%% file: UrdeitxEtAl.tex
\documentclass{article}

\usepackage{PRIMEarxiv}
\usepackage{hyperref}
\usepackage{algpseudocode}
\usepackage{algorithm}
\usepackage{xcolor}     
\usepackage{amsmath}    
\usepackage[utf8]{inputenc} 
\usepackage[T1]{fontenc}    
\usepackage{hyperref}       
\usepackage{url}            
\usepackage{booktabs}       
\usepackage{amsfonts}       
\usepackage{nicefrac}       
\usepackage{microtype}      
\usepackage{lipsum}
\usepackage{fancyhdr}       
\usepackage{graphicx}       
\graphicspath{{media/}}     
\usepackage{accents}

\newcommand{\bs}[1]{\boldsymbol{#1}}

\title{A comparison of Single and Double Generator Formalisms for Thermodynamics-Informed Neural Networks
}

\author{
  P. Urdeitx$^{1}$, I. Alfaro$^{1}$, D. Gonz\'alez$^{1}$, F. Chinesta$^{2,3}$, E. Cueto$^{1}$  \vspace{0.5cm}\\
  $^{1}$ ESI Group-UZ Chair of the National Strategy on Artificial Intelligence. \protect \\Aragon Institute of Engineering Research (I3A). Universidad de Zaragoza. Zaragoza, Spain. \\
  $^{2}$ ESI Group Chair. PIMM lab., ENSAM Arts et Métiers Institute of Technology, Paris, France. \\
$^3$ CNRS@CREATE LTD. Singapore.    \\
}

\begin{document}
\maketitle

\begin{abstract}
The development of inductive biases has been shown to be a very effective way to increase the accuracy and robustness of neural networks, particularly when they are used to predict physical phenomena. These biases significantly increase the certainty of predictions, decrease the error made and allow considerably smaller datasets to be used.

There are a multitude of methods in the literature to develop these biases. One of the most effective ways, when dealing with physical phenomena, is to introduce physical principles of recognised validity into the network architecture.

The problem becomes more complex without knowledge of the physical principles governing the phenomena under study. A very interesting possibility then is to turn to the principles of thermodynamics, which are universally valid, regardless of the level of abstraction of the description sought for the phenomenon under study.

To ensure compliance with the principles of thermodynamics, there are formulations that have a long tradition in many branches of science. In the field of rheology, for example, two main types of formalisms are used to ensure compliance with these principles: one-generator and two-generator formalisms. In this paper we study the advantages and disadvantages of each, using classical problems with known solutions and synthetic data.

\end{abstract}


\section{Introduction}

Since the recent re-emergence of machine learning after some ``artificial intelligence winters'', with neural networks and deep learning as major players, there has been a growing interest in constraining or controlling such learning, moving from "black box" learning to "grey box" learning for scientific machine learning purposes \cite{Cuomo2022,cranmer2020lagrangian,mattheakis2022hamiltonian}. When learning physical phenomena, the predictability and accuracy of the results become a major requisite. The imposition of certain mathematical or data structures, which allow us to establish inductive biases on the learned systems, has given rise to the development of different families of neural networks capable of learning the physical evolution of a system from the data to a great accuracy. Among them, Physics-Informed Neural Networks (PINNs) stand out, in which the learning algorithm tries to fit the solution to a known equation, defined by the governing partial differential equation, from the data \cite{Raissi2017, Mialon2023, Pichi2023, Sosanya2022,cai2021physics}. Taking advantage of the symmetries in the data, seen from a thermodynamic perspective, and imposing a specific, well-known structure on the evolution of its state variables of a dynamic system, the Structure Preserving Neural Networks (SPNN) and Hamiltonian Neural Networks (HNN), among others, can be found in the literature \cite{Greydanus2019, Hernandez2021a, Yu2021, Gruber2023}. The application of structures or formalisms during the learning process allows conservation laws (system symmetries) to be learned without direct supervision. These methods, which employ inductive biases, have been shown not only to be able to learn the dynamics of complex systems, ensuring the fulfillment of basic laws of thermodynamics but also these constraints can improve the robustness of the method during inference, limiting the appearance of incoherent responses. A review of the evolution of the integration of known physics---particularly, thermodynamics---in neural networks can be found in \cite{Cueto2023}.

These restrictions, however, do not always represent an advantage during learning. There is a trade-off between the expressiveness of a network, i.e., its ability to model complex functions, and the learnability of a network, the capability of a machine learning model to acquire knowledge or improve its performance from data \cite{zhang2017learnability}. Completely unrestricted (black box) networks represent the maximum expressiveness of the network but are often unable to capture the underlying physics of the problem for previously unseen situations, while by incorporating inductive biases, the representativeness increases at the cost of compromising the learning process.

From a physics perspective, different approaches have been presented for the representation of the evolution of a dynamical system out of equilibrium \cite{Jou}. At the molecular dynamics level, for instance, Newton's laws (or their equivalent Hamiltonian or Lagrangian alternatives) are sufficient for the description of the system. At this scale, everything is reversible or conservative. However, this entails the control of position and momentum of a number of molecules of the order of the Avogadro number at each instant of time, being unfathomable except in very specific cases. As we move from the microscopic to a meso and macroscopic description of the system, tracking the state variables becomes intricate, since unresolved variables play a role in the evolution of the physics, thus introducing the dependence on history \cite{ma2018model,gonzalez2021learning}. This lack of information is generally associated with a corrective term that allows us to go from an ideal system (reversible) to a real system (irreversible) which is associated with the generation of entropy of the system. In this sense, the total energy of the system is conserved, as dictated by the first law of thermodynamics.

A convenient way to define the dynamics of non-equilibrium systems is through a generalization of the Poisson bracket with its extension for irreversible systems with the so-called dissipative bracket \cite{Kaufman1984, Grmela1984, Morrison1984}. This way of defining systems compacts the main properties (invariants) by carefully defining the operators as well as considering Casimir invariants as constraints in the system \cite{Morrison1986}. The choice of one set of variables or another to represent the system can give rise to different bracket formalisms derived from this generalization of the Hamiltonian for non-conservative systems \cite{Edwards1998, Edwards1998a, Beris2001}. 

The level of compaction in the description of complex dynamical systems together with the preservation of the mathematical and thermodynamic properties of the system, make these formalisms ideal for learning systems through neural networks \cite{Gruber2023}. In this sense, structure-preserving models have demonstrated better performance regarding the use of black boxes, reinforcing the hypothesis of the benefit of considering these structures to shape inductive biases in data-driven models \cite{Hernandez2022, Gonzalez2019, Sosanya2022}. 

This paper analyzes the learning of different physical phenomena through the imposition of two alternative formalisms, with a thermodynamically consistent structure, widely used in the field of rheology, among other fields: single generator bracket and double generator bracket \cite{Eldred2020}. Recently, more elaborated formalisms of this type have been proposed that employ a 4-bracket formalism, but these, in general, will hinder the learning process \cite{zaidni2024thermodynamically}.  Both formalisms are mathematical structures for the description of dynamics in non-equilibrium systems, in which reversible mechanics is described by Hamilton's principle of least action and irreversible dynamics is a bilinear dissipation term \cite{Edwards1998, Edwards1998a}. The difference between the two structures analyzed lies in the definition of the energy generator functionals. While the single generator formalism defines a single generator, metriplectic systems, such as GENERIC, are defined with two generators, one associated with the reversible dynamics and the other with the dissipative part \cite{Grmela1984,grmela1997dynamics}. Although there are correlations between both formulations, and the transformation from one formalism to another can be obtained theoretically, the consideration of a generator or two establishes key differences that can be relevant during learning processes. Thus, the objective of this paper is to examine these differences and the advantages and limitations they confer to develop structure-preserving Neural Network systems. The pros and cons of each of the formulations will be analysed by considering two different problems, one discrete and one continuous, after the appropriate discretisation by means of finite elements. The different parameters affecting the learning process are analysed in detail and conclusions are drawn on the advantages and disadvantages of each method.

\section{Methods}
\subsection{Single and double generator bracket formalisms}

In 1984 different authors presented distinct, although very similar in spirit, formulations for irreversible phenomena as an extension of the classical Hamiltonian approach \cite{Kaufman1984, Grmela1984, Morrison1984}. For instance, \cite{Morrison1986} starts by considering that equilibrium is achieved by extremizing the energy at constant entropy,
\begin{equation}
\mathcal{F}_{\lambda}=\mathcal{H}+\lambda \mathcal{S},
\end{equation}
being $\mathcal{F}$ the generalized free energy of the system, $\lambda$ a Lagrange multiplier, and where $\mathcal{H}$ and $\mathcal{S}$, correspond to the Hamiltonian, and the entropy of the system, respectively. In general, one form of introducing dissipation in the Hamiltonian description of a system is by adding a Casimir or generalized entropy functional. Casimirs are functionals that are conserved for all Hamiltonians. Therefore, by considering Casimirs, we can drop the Lagrange multiplier to arrive at a very convenient description of the free energy of our system in the form
\begin{equation}
\mathcal{F}=\mathcal{H}+ \mathcal{S}.
\end{equation}

Based on the generalized free energy, different descriptions of the energy can be proposed which can be used to define different bracket formalisms \cite{Beris2024, Yu2021, Eldred2020}. If we consider a single energy potential, $\mathcal{F}$, the free energy in the system, the equations of motion for a set of state variables $\bs z$ will be:
\begin{equation} \label{General_Bracket}
\frac{d \boldsymbol{z}}{d t} = \{\{ \boldsymbol{z}, \mathcal{F} \} \} ,
\end{equation}
where the double braces are employed to denote a dissipative generalization of the Poisson bracket. Since any operator can be split into the self-adjoint and anti-self-adjoint parts, we arrive at 
\begin{equation} \label{General_Bracket2}
\frac{d \boldsymbol{z}}{d t} = \{\{ \boldsymbol{z}, \mathcal{F} \} \} = \{ \boldsymbol{z}, \mathcal{F} \} + [ \boldsymbol{z}, \mathcal{F} ]. 
\end{equation}

By considering the description of the brackets \cite{Kaufman1984}, this system can be written in the algebraic form with two operators, $\boldsymbol{L}$ and $\boldsymbol{M}$ as:
\begin{equation}
    \boldsymbol{L} : T^{\textasteriskcentered}\mathcal{M} \rightarrow T \mathcal{M}, \boldsymbol{M} : T^*\mathcal{M} \rightarrow T \mathcal{M},
\end{equation}
being $T^* \mathcal{M}$, and $T \mathcal{M}$ the cotangent and tangent bundles of $\mathcal{M}$, respectively. Eq. (\ref{General_Bracket}) can thus be rewritten as \cite{Kaufman1984}:
\begin{equation}\label{SBracket}
    \{\{ \boldsymbol{z}, \mathcal{F} \} \} =  \boldsymbol{L} \frac{\partial \mathcal{F}}{\partial \boldsymbol{z}} + \boldsymbol{M} \frac{\partial \mathcal{F}}{\partial \boldsymbol{z}}. 
\end{equation}
The operator $\boldsymbol{L}$ is the symplectic matrix, which is defined as a skew-symmetric matrix, while the operator $\boldsymbol{M}$ is the dissipative matrix, defined as a positive semi-definite matrix.

By decomposing Eq. (\ref{SBracket}) into the Hamiltonian or conservative energy, $\mathcal{H}$, and the entropy, $\mathcal{S}$, a two-generator bracket can be defined as:
\begin{equation}
\frac{d \boldsymbol{z}}{d t} = \{ \boldsymbol{z}, (\mathcal{H} + \mathcal{S}) \} + [ \boldsymbol{z}, (\mathcal{H} + \mathcal{S}) ],
\end{equation}
which also can be written as:
\begin{equation}\label{eq_e_s}
    \frac{d \boldsymbol{z}}{d t} = \boldsymbol{L} \frac{\partial (\mathcal{H} + \mathcal{S})}{\partial \boldsymbol{z}} + \boldsymbol{M} \frac{\partial (\mathcal{H} + \mathcal{S})}{\partial \boldsymbol{z}} .
\end{equation}

To ensure (i) conservation of the total energy $d\mathcal{H}/dt=0$, and (ii) non-negative entropy production $d\mathcal{S}/dt\ge0$, two additional conditions must be fulfilled. To ensure energy conservation we must impose that
\begin{equation}
    \frac{\partial \mathcal{H}}{\partial \boldsymbol{z}} \boldsymbol{M} = \boldsymbol{0},
\end{equation}
and, equivalently, to ensure non-negative entropy production,
\begin{equation}
    \frac{\partial \mathcal{S}}{\partial \boldsymbol{z}} \boldsymbol{L} = \boldsymbol{0}.  
\end{equation} 
In this way, it is straightforward to prove that the conservation of energy is obtained through
\begin{equation}
    \frac{d\mathcal{H}}{dt}=\frac{\partial \mathcal{H}}{\partial \boldsymbol{z}} \left( \boldsymbol{L} \frac{\partial \mathcal{H}}{\partial \boldsymbol{z}} + \boldsymbol{M} \frac{\partial \mathcal{S}}{\partial \boldsymbol{z}} \right) = 0, 
\end{equation} 
provided that
$$
\frac{\partial \mathcal{H}}{\partial \boldsymbol{z}} \boldsymbol{L} \frac{\partial \mathcal{H}}{\partial \boldsymbol{z}} = 0.
$$
In turn, non-negative entropy production results from
\begin{equation}
    \frac{d\mathcal{S}}{dt}=\frac{\partial \mathcal{S}}{\partial \boldsymbol{z}} \left( \boldsymbol{L} \frac{\partial \mathcal{H}}{\partial \boldsymbol{z}} + \boldsymbol{M} \frac{\partial \mathcal{S}}{\partial \boldsymbol{z}} \right) \ge 0, 
\end{equation} 
given that 
$$
\frac{\partial \mathcal{S}}{\partial \boldsymbol{z}} \boldsymbol{M} \frac{\partial \mathcal{S}}{\partial \boldsymbol{z}} \ge 0,
$$
by the semi-positive definiteness of the matrix $\bs M$.

By imposing these two degeneracy conditions on the double generator formalism we thus arrive at the so-called “General Equation for Non-Equilibrium Reversible-Irreversible Coupling” formalism, GENERIC, as \cite{Hernandez2021}: 
\begin{equation}\label{doubleform}
\frac{\partial \boldsymbol{z}}{\partial t} = \{ \boldsymbol{z}, \mathcal{H} \} + [ \boldsymbol{z}, \mathcal{S} ] .
\end{equation}

Although the equivalence between the two formalisms is demonstrated theoretically, the choice of one or the other formalism has practical implications in the development of methods that have to determine the particular structure of the equations of motion of our system from data. Basically, both formalisms will need to determine the particular form of the matrices $\bs L$ and $\bs M$, but one will also need to determine the form of a single potential, see Eq. (\ref{General_Bracket2}), while the second formalism uses two, see Eq. (\ref{doubleform}). In the latter case, it will be necessary to explicitly require the fulfilment of the degeneracy conditions. This is likely to result in a method with a more general structure, but subject to more constraints, and which will have the undeniable advantage of the explicit imposition of the two principles of thermodynamics (conservation of energy, non-negative production of entropy).\cite{Edwards1998}.

Our problem will then be defined as finding the precise form of the evolution of the state variables of the system, $\boldsymbol{z}$, from experimental measurements on the system, given predetermined initial conditions, $\boldsymbol{z}(0)$:
\begin{equation}
    \boldsymbol{\dot{z}}= \frac{d\boldsymbol{z}}{dt}=f(\boldsymbol{z},t), \boldsymbol{x}\in \Omega \in \mathbb{R}^D, t\in \mathcal{I}=(0,T], \boldsymbol{z}(0) =\boldsymbol{z}_0, 
\end{equation}
where $\boldsymbol{x}$ are the spatial coordinates on a domain $\Omega$, and  $f(\boldsymbol{x},\boldsymbol{z},t)$ being the function that describes the flow map $\boldsymbol{z}_0 \rightarrow \boldsymbol{z}(\boldsymbol{z}_0, T)$ in a prescribed time horizon $T$. The use of inductive biases will consist precisely in assuming that the precise form of the function $f$ sought will be either Eq. (\ref{General_Bracket2}) or Eq. (\ref{doubleform}).

For this purpose, both formalisms will be discretised in time, so that
\begin{equation}
    \boldsymbol{z}({t+\Delta t})=\boldsymbol{z}_{t} + (\boldsymbol{L}^S + \boldsymbol{M}^S) \frac{\partial \mathcal{F}}{\partial \boldsymbol{z}} \Delta t, 
    \label{SB_formalism}
\end{equation}
where $S$ refers to the single generator bracket formalism, and:
\begin{equation}
\boldsymbol{z}({t+\Delta t})=\boldsymbol{z}_{t} + \left( \boldsymbol{L}^G \frac{\partial \mathcal{H}}{\partial \boldsymbol{z}} + \boldsymbol{M}^G \frac{\partial \mathcal{S}}{\partial \boldsymbol{z}} \right) \Delta t, 
    \label{G_formalism}
\end{equation}
where $G$ stands for the double generator bracket or GENERIC, for short, formalism. These will be omitted when there is no risk of confusion. Note that the scheme in Eq. (\ref{SB_formalism}) resembles closely the so-called OnsagerNet \cite{Yu2021}. In that case, however, the system is assumed to be close to equilibrium and matrix $\bs M$ is assumed to be constant. It also includes an autoencoder in its architecture so as to unveil the latent variables governing the problem. This approach is also present in \cite{Hernandez2021a}, but has not been considered here in order to keep the analysis as simple as possible.

\subsection{Thermodynamics-Informed Neural Networks}

\begin{figure} 
    \centering
    \def\svgwidth{0.98\textwidth}
    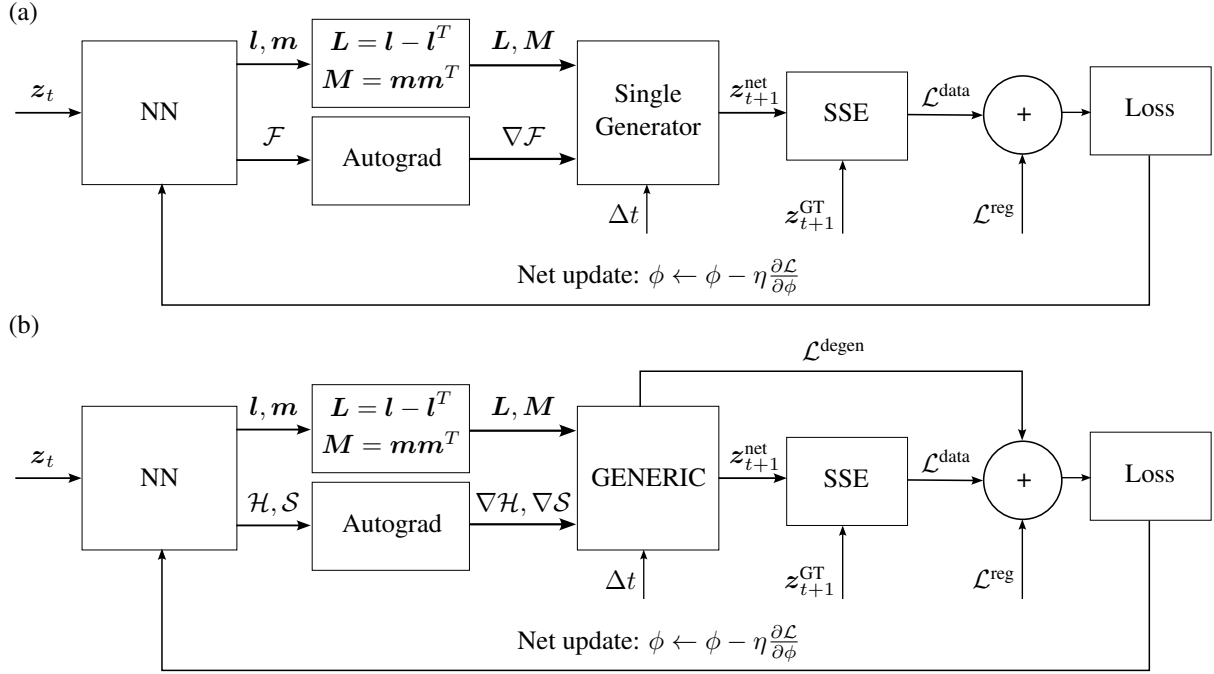
    \caption{Configuration scheme of the Structure Preserving Neural Network (SPNN) employed to learn single generator (a) and GENERIC (b) formalisms. The input parameters of the net are the state of the system, $\boldsymbol{z}(\bs x,t)$, at each time step. The output of the net includes the energy $\mathcal{F}$, $\mathcal{H}$, and $\mathcal{S}$, as well as the $\bs m$, and $\bs l$ components needed to reconstruct the formalism. The integration of the formalism gives the state of the system at the next time step $\boldsymbol{z}(\bs x,t+1)$. Then, the Data error and the degeneracy conditions are computed to define the loss of the net. }
    \label{SPNN_fig}
\end{figure}

Neural networks are well known to satisfy the universal approximation theorem, so the time evolution of the state variables, assumed in the form given by Eq. (\ref{General_Bracket2}) or by Eq. (\ref{doubleform}), will be determined using feed-forward neural networks (Fig \ref{SPNN_fig}). The input of the network is formed by the state vector of the system at each instant of time, $\boldsymbol{z}(\bs x,t)$, while the output of the net is taken as the parameters necessary for the reconstruction of the integration formalisms, including $\boldsymbol{L}(\bs x,t)$, and $\boldsymbol{M}(\bs x,t)$ operators, and the energy generators, $\mathcal{F}$, in the single generator formalism, and $\mathcal{H}$ and $\mathcal{S}$, in the GENERIC formlism. $\boldsymbol{L}$, the symplectic matrix, is well-known to be skew-symmetric. Therefore, it is more convenient to learn a matrix $\bs l$, such that $\bs L=\bs l-\bs l^T$. Conversely, $\boldsymbol{M}$, the friction matrix, is symmetric and positive semi-definite, so it is more convenient to learn a matrix $\bs m$ such that $\bs M=\bs m \bs m^T $ \cite{Hernandez2021}. Moreover, the gradient of the potentials is computed from the learned energy (scalar) by using the ${\tt autograd}$ function of PyTorch. 

The state of the variables of the system in the next time step, $\boldsymbol{z}_{n+1}=\boldsymbol{z}(\bs x,t+\Delta t)=\boldsymbol{z}(\bs x,(n+1)\Delta t)$, is then obtained by the integration with the reconstructed formalism, the current state of variables, $\bs z_n$, and a fixed time step increment, $\Delta t$. 

The loss function includes up to three contributions depending on the integration formalism. The first term of the loss function, the data loss, $\mathcal{L}^{\text{data}}_n$, enforces the agreement of the predicted values of the variables to the reference values. The data loss, $\mathcal{L}^{\text{data}}_n$, compares the mean square error between the predicted values, $\boldsymbol{z}_{n}^{\text{net}}$, and the ground truth values, $\boldsymbol{z}_{n}^{\text{GT}}$, throughout the time series, by employing the L2 norm. 
\begin{equation}
    \mathcal{L}^{\text{data}}_n = \parallel \boldsymbol{z}_{n+1}^{\text{GT}} - \boldsymbol{z}_{n+1}^{\text{net}} \parallel^2_2 .
\end{equation}
A second term of the loss, the degeneracy loss $\mathcal{L}^{\text{degen}}_n$, enforces the fulfillment of the  degeneracy conditions, 
\begin{equation}
    \mathcal{L}^{\text{degen}}_n = \parallel \boldsymbol{L} \nabla \mathcal{S} \parallel^2_2 + \parallel \boldsymbol{M} \nabla \mathcal H \parallel^2_2 , 
\end{equation}
A third term of the loss, the regularization loss, $\mathcal{L}^{\text{reg}}$, is considered to avoid the overfitting of the network, which is defined as the sum of the squared weight parameters of the network. 
\begin{equation}
    \mathcal{L}^{\text{reg}} = \sum_{l}^{L} \sum_{j}^{n^{[l]}} \sum_{i}^{n^{[l+1]}} (w_{i,j}^{[l]})^2,
\end{equation}
where $l$ is the index of the current network layer and $\bs{w}^{[l]}$ is the weight matrix of this same layer.  $n^{[l]}$ represents the number of neurons at layer $l\in [1,L]$.

Then, the global loss function is the sum of the contributions of the loss functions just considered. Due to the differences in the magnitude of each term in the loss, compensation weights were considered for the data and regularization losses,
\begin{equation}
     \mathcal{L} =  \sum_{n=1}^{N_T} (\lambda_d \mathcal{L}^{\text{data}}_n  + \mathcal{L}^{\text{degen}}_n )+ \lambda_r \mathcal{L}^{\text{reg}},
\end{equation}
where $\lambda_d$, and $\lambda_r$ were the data, and the regularization weight compensation hyperparameters, respectively. $N_T$ represents the number of snapshots in each simulation.

Based on the loss thus obtained, the parameters of the net  (weights and biases) are updated through the backpropagation algorithm with the gradient descent technique. This process is repeated for the fixed number of epoch $\tt n_{\text{epoch}}$, with a multistep learning rate scheduler with a decaying factor in $1/3 \tt n_{\text{epoch}}$, and $2/3 \tt n_{\text{epoch}}$. 

The training database was composed of the time series ($t \in \mathcal{I}(0, T]$) of a collection of different trajectories of the dynamic systems. The trajectories of the database were divided into training ($N_{\text{train}} = 80\%$ of the dataset) and test data ($N_{\text{test}} = 20\%$ of the dataset).

The performance of the network is evaluated based on the predicted state of the variables of the system by comparing them with the ground truth values by calculating the mean square error  (MSE) for all trajectories, and throughout all the time series, for every variable in the problem,
\begin{equation}
\text{MSE}^{\text{data}}(\boldsymbol{z})= \frac{1}{N_T} \sum_{n=1}^{N_T} (\bs z_{n}^{\text{GT}}-\bs z_{n}^{\text{net}})^2.
\end{equation}
The pseudocode of the training and test methods for the single generator bracket are presented in Algorithm \ref{SB_SPNN} and \ref{SB_SPNN_test}, respectively, while the pseudocode of the training and test for the GENERIC formalism can be seen in Algorithm \ref{GB_SPNN} and \ref{GB_SPNN_test}, respectively.

\begin{algorithm}
\caption{Pseudocode for the training algorithm of the single generator bracket SPNN}\label{SB_SPNN}
\begin{algorithmic}
\State Load train database: $\boldsymbol{z}^{\text{GT}}$(train partition), $\Delta t$;
\State Initialize $w_{i}, b_{i}$;
\For{$\text{epoch} \gets 1$ to $\tt n_{\text{epoch}}$}
    \For{$\text{train case} \gets 1$ to $N_{\text{train}}$}
        \State  Initialize state vector: $\bs z_{0}^{\text{net}}$ is $\bs z_{0}^{\text{GT}}$;
        \State  Initialize losses: $\mathcal{L}^{\text{data}}=0$;
        \For{$\text{snapshot} \gets 1$ to $N_{T}$}
            \State  Forward propagation: $[\bs l, \bs m, \mathcal{F}] \gets \text{Net}(\bs z_{t}^{\text{GT}})$;
            \State  Take the Energy gradient (autograd of PyTorch): $\nabla \mathcal{F} \gets \nabla_{\bs z} \mathcal{F}$;
            \State  Formalism construction: $\boldsymbol{L} \gets \bs l-\bs l^T$, $\boldsymbol{M} \gets \bs m·\bs m^T$;
            \State  Time integration $\bs z_{t+1}^{\text{net}} \gets \bs z_{t}^{\text{net}} + \Delta t (\boldsymbol{L} + \boldsymbol{M})\nabla \mathcal F$;
            \State  Update data loss $\mathcal{L}^{\text{data}} \gets \mathcal{L}^{\text{data}} + \mathcal{L}^{\text{data}}_n$;
            \State  Update degeneracy loss $\mathcal{L}^{\text{degen}} \gets \mathcal{L}^{\text{degen}} + \mathcal{L}^{\text{degen}}_n$;
        \EndFor
        \State  SSE loss function: $\mathcal L \gets \lambda_d \mathcal{L}^{\text{data}} +  \lambda_r \mathcal{L}^{\text{reg}}$
        \State  Backward propagation;
        \State  Optimizer step;
    \EndFor
    \State  Learning rate scheduler;
\EndFor
\end{algorithmic}
\end{algorithm}  

\begin{algorithm}
\caption{Pseudocode for the test algorithm of the single generator bracket SPNN}\label{SB_SPNN_test}
\begin{algorithmic}
\State Load test database: $\boldsymbol{z}^{\text{GT}}$(test partition), $\Delta t$;
\State Load network parameters;
\For{$\text{test case} \gets 1$ to $N_{\text{test}}$}
    \State  Initialize state vector: $\bs z_{0}^{\text{net}}$ is $\bs z_{0}^{\text{GT}}$;
    \For{$\text{snapshot} \gets 1$ to $N_{T}$}
        \State  Forward propagation $[\bs l, \bs m, \mathcal F] \gets \text{Net}(\bs z_{t}^{\text{GT}})$;
        \State  Formalism construction: $\boldsymbol{L} \gets \bs l-\bs l^T$, $\boldsymbol{M} \gets \bs m·\bs m^T$;
        \State  Take the Energy gradient (autograd of PyTorch): $\nabla \mathcal{F} \gets \nabla_{\bs z} \mathcal F$;
        \State  Time integration $\bs z_{t+1}^{\text{net}} \gets \bs z_{t}^{\text{net}} + \Delta t (\boldsymbol{L} + \boldsymbol{M})\nabla \mathcal F$;
    \EndFor
    \State  Compute $\text{MSE}$;
\EndFor
\end{algorithmic}
\end{algorithm}

\begin{algorithm}
\caption{Pseudocode for the training algorithm of the GENERIC SPNN}\label{GB_SPNN}
\begin{algorithmic}
\State Load train database: $\boldsymbol{z}^{\text{GT}}$(train partition), $\Delta t$;
\State Initialize $w_{i}, b_{i}$;
\For {epoch $\gets 1$ to $\tt n_{\text{epoch}}$}
    \For{train case $\gets 1$ to $N_{\text{train}}$}
        \State  Initialize state vector: $\bs z_{0}^{\text{net}}$ is $\bs z_{0}^{\text{GT}}$;
        \State  Initialize losses: $\mathcal{L}^{\text{data}},\mathcal{L}^{\text{deg}}=0$;
        \For{snapshot $\gets 1$ to $N_{T}$}
            \State  Forward propagation: $[\bs l, \bs m, \mathcal{H}, \mathcal{S}] \gets \text{Net}(\bs z_{t}^{\text{GT}})$;
            \State  Take the Energy gradient (autograd of PyTorch): $\nabla \mathcal{H} \gets \nabla_{\bs z} \mathcal{H}, \nabla \mathcal{S} \gets \nabla_{\bs z} \mathcal{S}$;
            \State  Formalism construction: $\boldsymbol{L} \gets \bs l-\bs l^T$, $\boldsymbol{M} \gets \bs m·\bs m^T$;
            \State  Time integration $\bs z_{t+1}^{\text{net}} \gets \bs z_{t}^{\text{net}} + \Delta t (\boldsymbol{L} \nabla \mathcal H + \boldsymbol{M} \nabla \mathcal S)$;
            \State  Update data loss $\mathcal{L}^{\text{data}} \gets \mathcal{L}^{\text{data}} + \mathcal{L}^{\text{data}}_n$;
            \State  Update degeneracy loss $\mathcal{L}^{\text{degen}} \gets \mathcal{L}^{\text{degen}} + \mathcal{L}^{\text{degen}}_n$;
        \EndFor
        \State  SSE loss function: $\mathcal L \gets \lambda_d \mathcal{L}^{\text{data}} + \mathcal{L}^{\text{degen}} +  \lambda_r \mathcal{L}^{\text{reg}}$
        \State  Backward propagation;
        \State  Optimizer step;
    \EndFor
    \State  Learning rate scheduler;
\EndFor
\end{algorithmic}
\end{algorithm} 

\begin{algorithm}
\caption{Pseudocode for the test algorithm of the GENERIC SPNN}\label{GB_SPNN_test}
\begin{algorithmic}
\State Load test database: $\boldsymbol{z}^{\text{GT}}$(test partition), $\Delta t$;
\State Load network parameters;
\For{test case $\gets 1$ to $N_{\text{test}}$}
    \State  Initialize state vector: $\bs z_{0}^{\text{net}}$ is $\bs z_{0}^{\text{GT}}$;
    \For{snapshot $\gets 1$ to $N_{T}$}
        \State  Forward propagation $[\bs l, \bs m, \mathcal H, \mathcal S] \gets \text{Net}(\bs z_{t}^{\text{GT}})$;
        \State  Formalism construction: $\boldsymbol{L} \gets \bs l-\bs l^T$, $\boldsymbol{M} \gets \bs m·\bs m^T$;
        \State  Take the Energy gradient (autograd of PyTorch): $\nabla \mathcal{H} \gets \nabla_{\bs z} \mathcal H$,  $\nabla \mathcal{S} \gets \nabla_{\bs z} \mathcal S$;
        \State  Time integration $\bs z_{t+1}^{\text{net}} \gets \bs z_{t}^{\text{net}} + \Delta t (\boldsymbol{L} \nabla \mathcal{H} + \boldsymbol{M} \nabla \mathcal S)$;
    \EndFor
    \State  Compute $\text{MSE}$;
\EndFor
\end{algorithmic}
\end{algorithm}

\section{Numerical results}

\subsection{Double thermoelastic pendulum}

\subsubsection{System description}

\begin{figure} 
    \centering
    \includegraphics[width=0.65\textwidth]{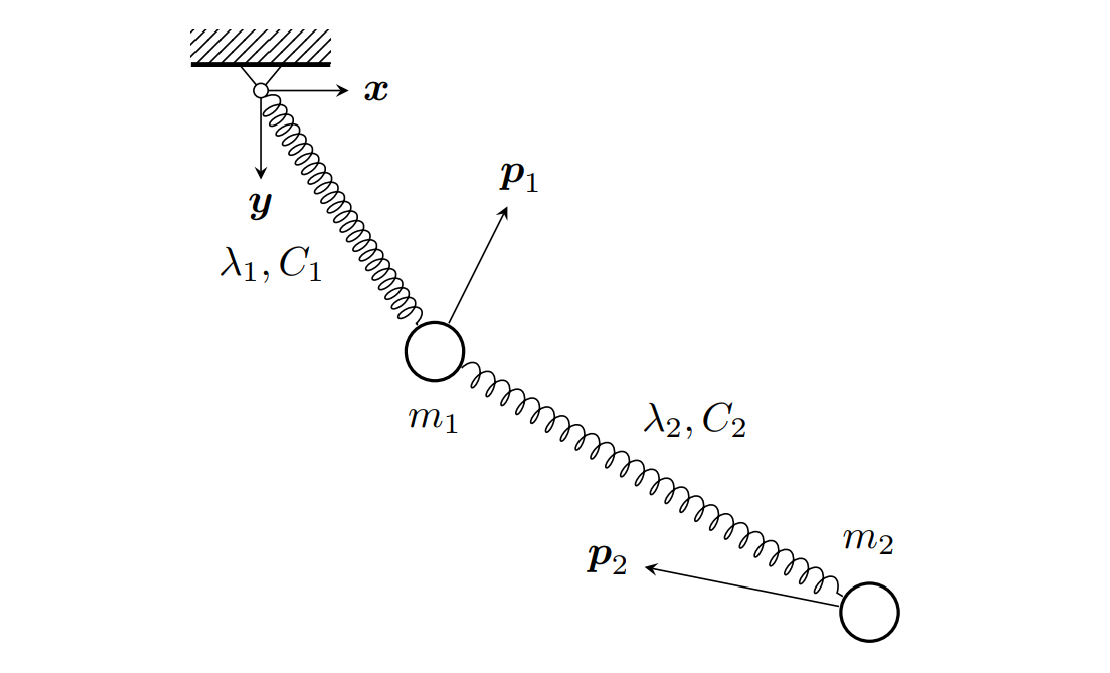}
    \caption{Double thermo-elastic pendulum system.}
    \label{DP_fig}
\end{figure}

The first example considers a double thermoelastic pendulum, which consists of two masses, $m_1$ and $m_2$, connected by two springs with natural lengths $\lambda_1^0$, and $\lambda_2^0$ (Fig. \ref{DP_fig}) \cite{Romero2009}. This model includes thermal effects due to the Gough-Joule effects, including the heat flux between springs (dissipative dynamics), and movements of masses (Hamiltonian mechanics). The set of variables which describe the system are:
\begin{equation}\label{Z_DP}
    S= \{ \boldsymbol{Z}(x,t)=(\boldsymbol{q}_1, \boldsymbol{q}_2, \boldsymbol{p}_1, \boldsymbol{p}_2, s_1, s_2) \in (\mathbb{R}^2 \times \mathbb{R}^2 \times \mathbb{R}^2 \times \mathbb{R}^2 \times \mathbb{R} \times \mathbb{R}) \}, 
\end{equation}
being, $\boldsymbol{q}_i$, $\boldsymbol{p}_i$, and $s_i$ the position, linear momentum, and entropy of each mass of the system.

The total energy of the system is defined by the sum of the kinetic energy of the masses, $K_i$, and the internal energy of the springs, $e_i$, as:
\begin{equation}\label{energyspring}
    E(\boldsymbol{z})= \sum_i \left( K_i(\boldsymbol{z}) + e_i(\lambda_i, s_i) \right),
\end{equation}
being $\lambda_i$ defined by the position of the masses as:
\begin{equation}
\lambda_1=\sqrt{\boldsymbol{q}_1·\boldsymbol{q}_1}, \ \lambda_2=\sqrt{(\boldsymbol{q}_2-\boldsymbol{q}_1)·(\boldsymbol{q}_2-\boldsymbol{q}_1)}, 
\end{equation}
and the kinetic energy, $K_i$, as:
\begin{equation}
    K_i=\frac{1}{2 m_i} |\boldsymbol{p}_i|^2.
\end{equation}

\subsubsection{Net hyperparameters and database}

A thermodynamically consistent algorithm following \cite{Romero2009} and implemented in MATLAB has been employed to generate the training database. The parameters associated with the generation of the synthetic data include the weights of the masses $m_1=1$ kg and $m_2=2$ kg, the natural lengths of the springs, $\lambda_1=2$ m and $\lambda_2=1$ m, and the thermal constant $C_1=0.02$ J and, $C_2=0.2$ J, of the first and second pendulum, respectively (see Fig. \ref{DP_fig}). The conductivity is $\kappa = 300$ and the simulation time is $T=60$ s in time increments of $\Delta t = 0.3$ s ($N_t=200$ snapshots). The database, state vector $\boldsymbol{Z}(x,t)$, Eq. (\ref{Z_DP}), contains $N_{x}=50$ different trajectories, split randomly into $40$ train and $10$ test trajectories. Each trajectory has been obtained with mean initial conditions around the initial position of $\boldsymbol{q}_1= [4.5, 4.5]^T$ m, and $\boldsymbol{q}_2= [2.0, 4.5]^T$ m, and initial momentum of $\boldsymbol{p}_1= -0.5, 1.5]^T$ kg·m/s, and $\boldsymbol{p}_2= [1.4, -0.2]^T$ kg·m/s (variations of $5\%$ around the mean initial position of $\boldsymbol{q}_1$, and $\boldsymbol{p}_1$ have been simulated for $20$ s to take the initial positions). 

\begin{figure} 
    \centering
    \def\svgwidth{0.98\textwidth}
    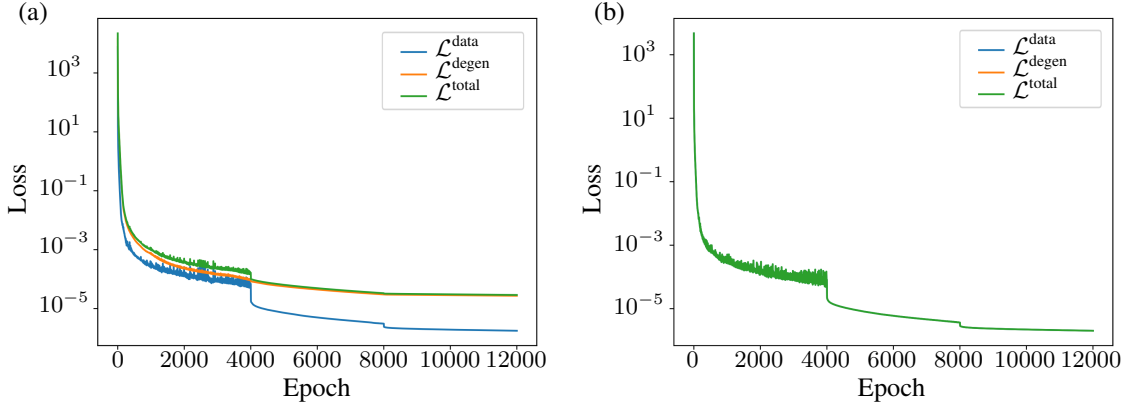
    \caption{Loss representation in training for GENERIC (a) and single generator (b) formalisms (Log scale). The loss function of the GENERIC includes data loss, $\mathcal{L}^{\text{data}}$, and degeneracy loss $\mathcal{L}^{\text{degen}}$, while the only contribution on the single generator is the data loss, $\mathcal{L}^{\text{data}}$. }
    \label{DP_Loss}
\end{figure}

The net is composed of an input layer, $N_{\text{in}}=10$, and an output layer, whose size depends on the chosen formalism, as $N_{\text{out}}^G=N_{\text{in}}^2+2=102$ and $N_{\text{out}}^S=N_{\text{in}}^2+1=101$, for GENERIC and single generator bracket, respectively. The number of hidden layers in both cases is $N_{\text{hidden}}=5$ with softplus function activation and with $N_{h}=2N_{\text{in}}^2=200$ units of neurons each. It is initialized according to the Kaiming method \cite{He2015}, with normal distribution, and the optimizer used is Adam \cite{Kingma2014}, with a weight decay of $\lambda_r= 10^{-5}$ and data loss weight of $\lambda_d = 10^2$. A total number of epochs of $N_{\text{epoch}}=12000$, with a multistep learning rate scheduler, is used, starting in $\mu = 10^{-4}$ and decaying by a factor of $\gamma=0.1$ in epochs $4000$, and $8000$ ($1/3·N_{\text{epoch}}$, and $2/3·N_{\text{epoch}}$, respectively). The evolution of the terms of data loss, $\mathcal{L}^{\text{data}}$, and degeneracy loss, $\mathcal{L}^{\text{degen}}$, have been represented in Fig. \ref{DP_Loss}. 

\subsubsection{Results}

\begin{figure} 
    \centering
    \def\svgwidth{0.97\textwidth}
    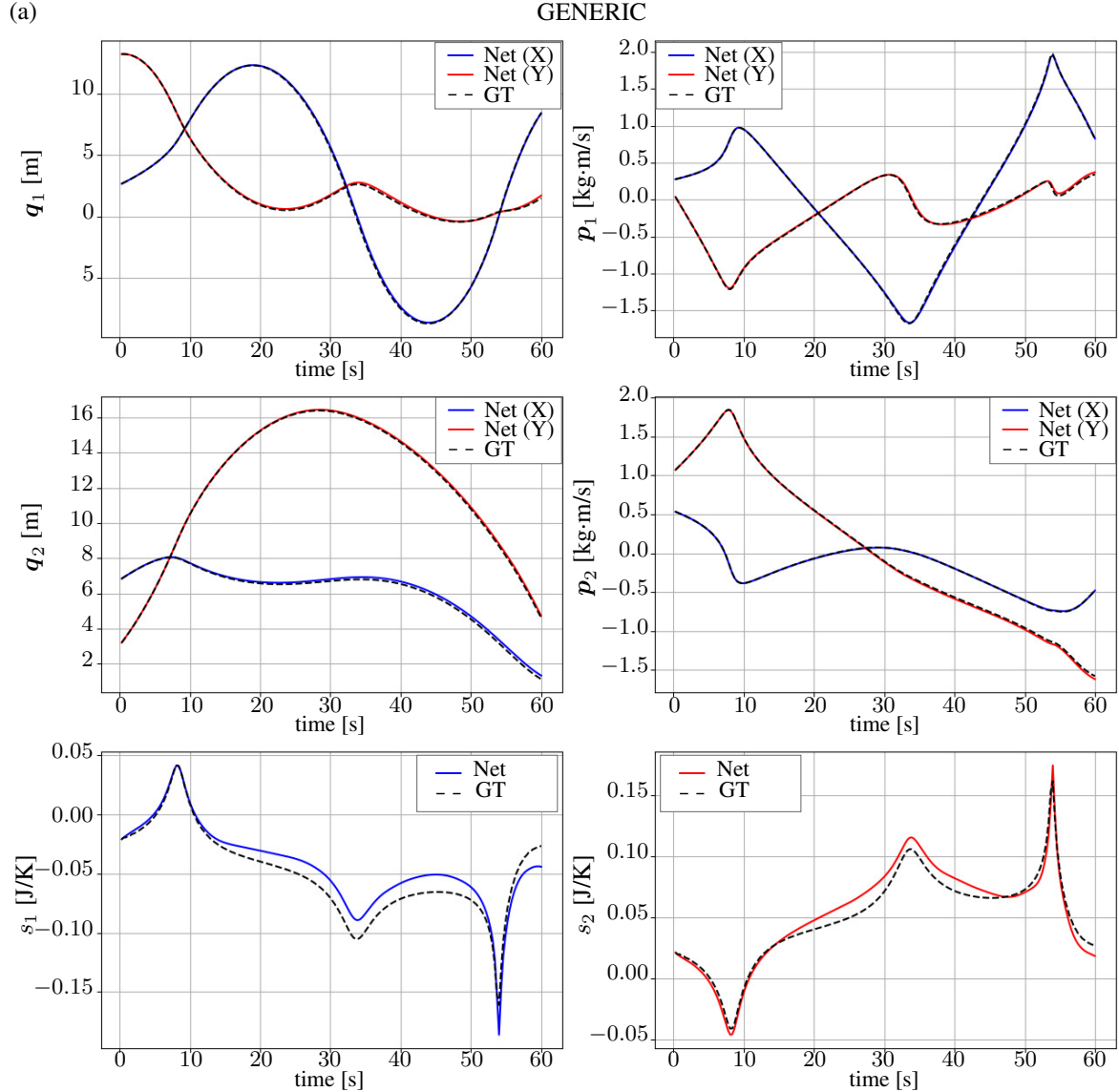
    \caption{Results of the reconstruction of the double thermo-elastic pendulum system. Test trajectory (Ground Truth, GT) and the reconstruction of the double thermo-elastic pendulum using time integration with GENERIC formalism. }
    \label{DP_Test_a} 
\end{figure}

\begin{figure} 
    \centering
    \def\svgwidth{0.97\textwidth}
    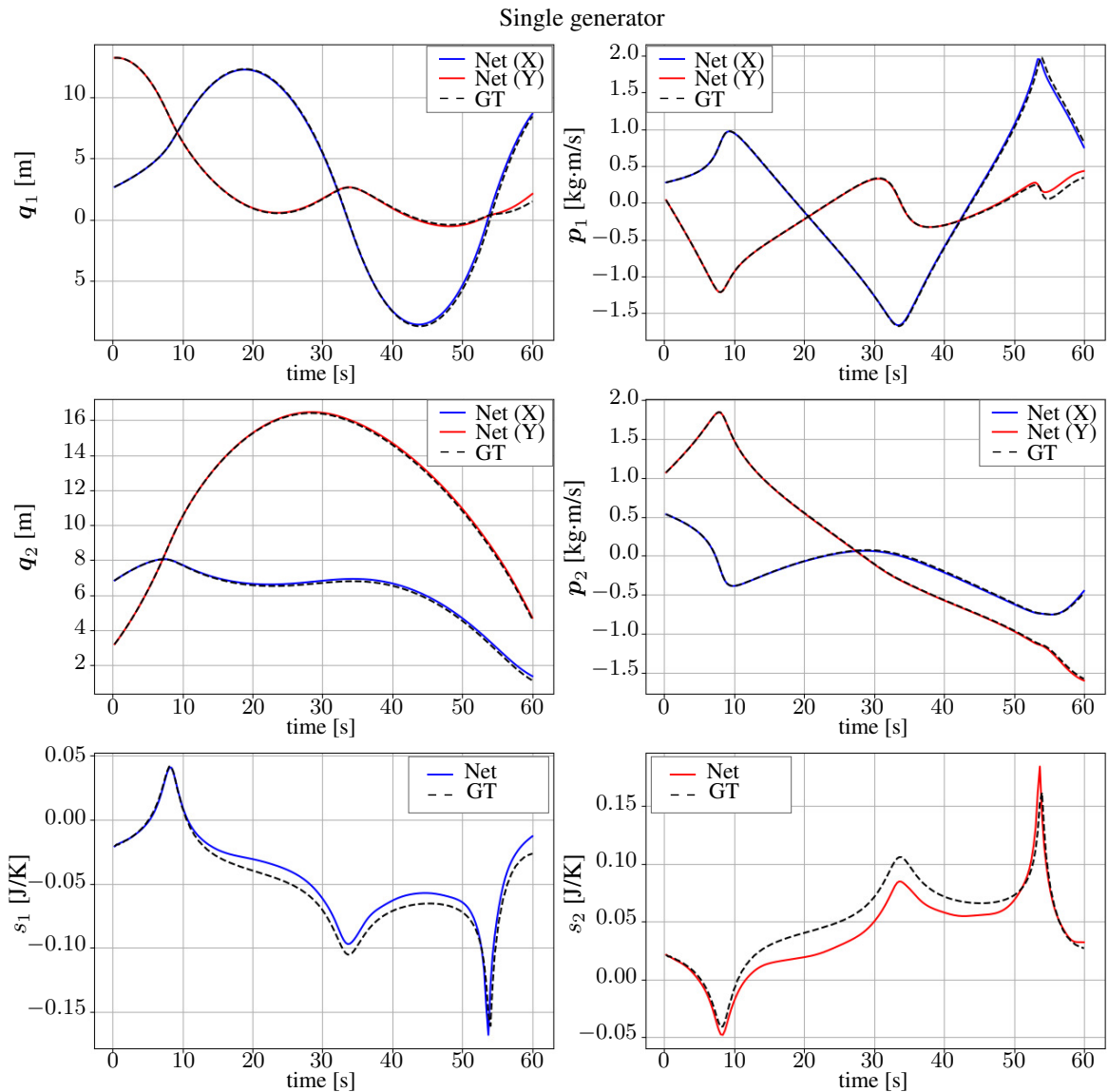
    \caption{Results of the reconstruction of the double thermo-elastic pendulum system. Test trajectory (Ground Truth, GT) and the reconstruction of the double thermo-elastic pendulum using time integration with single generator formalism. }
    \label{DP_Test_b} 
\end{figure}

\begin{figure} 
    \centering
    \def\svgwidth{0.98\textwidth}
    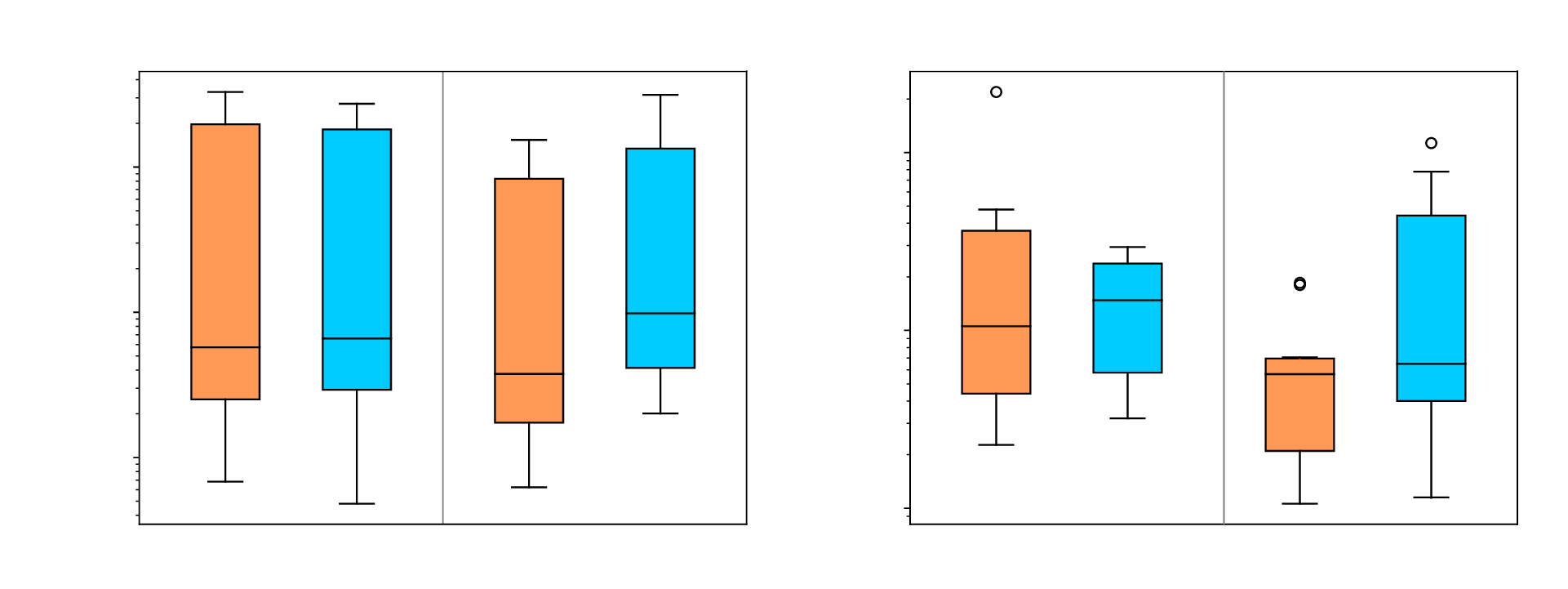
    \caption{Results of the Net. (a) MSE error in the system variables, obtained by the single generator and GENERIC formalisms. (b) Error in the reconstruction of the energy ($\mathcal{H}$) of the train and test trajectories calculated from the system variables. }
    \label{DP_Error}
\end{figure}

The state variables $\boldsymbol{z}_{n}^{\text{net}}$, obtained at each time increment $n$, from the reconstruction of the system with each formalism show a good degree of coherence with the synthetic ground truth data, $\boldsymbol{z}_{n}^{\text{GT}}$ (see Fig. \ref{DP_Test_a}, and Fig. \ref{DP_Test_b}). In both cases, the entropy variables show the highest error during the reconstruction. The comparison between the errors obtained in the reconstruction of the data does not show a significant difference between both formalisms (Fig. \ref{DP_Error} (a)). In the ground truth case, the error obtained in the train by the single generator formalism is lower than that obtained by the GENERIC formalism. However, GENERIC shows less error in the reconstruction of the test, previously unseen trajectories. 

To compare the thermodynamic consistency of both formalisms, the energy reconstructed with the predicted values of the state variables, $\mathcal{H}(\bs z_{n}^{\text{net}})$, with the GENERIC and single generator formalisms are compared to the real energy calculated from the system variables, $\mathcal{H}(\bs z_{n}^{\text{GT}})$) through Eq. (\ref{energyspring}). The error for the theoretical energy of the system has been represented in Fig. \ref{DP_Error} (b).

\subsection{Couette flow of an Oldroyd-B fluid}

\subsubsection{System description}

\begin{figure} 
    \centering
    \includegraphics[width=0.65\textwidth]{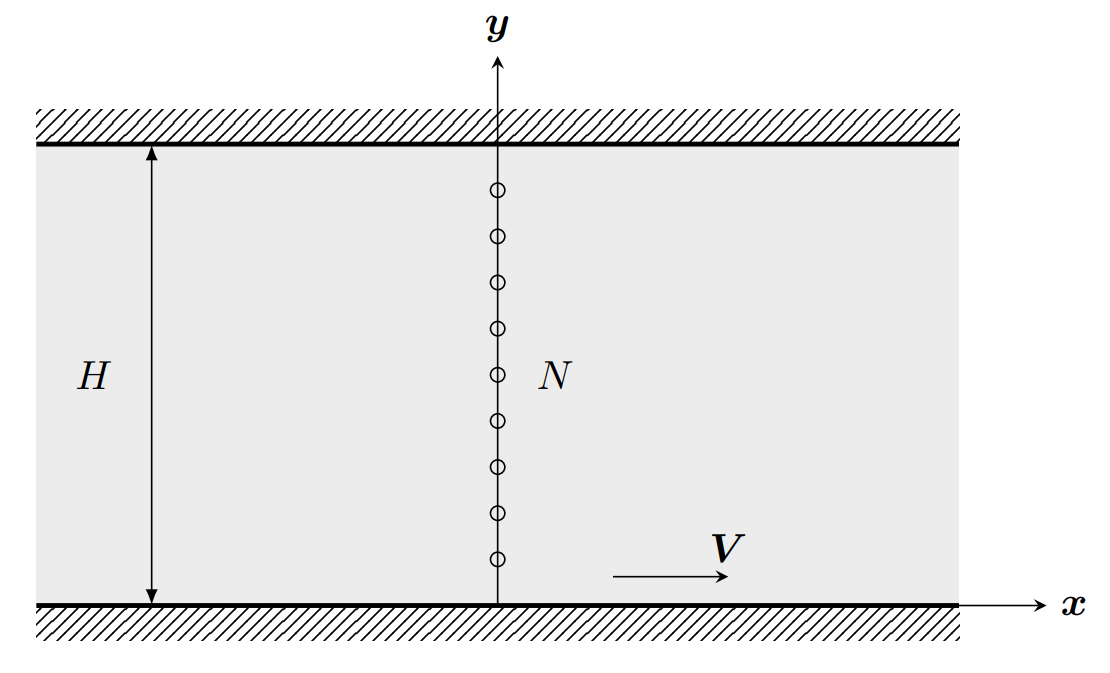}
    \caption{Couette flow in an Oldroyd-B fluid.}
    \label{VE_fig}
\end{figure}

The second example is an Oldroyd-B fluid within a shear flow (Couette flow), which can be modeled as a viscoelastic fluid composed of a series of linear elastic dumbbells (representing, for instance, the effect of polymer chains) immersed in a solvent \cite{Cherizol2015, Binns2024}. The dynamics of this system can be obtained from both, a macroscopic and microscopic perspective. The chosen set of variables to describe the system are (Fig. \ref{VE_fig}):
\begin{equation}\label{Z_VE}
    \mathcal D= \{\boldsymbol{z}(y,t)=(\boldsymbol{q}, \boldsymbol{v}, e, \tau) \in (\mathbb{R}^2 \times \mathbb{R} \times \mathbb{R} \times \mathbb{R}) \}, 
\end{equation}
being, $\boldsymbol{q}$, and $\boldsymbol{v}$, the position and velocity vectors, $e$, the internal energy, and, $\tau$ the stress-shear component of the conformal tensor. 

These parameters arise from the solution of the problem in two different scales. The macroscopic solution of the problem can be obtained through the Fokker-Plank equation, by applying the CONNFFESSIT technique, and by its transformation into the It\^o stochastic differential equation as \cite{Laso1993, Bris2009}:
\begin{alignat}{1}
    dq_x&= \left( \frac{\partial \boldsymbol{v}}{\partial y} q_y - \frac{1}{2\text{We}} q_x\right) dt + \frac{1}{\sqrt{\text{We}}}dV_t, \\
    dq_y&= -\frac{1}{2\text{We}}q_y dt + \frac{1}{\sqrt{\text{We}}}dW_t , 
\end{alignat}
being $\boldsymbol{v}=[v_x, v_y]^T$, and $\boldsymbol{q}=[q_x, q_y]^T$, the velocity and position vectors, $\text{We}$, the Weissenberg number, and $V_t$ and $W_t$ are two independent one-dimensional Brownian motions. Under the assumption of the Couette Flow, the dependencies of the positions are given by  $q_x=q_x(y,t)$ and $q_y=q_y(t)$. The solution of this equation can be obtained by Monte Carlo techniques, considering the empirical mean to replace the mathematical expectation. 

In the microscopic scale, the evolution of the conformation tensor $\boldsymbol{c}=\langle \boldsymbol{rr} \rangle$, describes the state of the dumbbells through the expected $\tau_{xy}$ component of this tensor. This acts as an internal variable in the system and is given by: 
\begin{equation}
    \tau_{xy}= \frac{\epsilon}{\text{We}}\frac{1}{K}\sum_{k=1}^{K} q_x q_y, 
\end{equation}
being $\epsilon=\frac{\nu_p}{\nu_s}$, the ratio of the polymer to solvent viscosities, and $K$ the number of dumbbells in the simulation.

For its part, the viscoelastic behavior of the model can be defined by the mechanical model of the solvent ($s$), as a Newtonian substrate, and polymer chains ($p$), as linear dumbbells diluted in the substrate. Thus, the deviatoric part, $\boldsymbol{T}$, of the stress tensor, $\boldsymbol{\sigma}$, is defined as:
\begin{equation}
    \boldsymbol{T} + \lambda_1 \stackrel{\nabla}{\boldsymbol{T}}  = \eta_0 \left( \boldsymbol{\dot\gamma} + \lambda_2 \stackrel{\nabla}{ \boldsymbol{\dot\gamma}} \right), 
\end{equation}
being, $\lambda_1$, $\lambda_2$, and, $\eta_0$, parameters of the model, $\boldsymbol{\dot\gamma}$ is the strain rate tensor, given by $\boldsymbol{\dot\gamma}=(\nabla^s \boldsymbol{v})=\boldsymbol{D}$, and where the triangle denotes non-linear Oldroyd upper convected derivative \cite{Cherizol2015}. 

Then, the total stress is given by the stress in the solvent ($s$) and polymer ($p$), given by:
\begin{equation}
    \boldsymbol{T} = \eta_s \boldsymbol{\dot\gamma} + \boldsymbol{\tau}, 
\end{equation}
so that
\begin{equation}
    \boldsymbol{\tau} + \lambda_1 \stackrel{\nabla}{\boldsymbol{\tau}} = \eta_p \boldsymbol{\dot\gamma}, 
\end{equation}
which is the constitutive equation of the elastic stress.

\subsubsection{Net hyperparameters and database}

A dimensionless multi-scale algorithm implemented in MATLAB has been employed to generate the training database. The parameters associated with the generation of the synthetic data include the lid velocity, $V=1$ m/s, the number of dumbbells at each node of the model, $N_d=10000$, the number of Reynolds, Re=$0.1$, and Weissenberg, We$=1.0$. The fluid vertical direction has been discretized with $N_x=100$ nodes, and the simulation time is $T=1$ s in time increments of $\Delta t = 0.0067$ s ($N_t=150$ snapshots). At the node at $h=H$ a condition of no-slip conditions has been imposed ($v=0$ m/s), therefore, it has been excluded from the database. The database state vector $\boldsymbol{z}(y,t)$, Eq. (\ref{Z_VE}), contains $N_{z}=100$ trajectories, split into $80$ training and $20$ test trajectories. 

\begin{figure} 
    \centering
    \def\svgwidth{0.98\textwidth}
    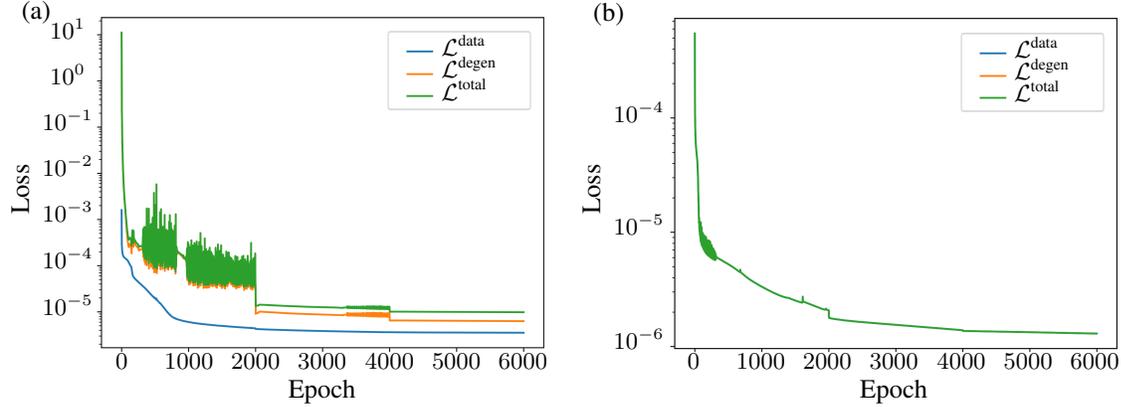
    \caption{Loss representation for the Couette flow in the Oldroyd Fluid system, for GENERIC (a), including data loss, $\mathcal{L}^{\text{data}}$, and degeneracy loss $\mathcal{L}^{\text{degen}}$, and single generator (b) whose only contribution is the data loss, $\mathcal{L}^{\text{data}}$ (Log scale). }
    \label{VE_Loss}
\end{figure}

The net is composed of an input layer, $N_{\text{in}}=5$, and an output layer, depending on the training Formalism, of $N_{\text{out}}^G=N_{\text{in}}^2+2=27$ and $N_{\text{out}}^S=N_{\text{in}}^2+1=26$, for GENERIC and single generator formalisms, respectively. The number of hidden layers in both cases is $N_{\text{hidden}}=5$ with Softplus function activation and with $N_{h}=2N_{\text{in}}^2=50$ units of neurons each. It is initialized according to the Kaiming method \cite{He2015}, with normal distribution, and the optimizer used is Adam \cite{Kingma2014}, with a weight decay of $\lambda_r= 10^{-5}$ and data loss weight of $\lambda_d = 10^2$. A total number of epochs of $N_{\text{epoch}}=6000$, with a multistep learning rate scheduler, is used, starting in $\mu = 10^{-4}$ and decaying by a factor of $\gamma=0.1$ in epochs $2000$ and $4000$ ($1/3·N_{\text{epoch}}$, and $2/3·N_{\text{epoch}}$, respectively). The evolution of the terms of data loss, $\mathcal{L}^{\text{data}}$, and degeneracy loss, $\mathcal{L}^{\text{degen}}$, has been represented in Fig. \ref{VE_Loss}. 

\subsubsection{Results}

\begin{figure} 
    \centering
    \def\svgwidth{0.95\textwidth}
    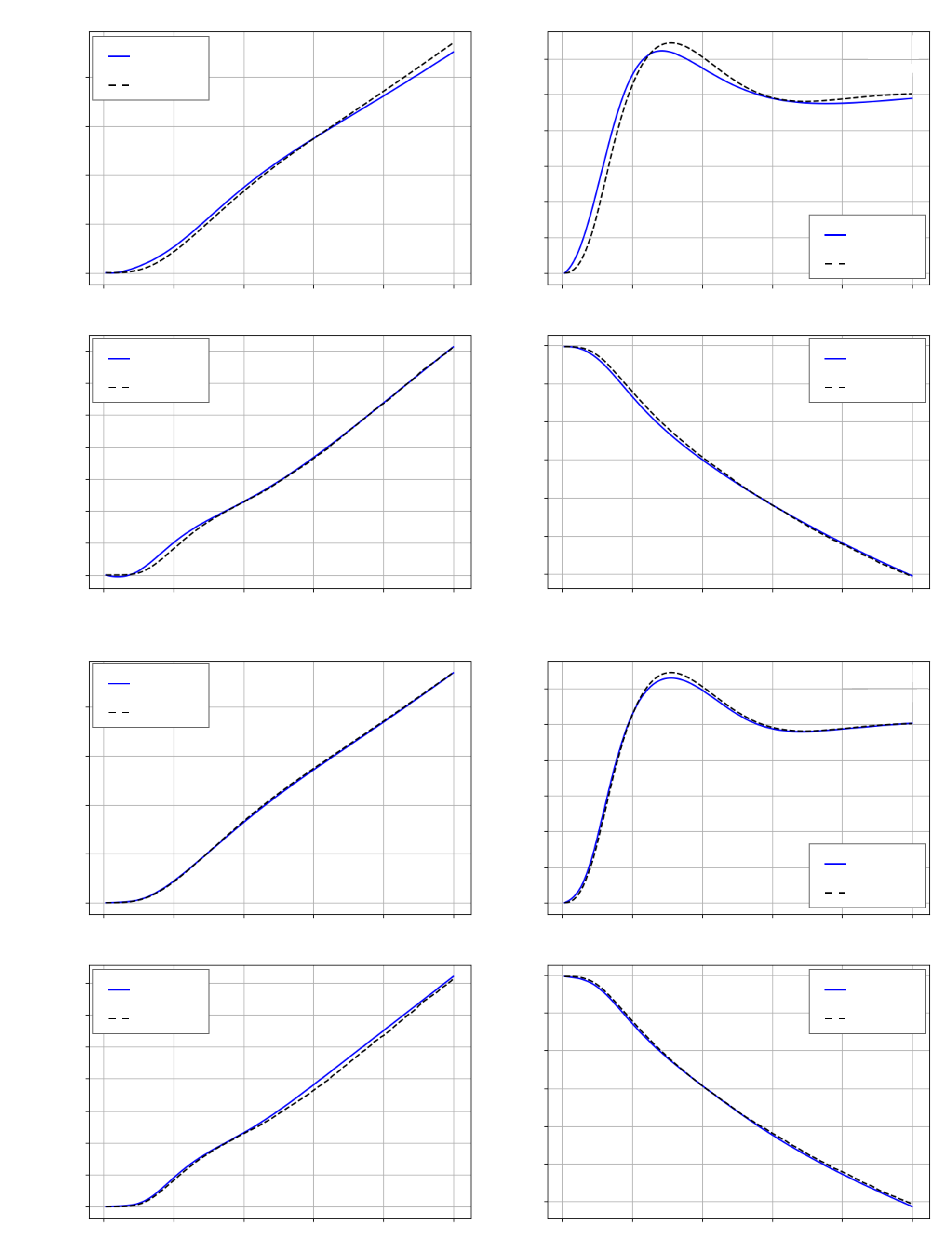
    \caption{Test trajectory reconstruction of the Couette flow in an Oldroyd fluid with GENERIC (a) and single generator bracket (b) formalisms. }
    \label{VE_Test}
\end{figure}

\begin{figure} 
    \centering
    \def\svgwidth{0.98\textwidth}
    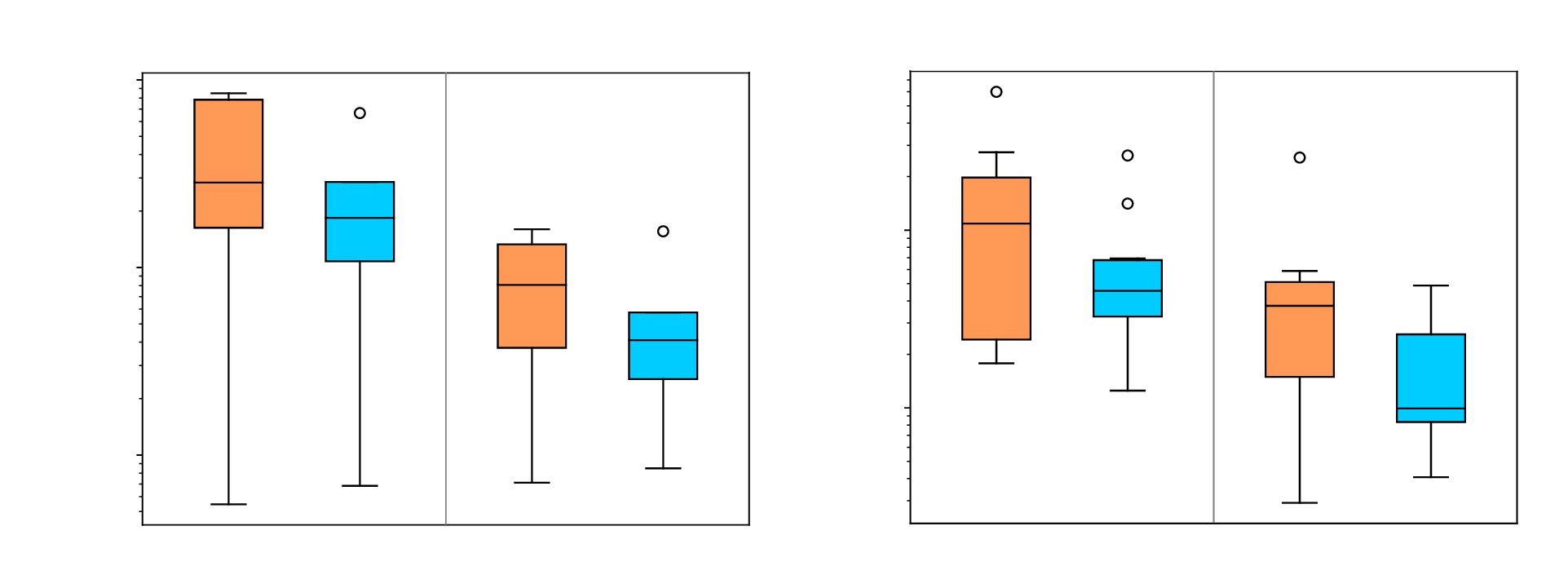
    \caption{Box plots for the data integration mean square error (MSE) (a), and the energy (Hamiltonian) mean square error (b), for the Couette flow in an Oldroyd fluid. }
    \label{VE_Error}
\end{figure}

The evolution of the state variables of the model, reconstructed by the GENERIC, and single generator, has been represented in Fig. \ref{VE_Test} (a), and Fig. \ref{VE_Test} (b), respectively. Even though the capacity of the net is lower than in the previous example, the obtained reconstruction shows at least one order of magnitude less error than in the previous system (Fig. \ref{VE_Error}). The results show a lower error for the reconstruction of the state variables in the single generator than in the GENERIC case. Likewise, a lower error is observed in the test than in the train, which can be attributed to specific trajectories close to the limits ($x=0$). The MSE error in the energy (internal energy of the system, $e$) shows better performance of the single generator than in the GENERIC formalism (Fig. \ref{VE_Error} (b)). 

\section{Discussion}

The inductive bias introduced when learning the different systems through the enforcement of well-known physics formalisms introduces interesting characteristics that are worth commenting on. We would like to point out that a comparison with a black box algorithm has not been made, since this has already been done with the GENERIC formalism in \cite{Hernandez2021}. In all the cases studied, GENERIC improved the results by about an order of magnitude. 

In the example of the thermo-elastic double pendulum, both formalisms show very similar system reconstruction errors, with less error obtained by GENERIC. On the contrary, in the Oldroyd-B fluid example, the single generator formalism reports the best results. These differences can be attributed to the differences between the two systems analyzed. Since the double pendulum is a chaotic system, the trajectories within the database can be much more different from each other than those observed in the Couette flow.   

By comparing the energy of the system, computed from the reconstructed variables of the system ($\mathcal{H}(\bs z_{n}^{\text{net}})$), with the ground truth energy ($\mathcal{H}(\bs z_{n}^{\text{GT}})$), less error has been observed in the single generator than in the GENERIC case, even without the conservation of energy being explicitly restricted, see Fig. \ref{DP_Error} (b). 

For comparison and application of these formalisms in learning the structure of a system through neural networks, it is important to understand the advantages and limitations they may present. The single generator structure, with fewer restrictions than GENERIC, has therefore a higher expressiveness, while the GENERIC structure, by the separation of the energy generators, is more representative, in the sense that it allows for an explicit imposition of the laws of thermodynamics.  However, this comes at the cost of including a new hyperparameter in the loss function. This hyperparameter can increase the weight to one part of the problem or another: reconstruction of the data, or imposition of thermodynamic correctness. In this sense, forcing a very strict imposition of thermodynamics can make it difficult to find the solution to the problem, which, in part, responds to the observed advantage of the single generator over GENERIC. On the contrary, the degeneracy conditions can be seen as additional information available to the network to find the solution space for the problem. 

\begin{figure} 
    \centering
    \def\svgwidth{0.98\textwidth}
    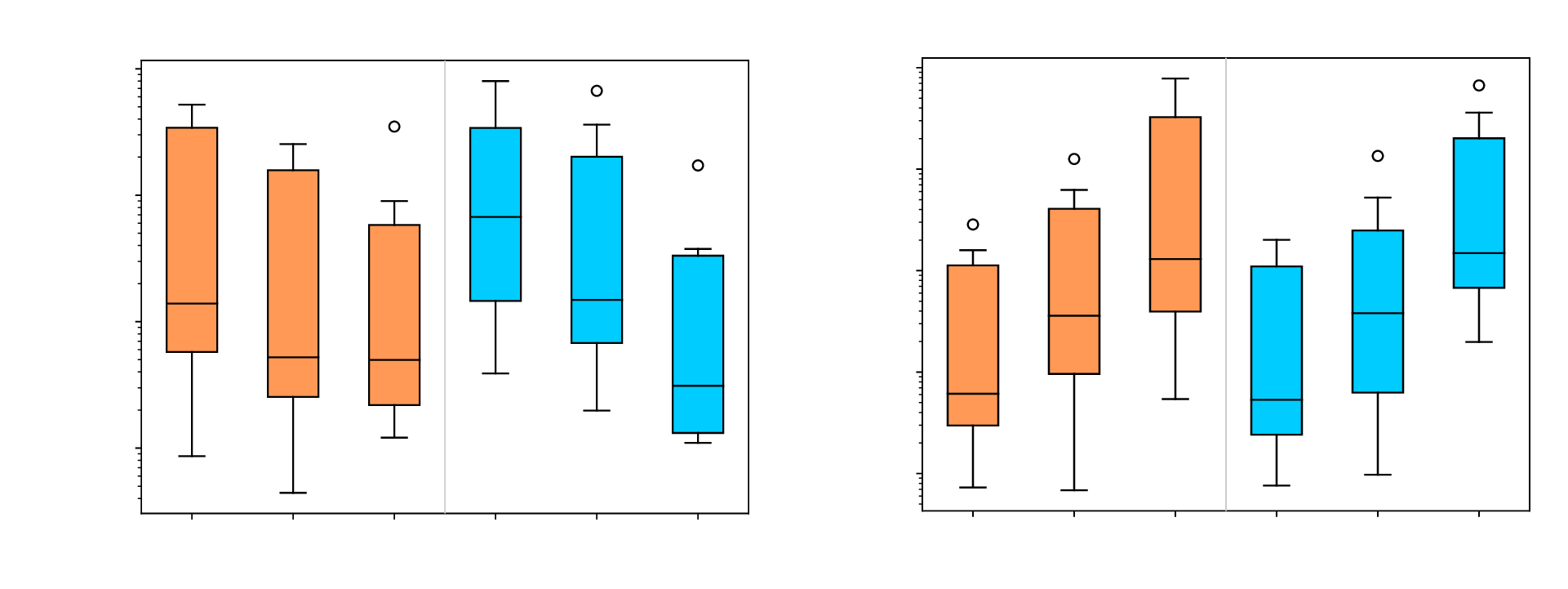
    \caption{Box plots for the test data integration mean square error (MSE) for the double thermo-elastic pendulum. (a) Increasing the number of snapshots (decrease in the step time) increases data error which is associated with the integration method. (b) Increasing the number of trajectories in the database improves the generalization of the network which highly reduces the error. }
    \label{DP_Ndata}
\end{figure}

\begin{figure} 
    \centering   
    \def\svgwidth{0.98\textwidth}
    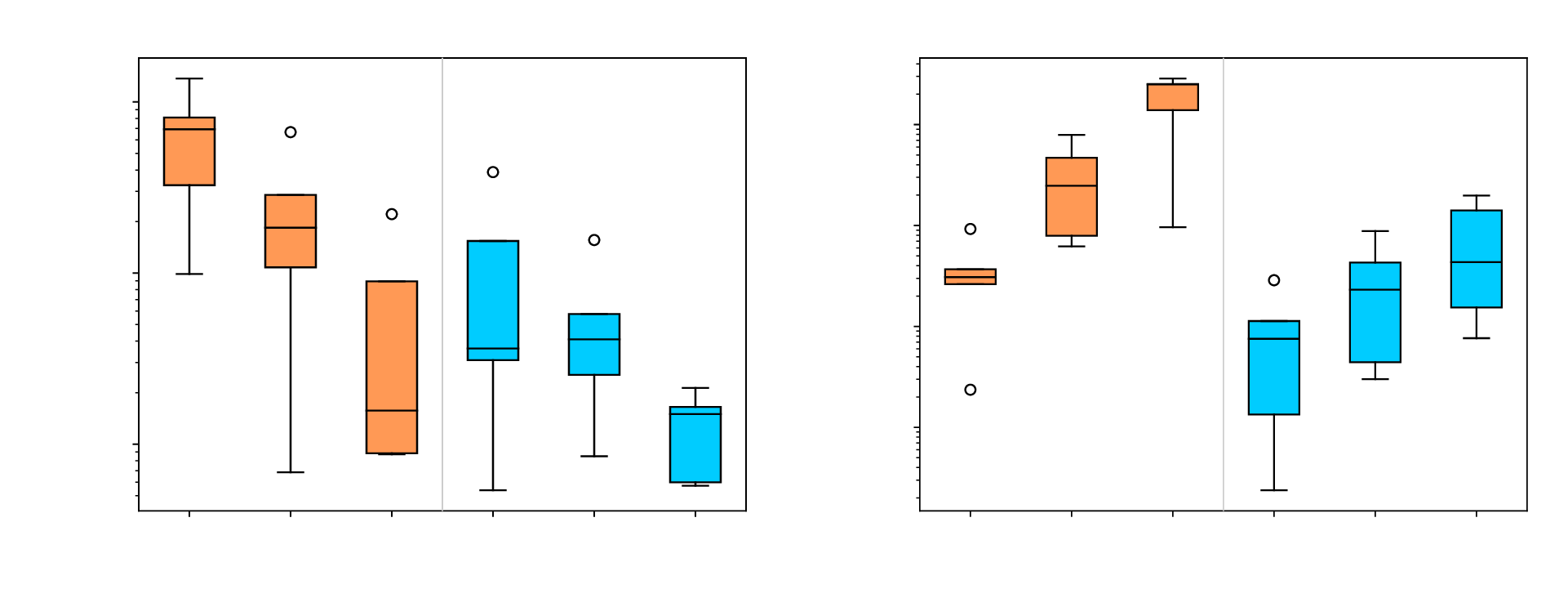
    \caption{Box plots for the test data integration mean square error (MSE) for the Couette flow in an Oldroyd-B fluid. (a) By increasing the number of snapshots ($300$, $200$, and $50$ snapshots), the error obtained increases. (b) By increasing the number of points ($400$, $200$, and $50$ discretized points), the error obtained significantly decreases.}
    \label{VE_Ndata}
\end{figure}

Due to the simple time integration scheme (forward Euler), small errors obtained in the integration of the variables cause a magnification of the successive error. Thus, by increasing the number of snapshots in the database, the error of the reconstruction of the state variables increases, see Fig. \ref{DP_Ndata} (a), and Fig. \ref{VE_Ndata} (a). Besides, by increasing the number of trajectories in the database, the generalization of the net is improved which decreases the error of the reconstruction of the state variables, see Fig. \ref{DP_Ndata} (b), and Fig. \ref{VE_Ndata} (b). Moreover, we observed that an increase in the number of train trajectories promotes thermodynamic consistency in GENERIC, see Appendix \hyperref[App_data]{B} and Fig. \ref{DP_LearnedEnergy}. 

For the case of the Couette flow in the Oldroyd-B fluid, the variation of the {\it Re} and {\it We} numbers have been studied. As {\it Re} and {\it We} change, the relative importance of the dissipative and reversible dynamics of the example is varied. It has been observed that the behavior of both formalisms differs, with GENERIC showing better behavior for weakly dissipative dynamics (Appendix \hyperref[App_data]{B}, Fig. \ref{VE_RexxWexx}). Thus, GENERIC seems to be more advisable with dynamics including low dissipative content, while the single generator scheme shows a clear advantage with higher dissipative dynamics.

The results obtained with both formalisms are dependent on the hyperparameters of the network. The value of the learning rate seems to be the most critical parameter for the GENERIC formalism. With values of $lr \ge 1e-3$, for the case of the double thermo-elastic pendulum, the network can stagnate in a local minimum for which the dissipative component is the trivial solution $\boldsymbol{M}=\bs 0$. For its part, the single generator seems to show a higher dependence on the network capacity and the training process, being unable to reconstruct some trajectories (test, but also train) with a low capacity, and a low number of training epochs. At the same time, its dependence on the data to generalize the solution means that, as the number of training trajectories in the database is reduced, the single generator error increases and vice versa. Thus, GENERIC shows more stable behavior and higher robustness to the conditions of the database and the hyperparameters of the network.

\section{Conclusion}

Both formalisms show high accuracy in the results, being coherent with the laws of thermodynamics. Even when considering a single generator, where the imposition of these is not explicit, the energy, computed by post-processing, shows a good degree of conservation. Although the single generator formalism indeed seems to yield better results in general terms, with lower computing costs (less training time), the stability of the predictions depends largely on the adequate adjustment of the network hyperparameters. Thus, the effect of the different hyperparameters, including learning rate, number of training epochs, and neurons in hidden layers, as well as differences in the database have been compared in the reconstruction of the system with both formalisms. In this sense, decreasing the capacity of the network (fewer units per hidden layer) limits the network's ability to generalize and prevents the reconstruction of some trajectories with the single generator formalism. Besides, reducing the learning rate to $lr=1e-4$ within the single generator paradigm requires increasing the number of training epochs to be able to reconstruct the solution. This is not observed in the GENERIC formalism, which, maybe supported by the degeneracy conditions, shows higher robustness in the reconstruction of the trajectories in every tested example. 

For its part, the structure in the GENERIC formalism is more representative since it separates the dynamics into two independent terms. This can be seen as an advantage but also can imply some limitations. As the energy generators were defined separately, the thermodynamic laws can be imposed explicitly by the degeneracy conditions. However, this implies the addition of an extra term on the loss function, which needs to be compensated with the data loss with an extra hyperparameter $\lambda_d$. An additional limitation of the GENERIC formalism is derived from the separation of energy generators. This separation into two independent terms can guide the learning process to a solution in which one of these energy potential generators is no longer relevant (trivial solution with $\boldsymbol{M}\nabla \mathcal{S}=\bs 0$) and the entire weight of the dynamics of the problem is learned through the other term of the formalism. 

Some of these conclusions depend on the adjustment of various hyperparameters of the network and the physics of the particular example studied. In this work, we have tried to simplify all this content to focus on a clear answer to the main question: single generator or GENERIC for learning physics? In this sense, there is no definitive winner with conclusive arguments. Both formalisms seem to be equivalent---something already demonstrated in the literature---. The addition of degeneracy conditions acts as an additional data term to obtain the solution space of the system. Moreover, the degeneracy conditions provide higher robustness to the model, allowing for a better generalization and lower errors with fewer data samples, in chaotic systems, and lower network capacity (number of neurons). For its part, the single generator formalism has shown to be capable of learning, to a certain extent, these implicit restrictions in the system, reporting generally less error in the system variables reconstruction, with less computational cost, but with a high dependence on the adjustment of the network's hyperparameters. Thus, as was exposed by B. Edwards et al. in their analysis of complex fluids, even though both formalisms can reconstruct the dynamics of different complex systems, the description of the energy through the double generator with the GENERIC structure (separated Hamiltonian and entropy) is more natural and reports some benefits in the description of dissipative dynamics \cite{Edwards1998, Edwards1998a}.

\section*{Aclnowledgements}

This material is also based upon work supported in part by the Army Research Laboratory and the Army Research Office under contract/grant number W911NF2210271.

This work was also supported by the Spanish Ministry of Science and Innovation, AEI/10.13039/501100011033, through Grants number PID2020-113463RB-C31 and TED2021-130105B-I00 and by the Ministry for Digital Transformation and the Civil Service, through the ENIA 2022 Chairs for the creation of university-industry chairs in AI, through Grant TSI-100930-2023-1.

This research is also part of the DesCartes programme and is supported by the National
Research Foundation, Prime Minister Office, Singapore under its Campus for Research
Excellence and Technological Enterprise (CREATE) programme.

The authors also acknowledge the support of ESI Group through the chairs at the
University of Zaragoza and at ENSAM Institute of Technology.

\appendix
\section*{Appendix A. Influence of hyperparameter values}\label{App_hyp}

The effect of different hyperparameters in the reconstruction of the system with the GENERIC and single generator formalisms has been analyzed. The single generator shows a high sensibility on the hyperparameters of the net as compared with GENERIC. Low capacity networks and insufficient training epochs make the single generator approach to be unable to reconstruct the state variables of the double thermo-elastic pendulum. For its part, the thermodynamic consistency of GENERIC shows a stronger dependence on the choice of an appropriate learning rate value. We detail the effect of these values in the following sections.

\subsection*{Training epochs}

\begin{figure} 
    \centering
    \def\svgwidth{0.65\textwidth}
    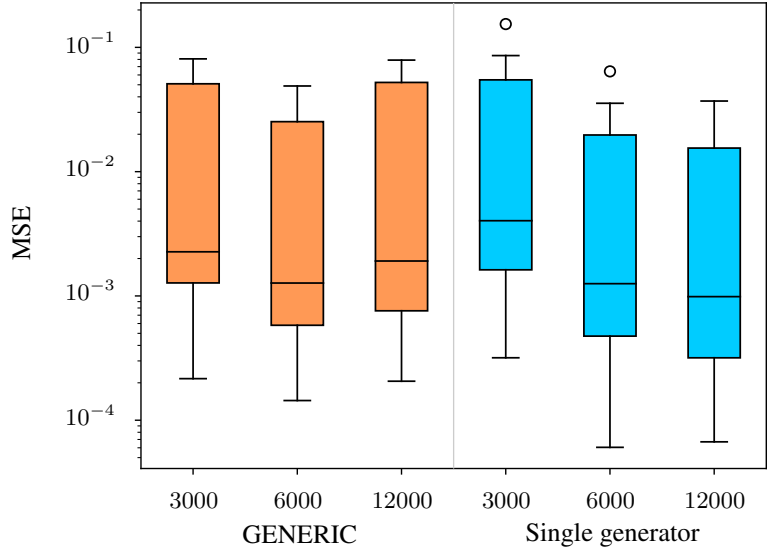
    \caption{Box plots for the data test integration mean square error (MSE). Effect of increasing the number of training epochs in GENERIC and single generator bracket formalisms.}
    \label{DP_epoch}
\end{figure}

Generally speaking, as the number of training epochs is increased, the reported error of the net is reduced. This tendency, on the limit, will contribute to the net overfitting, which reduces the generalization of the forecasts and increases the error in the test trajectories. We have compared the effect of the increase of the maximum training epochs in the double pendulum example, by training both formalisms with $N_{\text{epoch}}=3000$, $N_{\text{epoch}}=6000$, and $N_{\text{epoch}}=12000$. As observed in Fig. \ref{DP_epoch}, in general, for both formalisms, the error on the state variables reconstruction is reduced as the number of training epochs increases. However, for the maximum number of epochs considered in GENERIC, the error is increased. As observed in detail in the reconstruction of the model by Eq. (\ref{G_formalism}), the increase of the maximum epoch moves the solution to a local minimum (trivial solution with $\bs M=\bs 0$) which reduces the generalization of the problem and increases the test error. This seems to indicate that the system solution encounters a ``cliff'' point, and the use of a high learning rate causes it to systematically move away from the solution. If we look at Fig. \ref{DP_Loss}, we can see a first region where the term associated with the data in the loss function decreases continuously but fluctuates in the term associated with the degeneracy conditions, which denotes a complicated balance between these two contributions in the Loss function. This problem is solved by decreasing the learning rate to $lr=1e-4$. 

\subsection*{Learning rate}
\begin{figure} 
    \centering
    \def\svgwidth{0.65\textwidth}
    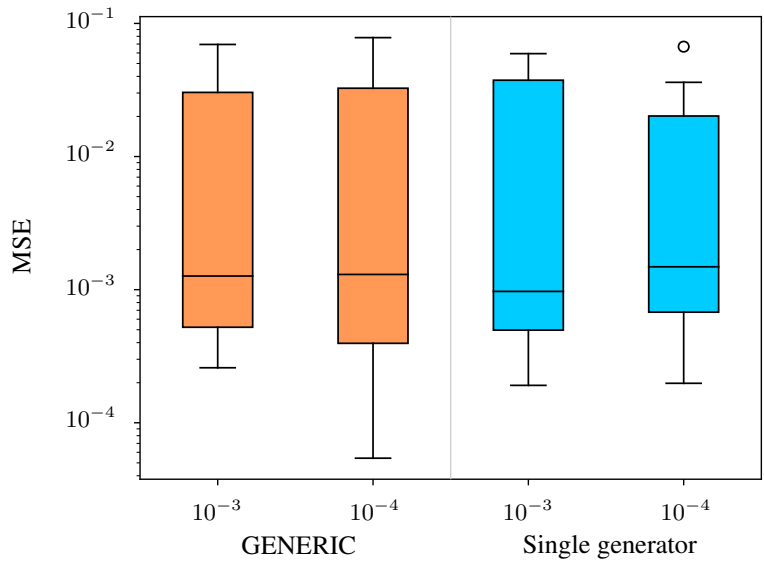
    \caption{Box plots for the data integration mean square error (MSE), with initial learning rates of $lr=1e-3$, and $lr=1e-4$ for the GENERIC and single generator bracket formalisms.}
    \label{DP_lrate}
\end{figure}

The effect of different learning rates has been studied in the reconstruction of the double pendulum system with the GENERIC and single generator formalisms (Fig. \ref{DP_lrate}). The error in the variables reconstruction represented shows a worsening in the error for both formalisms, much more significant in GENERIC. Then, the single generator with a learning rate of $lr=1e-3$ reports a lower error than GENERIC. Additionally, for the same learning rate value, GENERIC has reported falling into a local minimum (trivial solution $M=0$) that has no physical meaning which cannot be considered valid from the perspective of this work. However, for the different learning rates studied ($1e-2$, $1e-3$, $1e-4$, and $1e-5$), single generator only was capable of reconstructing the test trajectories for the represented learning rates and at the cost of an increase in the number of training epoch in case of $lr=1e-4$ (from $6000$ to $12000$). Thus, in terms of stability in the solution, the robustness of the GENERIC formalism against the single generator must also be assessed, the latter being much more susceptible to failure.

\subsection*{Number of neurons per hidden layer}

\begin{figure} 
    \centering
    \def\svgwidth{0.65\textwidth}
    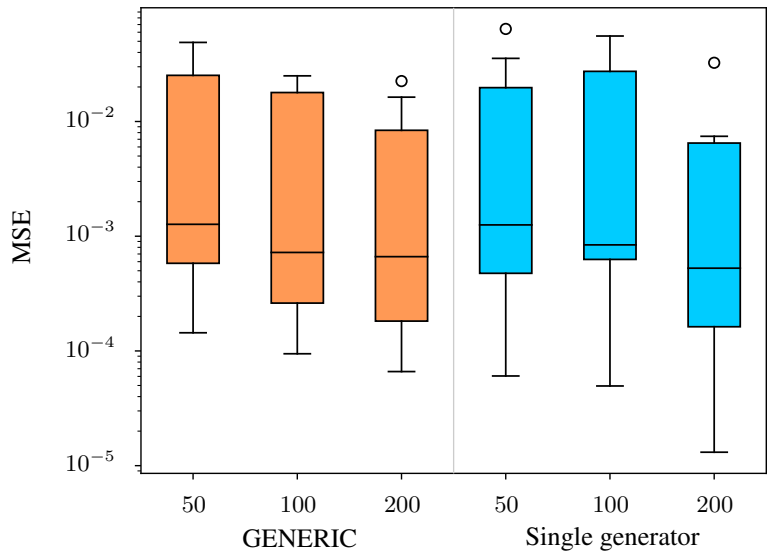
    \caption{Box plots for the data test integration mean square error (MSE) in the double thermo-elastic pendulum system. Effect of the net capacity, number of neurons on hidden layers, for GENERIC and single generator bracket formalisms. As the capacity of the net increases, the MSE on the reconstructed trajectories is reduced. }
    \label{DP_HiddenLayer}
\end{figure}

By increasing the number of neurons in hidden layers, both formalisms reduce the mean square error on the reconstructed system variables (Fig. \ref{DP_HiddenLayer}). However, a high dependency on the network capacity has been observed for the single generator formalism. While the GENERIC formalism can reconstruct the trajectories even with 20 neurons in the hidden layer, the single generator is not able to reconstruct some trajectories with less than 50 neurons. If the learning rate is decreased to 1e-4, the single generator reports an error in the reconstruction of the variables with less than 200 neurons in the hidden layer.

\section*{Appendix B. Influence of data in the learning process}\label{App_data}

\subsection*{Amount of data}

The training database plays a key role in the learning process of the system. The quantity and diversity of the data are crucial to allow adequate learning with sufficient generalization. Thus, we analyzed the effect of increasing the database in both formalisms and for both examples (double thermo-elastic pendulum and Couette flow in an Oldroyd-B fluid). 

\begin{figure} 
    \centering
    \def\svgwidth{0.98\textwidth}
    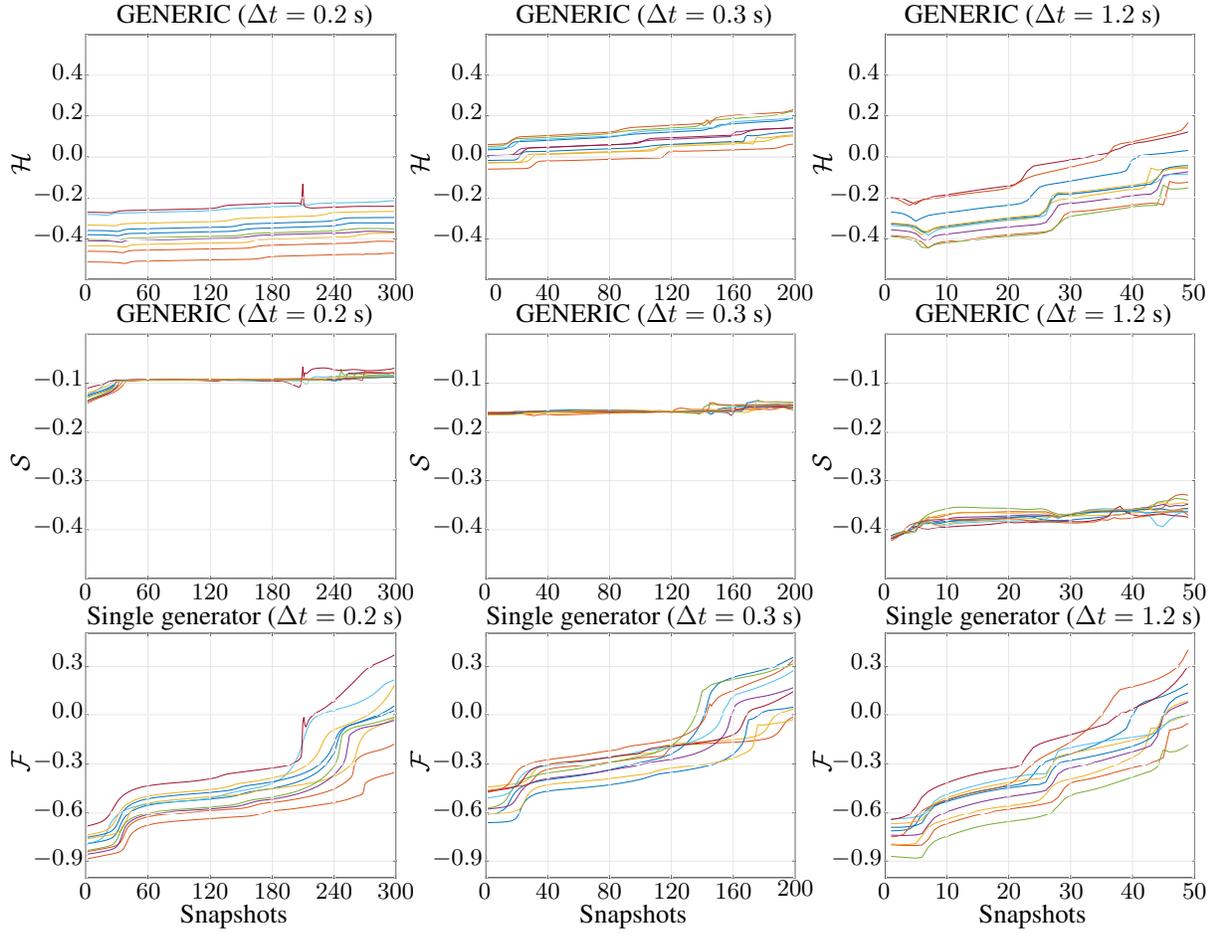
    \caption{Plot of the learned energy generators in GENERIC ($\mathcal{H}$, and $\mathcal{S}$) and single generator bracket ($\mathcal{F}$) for different runs of the double thermo-elastic pendulum. Increasing the number of snapshots improves thermodynamic consistency in GENERIC formalism. }
    \label{DP_LearnedEnergy}
\end{figure}

By increasing the number of snapshots (decreasing the time step increment) a data augmentation is obtained. Moreover, the increase in the number of snapshots increases the number of integration points in the reconstruction of the system. Thus, even though the increase of snapshots will decrease the error of the net at each step of integration compared with the case of reference, the effect of the accumulated errors at each integration step will increase the error obtained (Fig. \ref{DP_Ndata} (a), and Fig. \ref{VE_Ndata} (a)). However, in a detailed view of the learned system, the increase in the number of snapshots benefits the thermodynamic consistency of the learned system. Thus, the learned energy generators, the Hamiltonian, $\mathcal{H}$, and the Entropy, $\mathcal{S}$, in GENERIC, and, the free energy, $\mathcal{F}$ in the single generator, have been represented in Fig. \ref{DP_LearnedEnergy}. The energy conservation principle imposed in GENERIC implies that $\mathcal{\dot{H}}=0$, which is better  fulfilled for smaller time increments, as expected. As the number of snapshots increases, the results for the GENERIC formalism tend to converge to this imposed condition.

The double thermo-elastic is a chaotic system, thus increasing the number of trajectories in the database will introduce additional information on the solution space of the system. In the case of the Oldroyd-B fluid Couette flow, the increase in the trajectories is achieved by considering extra points in the discretization in the vertical direction of the model. In this sense, as the number of trajectories in the database increases, the error of the reconstruction of the variables of the system is reduced in both formalisms and for both analyzed systems (Fig. \ref{DP_Ndata} (b), and Fig. \ref{VE_Ndata} (b)). 

\subsection*{Influence of the physics}

\begin{figure} 
    \centering
    \def\svgwidth{0.98\textwidth}
    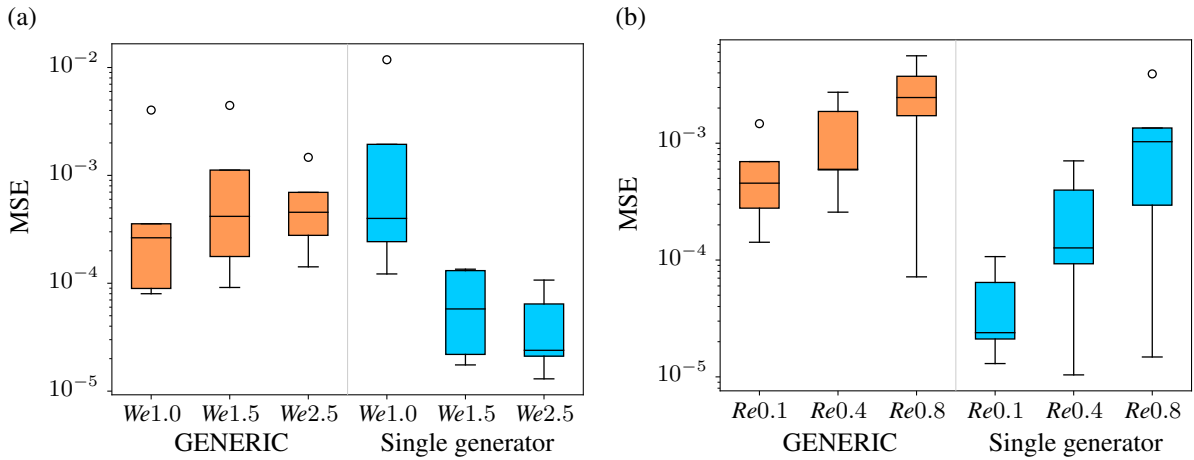
    \caption{Box plots for the data integration mean square error (MSE).  (a) Errors corresponding to $\textit{Re}=0.1$ and increasing values for $\textit{We}$ number ($\textit{We}=1.0$, $\textit{We}=1.5$, and $\textit{We}=2.5$). (b) Errors corresponding to $\textit{We}=2.5$ and increasing values for $\textit{Re}$ number ($\textit{Re}=0.1$, $\textit{Re}=0.4$, and $\textit{Re}=0.8$).}
    \label{VE_RexxWexx}
\end{figure}

Different Reynolds, {\it Re} and Weissenberg, {\it We} numbers have been used in the generation of Couette flow databases in the Oldroyd-B fluid, to analyze the contributions of dissipative and conservative effects in the training of the two formalisms. As {\it We} increases, the elasticity of the system increases, which in turn decreases the relative importance of viscous effects and the dissipative effects. On the contrary, as {\it Re} is increased, the internal energy of the system increases which can be associated with the increase in the Hamiltonian. The effect of the variation of the {\it We} number is different in both models. Increasing {\it We} in GENERIC increases the error, while a reduction in the error is observed in the single generator formalism (Fig. \ref{VE_RexxWexx} (a)). However, for the minimum {\it We} value studied,  GENERIC reports a lower error than the single generator. On the other hand, as {\it Re} is increased, the errors of both grow but not to the same extent, the effect on GENERIC being smaller, Fig. \ref{VE_RexxWexx} (b). The tendency of the errors for both formalisms seems to imply that a higher increase in the {\it Re} number will eventually cause GENERIC to equal or improve the results of the single generator.

Whether due to the dissipative effect reduction or an increase in the conservative effect, the relationship between the conservative and dissipative effect seems to play a key role in the differences in the behavior of both formalisms. As the system becomes more dissipative the single generator seems to report better results. On the contrary, as the system becomes conservatively dominated (reduced $\mathcal S/\mathcal H$) the GENERIC reports better results. This can be justified by the separation of both energy generators and the degeneracy conditions in GENERIC, which gives support to learning the dissipative dynamics of the system even when their contribution is reduced. Besides, given that in single generator the energy generators are not separated, a reduction in dynamics can be diluted and lose relevance compared to the conservative part, increasing the error to a greater extent than in GENERIC in systems with a low dissipative component. 

\bibliographystyle{unsrt}  

\end{document}

%% file: SPNN_fig.eps_tex
\begingroup%
  \makeatletter%
  \providecommand\color[2][]{%
    \errmessage{(Inkscape) Color is used for the text in Inkscape, but the package 'color.sty' is not loaded}%
    \renewcommand\color[2][]{}%
  }%
  \providecommand\transparent[1]{%
    \errmessage{(Inkscape) Transparency is used (non-zero) for the text in Inkscape, but the package 'transparent.sty' is not loaded}%
    \renewcommand\transparent[1]{}%
  }%
  \providecommand\rotatebox[2]{#2}%
  \newcommand*\fsize{\dimexpr\f@size pt\relax}%
  \newcommand*\lineheight[1]{\fontsize{\fsize}{#1\fsize}\selectfont}%
  \ifx\svgwidth\undefined%
    \setlength{\unitlength}{586.18272833bp}%
    \ifx\svgscale\undefined%
      \relax%
    \else%
      \setlength{\unitlength}{\unitlength * \real{\svgscale}}%
    \fi%
  \else%
    \setlength{\unitlength}{\svgwidth}%
  \fi%
  \global\let\svgwidth\undefined%
  \global\let\svgscale\undefined%
  \makeatother%
  \begin{picture}(1,0.56047078)%
    \lineheight{1}%
    \setlength\tabcolsep{0pt}%
    \put(0,0){\includegraphics[width=\unitlength]{SPNN_fig.eps}}%
    \put(0.12365312,0.16437401){\color[rgb]{0,0,0}\makebox(0,0)[t]{\lineheight{1.25}\smash{\begin{tabular}[t]{c}NN\end{tabular}}}}%
    \put(0.52557876,0.16437401){\color[rgb]{0,0,0}\makebox(0,0)[t]{\lineheight{1.25}\smash{\begin{tabular}[t]{c}GENERIC\end{tabular}}}}%
    \put(0.53410767,0.03041485){\color[rgb]{0,0,0}\makebox(0,0)[t]{\lineheight{1.25}\smash{\begin{tabular}[t]{c}Net update: $\phi \leftarrow \phi - \eta \frac{\partial \mathcal{L}}{\partial \phi}$\end{tabular}}}}%
    \put(0.01730846,0.18283083){\color[rgb]{0,0,0}\makebox(0,0)[lt]{\lineheight{1.25}\smash{\begin{tabular}[t]{l}$\boldsymbol{z}_t$\end{tabular}}}}%
    \put(0.61278155,0.18283083){\color[rgb]{0,0,0}\makebox(0,0)[t]{\lineheight{1.25}\smash{\begin{tabular}[t]{c}$\boldsymbol{z}_{t+1}^{\text{net}}$\end{tabular}}}}%
    \put(0.67936335,0.07977085){\color[rgb]{0,0,0}\makebox(0,0)[rt]{\lineheight{1.25}\smash{\begin{tabular}[t]{r}$\boldsymbol{z}_{t+1}^{\text{GT}}$\end{tabular}}}}%
    \put(0.51528548,0.0795797){\color[rgb]{0,0,0}\makebox(0,0)[rt]{\lineheight{1.25}\smash{\begin{tabular}[t]{r}$\Delta t$\end{tabular}}}}%
    \put(0.21743424,0.22276944){\color[rgb]{0,0,0}\makebox(0,0)[t]{\lineheight{1.25}\smash{\begin{tabular}[t]{c}$\boldsymbol{l}, \boldsymbol{m}$\end{tabular}}}}%
    \put(0.31396211,0.22155515){\color[rgb]{0,0,0}\makebox(0,0)[t]{\lineheight{1.25}\smash{\begin{tabular}[t]{c}$\boldsymbol{L}=\boldsymbol{l}-\boldsymbol{l}^T$\end{tabular}}}}%
    \put(0.31396212,0.19086778){\color[rgb]{0,0,0}\makebox(0,0)[t]{\lineheight{1.25}\smash{\begin{tabular}[t]{c}$\boldsymbol{M}=\boldsymbol{m}·\boldsymbol{m}^T$\end{tabular}}}}%
    \put(0.21743424,0.1433138){\color[rgb]{0,0,0}\makebox(0,0)[t]{\lineheight{1.25}\smash{\begin{tabular}[t]{c}$\mathcal{H}, \mathcal{S}$\end{tabular}}}}%
    \put(0.42243525,0.22276944){\color[rgb]{0,0,0}\makebox(0,0)[t]{\lineheight{1.25}\smash{\begin{tabular}[t]{c}$\boldsymbol{L}, \boldsymbol{M}$\end{tabular}}}}%
    \put(0.42243526,0.1433138){\color[rgb]{0,0,0}\makebox(0,0)[t]{\lineheight{1.25}\smash{\begin{tabular}[t]{c}$\nabla \mathcal{H}, \nabla \mathcal{S}$\end{tabular}}}}%
    \put(0.82681793,0.07981209){\color[rgb]{0,0,0}\makebox(0,0)[rt]{\lineheight{1.25}\smash{\begin{tabular}[t]{r}$\mathcal{L}^{\text{reg}}$\end{tabular}}}}%
    \put(0.67773367,0.26746802){\color[rgb]{0,0,0}\makebox(0,0)[t]{\lineheight{1.25}\smash{\begin{tabular}[t]{c}$\mathcal{L}^{\text{degen}}$\end{tabular}}}}%
    \put(0.76998042,0.1756321){\color[rgb]{0,0,0}\makebox(0,0)[t]{\lineheight{1.25}\smash{\begin{tabular}[t]{c}$\mathcal{L}^{\text{data}}$\end{tabular}}}}%
    \put(0.31460683,0.12602999){\color[rgb]{0,0,0}\makebox(0,0)[t]{\lineheight{1.25}\smash{\begin{tabular}[t]{c}Autograd\end{tabular}}}}%
    \put(0.68777231,0.16299908){\color[rgb]{0,0,0}\makebox(0,0)[t]{\lineheight{1.25}\smash{\begin{tabular}[t]{c}SSE\end{tabular}}}}%
    \put(0.937573,0.16728675){\color[rgb]{0,0,0}\makebox(0,0)[t]{\lineheight{1.25}\smash{\begin{tabular}[t]{c}Loss\end{tabular}}}}%
    \put(0.8339581,0.16327689){\color[rgb]{0,0,0}\makebox(0,0)[t]{\lineheight{1.25}\smash{\begin{tabular}[t]{c}+\end{tabular}}}}%
    \put(0.12365312,0.46522891){\color[rgb]{0,0,0}\makebox(0,0)[t]{\lineheight{1.25}\smash{\begin{tabular}[t]{c}NN\end{tabular}}}}%
    \put(0.52511773,0.47980034){\color[rgb]{0,0,0}\makebox(0,0)[t]{\lineheight{1.25}\smash{\begin{tabular}[t]{c}Single\\Generator\end{tabular}}}}%
    \put(0.53410741,0.33126946){\color[rgb]{0,0,0}\makebox(0,0)[t]{\lineheight{1.25}\smash{\begin{tabular}[t]{c}Net update: $\phi \leftarrow \phi - \eta \frac{\partial \mathcal{L}}{\partial \phi}$\end{tabular}}}}%
    \put(0.01730846,0.48368583){\color[rgb]{0,0,0}\makebox(0,0)[lt]{\lineheight{1.25}\smash{\begin{tabular}[t]{l}$\boldsymbol{z}_t$\end{tabular}}}}%
    \put(0.61278133,0.48368583){\color[rgb]{0,0,0}\makebox(0,0)[t]{\lineheight{1.25}\smash{\begin{tabular}[t]{c}$\boldsymbol{z}_{t+1}^{\text{net}}$\end{tabular}}}}%
    \put(0.67936335,0.38062552){\color[rgb]{0,0,0}\makebox(0,0)[rt]{\lineheight{1.25}\smash{\begin{tabular}[t]{r}$\boldsymbol{z}_{t+1}^{\text{GT}}$\end{tabular}}}}%
    \put(0.51784441,0.38043441){\color[rgb]{0,0,0}\makebox(0,0)[rt]{\lineheight{1.25}\smash{\begin{tabular}[t]{r}$\Delta t$\end{tabular}}}}%
    \put(0.21743424,0.52362452){\color[rgb]{0,0,0}\makebox(0,0)[t]{\lineheight{1.25}\smash{\begin{tabular}[t]{c}$\boldsymbol{l}, \boldsymbol{m}$\end{tabular}}}}%
    \put(0.31396211,0.52241034){\color[rgb]{0,0,0}\makebox(0,0)[t]{\lineheight{1.25}\smash{\begin{tabular}[t]{c}$\boldsymbol{L}=\boldsymbol{l}-\boldsymbol{l}^T$\end{tabular}}}}%
    \put(0.31396212,0.49172283){\color[rgb]{0,0,0}\makebox(0,0)[t]{\lineheight{1.25}\smash{\begin{tabular}[t]{c}$\boldsymbol{M}=\boldsymbol{m}·\boldsymbol{m}^T$\end{tabular}}}}%
    \put(0.21743423,0.44416874){\color[rgb]{0,0,0}\makebox(0,0)[t]{\lineheight{1.25}\smash{\begin{tabular}[t]{c}$\mathcal{F}$\end{tabular}}}}%
    \put(0.42243525,0.52362452){\color[rgb]{0,0,0}\makebox(0,0)[t]{\lineheight{1.25}\smash{\begin{tabular}[t]{c}$\boldsymbol{L}, \boldsymbol{M}$\end{tabular}}}}%
    \put(0.42243526,0.44416874){\color[rgb]{0,0,0}\makebox(0,0)[t]{\lineheight{1.25}\smash{\begin{tabular}[t]{c}$\nabla \mathcal{F}$\end{tabular}}}}%
    \put(0.82681793,0.38066681){\color[rgb]{0,0,0}\makebox(0,0)[rt]{\lineheight{1.25}\smash{\begin{tabular}[t]{r}$\mathcal{L}^{\text{reg}}$\end{tabular}}}}%
    \put(0.76998042,0.476487){\color[rgb]{0,0,0}\makebox(0,0)[t]{\lineheight{1.25}\smash{\begin{tabular}[t]{c}$\mathcal{L}^{\text{data}}$\end{tabular}}}}%
    \put(0.31460683,0.42688482){\color[rgb]{0,0,0}\makebox(0,0)[t]{\lineheight{1.25}\smash{\begin{tabular}[t]{c}Autograd\end{tabular}}}}%
    \put(0.68777231,0.46385405){\color[rgb]{0,0,0}\makebox(0,0)[t]{\lineheight{1.25}\smash{\begin{tabular}[t]{c}SSE\end{tabular}}}}%
    \put(0.937573,0.46814169){\color[rgb]{0,0,0}\makebox(0,0)[t]{\lineheight{1.25}\smash{\begin{tabular}[t]{c}Loss\end{tabular}}}}%
    \put(0.8339581,0.46413172){\color[rgb]{0,0,0}\makebox(0,0)[t]{\lineheight{1.25}\smash{\begin{tabular}[t]{c}+\end{tabular}}}}%
    \put(-0.00040983,0.54706389){\color[rgb]{0,0,0}\makebox(0,0)[lt]{\lineheight{1.25}\smash{\begin{tabular}[t]{l}(a)\end{tabular}}}}%
    \put(-0.00040983,0.28972359){\color[rgb]{0,0,0}\makebox(0,0)[lt]{\lineheight{1.25}\smash{\begin{tabular}[t]{l}(b)\end{tabular}}}}%
  \end{picture}%
\endgroup%

%% file: DP_Losses.eps_tex
\begingroup%
  \makeatletter%
  \providecommand\color[2][]{%
    \errmessage{(Inkscape) Color is used for the text in Inkscape, but the package 'color.sty' is not loaded}%
    \renewcommand\color[2][]{}%
  }%
  \providecommand\transparent[1]{%
    \errmessage{(Inkscape) Transparency is used (non-zero) for the text in Inkscape, but the package 'transparent.sty' is not loaded}%
    \renewcommand\transparent[1]{}%
  }%
  \providecommand\rotatebox[2]{#2}%
  \newcommand*\fsize{\dimexpr\f@size pt\relax}%
  \newcommand*\lineheight[1]{\fontsize{\fsize}{#1\fsize}\selectfont}%
  \ifx\svgwidth\undefined%
    \setlength{\unitlength}{990.71380135bp}%
    \ifx\svgscale\undefined%
      \relax%
    \else%
      \setlength{\unitlength}{\unitlength * \real{\svgscale}}%
    \fi%
  \else%
    \setlength{\unitlength}{\svgwidth}%
  \fi%
  \global\let\svgwidth\undefined%
  \global\let\svgscale\undefined%
  \makeatother%
  \begin{picture}(1,0.32952644)%
    \lineheight{1}%
    \setlength\tabcolsep{0pt}%
    \put(0,0){\includegraphics[width=\unitlength]{DP_Losses.eps}}%
    \put(0.48731965,0.31069335){\color[rgb]{0,0,0}\makebox(0,0)[lt]{\lineheight{1.25}\smash{\begin{tabular}[t]{l}(b)\end{tabular}}}}%
    \put(0.83096941,0.24617811){\makebox(0,0)[lt]{\lineheight{1.25}\smash{\begin{tabular}[t]{l}\footnotesize{$\mathcal{L}^{\text{total}}$}\end{tabular}}}}%
    \put(0.83096941,0.28242723){\makebox(0,0)[lt]{\lineheight{1.25}\smash{\begin{tabular}[t]{l}\footnotesize{$\mathcal{L}^{\text{data}}$}\end{tabular}}}}%
    \put(0.83096941,0.26430267){\makebox(0,0)[lt]{\lineheight{1.25}\smash{\begin{tabular}[t]{l}\footnotesize{$\mathcal{L}^{\text{degen}}$}\end{tabular}}}}%
    \put(0.35670544,0.24457537){\makebox(0,0)[lt]{\lineheight{1.25}\smash{\begin{tabular}[t]{l}\footnotesize{$\mathcal{L}^{\text{total}}$}\end{tabular}}}}%
    \put(0.35670544,0.28082448){\makebox(0,0)[lt]{\lineheight{1.25}\smash{\begin{tabular}[t]{l}\footnotesize{$\mathcal{L}^{\text{data}}$}\end{tabular}}}}%
    \put(0.35670544,0.26269993){\makebox(0,0)[lt]{\lineheight{1.25}\smash{\begin{tabular}[t]{l}\footnotesize{$\mathcal{L}^{\text{degen}}$}\end{tabular}}}}%
    \put(0.01422271,0.31069335){\color[rgb]{0,0,0}\makebox(0,0)[lt]{\lineheight{1.25}\smash{\begin{tabular}[t]{l}(a)\end{tabular}}}}%
    \put(0.02025733,0.17189504){\rotatebox{90}{\makebox(0,0)[t]{\lineheight{1.25}\smash{\begin{tabular}[t]{c}Loss\end{tabular}}}}}%
    \put(0.25938976,0.001){\makebox(0,0)[t]{\lineheight{1.25}\smash{\begin{tabular}[t]{c}Epoch\end{tabular}}}}%
    \put(0.49254685,0.17189507){\rotatebox{90}{\makebox(0,0)[t]{\lineheight{1.25}\smash{\begin{tabular}[t]{c}Loss\end{tabular}}}}}%
    \put(0.73167922,0.001){\makebox(0,0)[t]{\lineheight{1.25}\smash{\begin{tabular}[t]{c}Epoch\end{tabular}}}}%
    \put(0.56815829,0.02514894){\makebox(0,0)[t]{\lineheight{1.25}\smash{\begin{tabular}[t]{c}\footnotesize{$0$}\end{tabular}}}}%
    \put(0.52406549,0.276526){\makebox(0,0)[t]{\lineheight{1.25}\smash{\begin{tabular}[t]{c}\footnotesize{$10^{3}$}\end{tabular}}}}%
    \put(0.52406549,0.22371404){\makebox(0,0)[t]{\lineheight{1.25}\smash{\begin{tabular}[t]{c}\footnotesize{$10^{1}$}\end{tabular}}}}%
    \put(0.52406549,0.17285035){\makebox(0,0)[t]{\lineheight{1.25}\smash{\begin{tabular}[t]{c}\footnotesize{$10^{-1}$}\end{tabular}}}}%
    \put(0.52406549,0.12078785){\makebox(0,0)[t]{\lineheight{1.25}\smash{\begin{tabular}[t]{c}\footnotesize{$10^{-3}$}\end{tabular}}}}%
    \put(0.52406549,0.06855599){\makebox(0,0)[t]{\lineheight{1.25}\smash{\begin{tabular}[t]{c}\footnotesize{$10^{-5}$}\end{tabular}}}}%
    \put(0.62268971,0.02514894){\makebox(0,0)[t]{\lineheight{1.25}\smash{\begin{tabular}[t]{c}\footnotesize{$2000$}\end{tabular}}}}%
    \put(0.67741605,0.02514894){\makebox(0,0)[t]{\lineheight{1.25}\smash{\begin{tabular}[t]{c}\footnotesize{$4000$}\end{tabular}}}}%
    \put(0.73214226,0.02514894){\makebox(0,0)[t]{\lineheight{1.25}\smash{\begin{tabular}[t]{c}\footnotesize{$6000$}\end{tabular}}}}%
    \put(0.7868686,0.02514894){\makebox(0,0)[t]{\lineheight{1.25}\smash{\begin{tabular}[t]{c}\footnotesize{$8000$}\end{tabular}}}}%
    \put(0.84143538,0.02514894){\makebox(0,0)[t]{\lineheight{1.25}\smash{\begin{tabular}[t]{c}\footnotesize{$10000$}\end{tabular}}}}%
    \put(0.89576657,0.02514894){\makebox(0,0)[t]{\lineheight{1.25}\smash{\begin{tabular}[t]{c}\footnotesize{$12000$}\end{tabular}}}}%
    \put(0.05177597,0.26441349){\makebox(0,0)[t]{\lineheight{1.25}\smash{\begin{tabular}[t]{c}\footnotesize{$10^{3}$}\end{tabular}}}}%
    \put(0.05177597,0.21614371){\makebox(0,0)[t]{\lineheight{1.25}\smash{\begin{tabular}[t]{c}\footnotesize{$10^{1}$}\end{tabular}}}}%
    \put(0.05177597,0.16679408){\makebox(0,0)[t]{\lineheight{1.25}\smash{\begin{tabular}[t]{c}\footnotesize{$10^{-1}$}\end{tabular}}}}%
    \put(0.05177597,0.11927379){\makebox(0,0)[t]{\lineheight{1.25}\smash{\begin{tabular}[t]{c}\footnotesize{$10^{-3}$}\end{tabular}}}}%
    \put(0.05177597,0.07007005){\makebox(0,0)[t]{\lineheight{1.25}\smash{\begin{tabular}[t]{c}\footnotesize{$10^{-5}$}\end{tabular}}}}%
    \put(0.09586883,0.02514894){\makebox(0,0)[t]{\lineheight{1.25}\smash{\begin{tabular}[t]{c}\footnotesize{$0$}\end{tabular}}}}%
    \put(0.15040025,0.02514894){\makebox(0,0)[t]{\lineheight{1.25}\smash{\begin{tabular}[t]{c}\footnotesize{$2000$}\end{tabular}}}}%
    \put(0.20512659,0.02514894){\makebox(0,0)[t]{\lineheight{1.25}\smash{\begin{tabular}[t]{c}\footnotesize{$4000$}\end{tabular}}}}%
    \put(0.2598528,0.02514894){\makebox(0,0)[t]{\lineheight{1.25}\smash{\begin{tabular}[t]{c}\footnotesize{$6000$}\end{tabular}}}}%
    \put(0.31457914,0.02514894){\makebox(0,0)[t]{\lineheight{1.25}\smash{\begin{tabular}[t]{c}\footnotesize{$8000$}\end{tabular}}}}%
    \put(0.36914598,0.02514894){\makebox(0,0)[t]{\lineheight{1.25}\smash{\begin{tabular}[t]{c}\footnotesize{$10000$}\end{tabular}}}}%
    \put(0.42347717,0.02514894){\makebox(0,0)[t]{\lineheight{1.25}\smash{\begin{tabular}[t]{c}\footnotesize{$12000$}\end{tabular}}}}%
  \end{picture}%
\endgroup%

%% file: DP_variablesGeneric.eps_tex
\begingroup%
  \makeatletter%
  \providecommand\color[2][]{%
    \errmessage{(Inkscape) Color is used for the text in Inkscape, but the package 'color.sty' is not loaded}%
    \renewcommand\color[2][]{}%
  }%
  \providecommand\transparent[1]{%
    \errmessage{(Inkscape) Transparency is used (non-zero) for the text in Inkscape, but the package 'transparent.sty' is not loaded}%
    \renewcommand\transparent[1]{}%
  }%
  \providecommand\rotatebox[2]{#2}%
  \newcommand*\fsize{\dimexpr\f@size pt\relax}%
  \newcommand*\lineheight[1]{\fontsize{\fsize}{#1\fsize}\selectfont}%
  \ifx\svgwidth\undefined%
    \setlength{\unitlength}{949.02808665bp}%
    \ifx\svgscale\undefined%
      \relax%
    \else%
      \setlength{\unitlength}{\unitlength * \real{\svgscale}}%
    \fi%
  \else%
    \setlength{\unitlength}{\svgwidth}%
  \fi%
  \global\let\svgwidth\undefined%
  \global\let\svgscale\undefined%
  \makeatother%
  \begin{picture}(1,0.9501938)%
    \lineheight{1}%
    \setlength\tabcolsep{0pt}%
    \put(0,0){\includegraphics[width=\unitlength]{DP_variablesGeneric.eps}}%
    \put(0.61341307,0.89914536){\makebox(0,0)[lt]{\lineheight{1.25}\smash{\begin{tabular}[t]{l}\footnotesize{Net (X)}\end{tabular}}}}%
    \put(0.61341307,0.88268566){\makebox(0,0)[lt]{\lineheight{1.25}\smash{\begin{tabular}[t]{l}\footnotesize{Net (Y)}\end{tabular}}}}%
    \put(0.61341307,0.86478805){\makebox(0,0)[lt]{\lineheight{1.25}\smash{\begin{tabular}[t]{l}\footnotesize{GT}\end{tabular}}}}%
    \put(0.90371795,0.59035952){\makebox(0,0)[lt]{\lineheight{1.25}\smash{\begin{tabular}[t]{l}\footnotesize{Net (X)}\end{tabular}}}}%
    \put(0.90371795,0.57389982){\makebox(0,0)[lt]{\lineheight{1.25}\smash{\begin{tabular}[t]{l}\footnotesize{Net (Y)}\end{tabular}}}}%
    \put(0.90371795,0.55600221){\makebox(0,0)[lt]{\lineheight{1.25}\smash{\begin{tabular}[t]{l}\footnotesize{GT}\end{tabular}}}}%
    \put(0.42187822,0.59035952){\makebox(0,0)[lt]{\lineheight{1.25}\smash{\begin{tabular}[t]{l}\footnotesize{Net (X)}\end{tabular}}}}%
    \put(0.42187822,0.57389982){\makebox(0,0)[lt]{\lineheight{1.25}\smash{\begin{tabular}[t]{l}\footnotesize{Net (Y)}\end{tabular}}}}%
    \put(0.42187822,0.55600221){\makebox(0,0)[lt]{\lineheight{1.25}\smash{\begin{tabular}[t]{l}\footnotesize{GT}\end{tabular}}}}%
    \put(0.42131362,0.89914536){\makebox(0,0)[lt]{\lineheight{1.25}\smash{\begin{tabular}[t]{l}\footnotesize{Net (X)}\end{tabular}}}}%
    \put(0.42131362,0.88268566){\makebox(0,0)[lt]{\lineheight{1.25}\smash{\begin{tabular}[t]{l}\footnotesize{Net (Y)}\end{tabular}}}}%
    \put(0.42131362,0.86478805){\makebox(0,0)[lt]{\lineheight{1.25}\smash{\begin{tabular}[t]{l}\footnotesize{GT}\end{tabular}}}}%
    \put(0.02881166,0.17445107){\color[rgb]{0,0,0}\rotatebox{90}{\makebox(0,0)[t]{\lineheight{1.25}\smash{\begin{tabular}[t]{c}$s_1$ [J/K]\end{tabular}}}}}%
    \put(0.62387829,0.27782859){\makebox(0,0)[lt]{\lineheight{1.25}\smash{\begin{tabular}[t]{l}\footnotesize{Net}\end{tabular}}}}%
    \put(0.62584911,0.25960515){\makebox(0,0)[lt]{\lineheight{1.25}\smash{\begin{tabular}[t]{l}\footnotesize{GT}\end{tabular}}}}%
    \put(0.10577662,0.64299827){\makebox(0,0)[t]{\lineheight{1.25}\smash{\begin{tabular}[t]{c}\footnotesize{$0$}\end{tabular}}}}%
    \put(0.41149333,0.64299827){\makebox(0,0)[t]{\lineheight{1.25}\smash{\begin{tabular}[t]{c}\footnotesize{$50$}\end{tabular}}}}%
    \put(0.47190892,0.64299827){\makebox(0,0)[t]{\lineheight{1.25}\smash{\begin{tabular}[t]{c}\footnotesize{$60$}\end{tabular}}}}%
    \put(0.16741589,0.64299827){\makebox(0,0)[t]{\lineheight{1.25}\smash{\begin{tabular}[t]{c}\footnotesize{$10$}\end{tabular}}}}%
    \put(0.28787416,0.64299827){\makebox(0,0)[t]{\lineheight{1.25}\smash{\begin{tabular}[t]{c}\footnotesize{$30$}\end{tabular}}}}%
    \put(0.3491475,0.64299827){\makebox(0,0)[t]{\lineheight{1.25}\smash{\begin{tabular}[t]{c}\footnotesize{$40$}\end{tabular}}}}%
    \put(0.22846545,0.64299827){\makebox(0,0)[t]{\lineheight{1.25}\smash{\begin{tabular}[t]{c}\footnotesize{$20$}\end{tabular}}}}%
    \put(0.58720038,0.64299827){\makebox(0,0)[t]{\lineheight{1.25}\smash{\begin{tabular}[t]{c}\footnotesize{$0$}\end{tabular}}}}%
    \put(0.89291713,0.64299827){\makebox(0,0)[t]{\lineheight{1.25}\smash{\begin{tabular}[t]{c}\footnotesize{$50$}\end{tabular}}}}%
    \put(0.95333272,0.64299827){\makebox(0,0)[t]{\lineheight{1.25}\smash{\begin{tabular}[t]{c}\footnotesize{$60$}\end{tabular}}}}%
    \put(0.64883969,0.64299827){\makebox(0,0)[t]{\lineheight{1.25}\smash{\begin{tabular}[t]{c}\footnotesize{$10$}\end{tabular}}}}%
    \put(0.76929796,0.64299827){\makebox(0,0)[t]{\lineheight{1.25}\smash{\begin{tabular}[t]{c}\footnotesize{$30$}\end{tabular}}}}%
    \put(0.8305713,0.64299827){\makebox(0,0)[t]{\lineheight{1.25}\smash{\begin{tabular}[t]{c}\footnotesize{$40$}\end{tabular}}}}%
    \put(0.70988925,0.64299827){\makebox(0,0)[t]{\lineheight{1.25}\smash{\begin{tabular}[t]{c}\footnotesize{$20$}\end{tabular}}}}%
    \put(0.28763205,0.6257972){\color[rgb]{0,0,0}\makebox(0,0)[t]{\lineheight{1.25}\smash{\begin{tabular}[t]{c}\footnotesize{time [s]}\end{tabular}}}}%
    \put(0.76996201,0.6257972){\color[rgb]{0,0,0}\makebox(0,0)[t]{\lineheight{1.25}\smash{\begin{tabular}[t]{c}\footnotesize{time [s]}\end{tabular}}}}%
    \put(0.10577667,0.33406282){\makebox(0,0)[t]{\lineheight{1.25}\smash{\begin{tabular}[t]{c}\footnotesize{$0$}\end{tabular}}}}%
    \put(0.41149333,0.33406282){\makebox(0,0)[t]{\lineheight{1.25}\smash{\begin{tabular}[t]{c}\footnotesize{$50$}\end{tabular}}}}%
    \put(0.47190892,0.33406282){\makebox(0,0)[t]{\lineheight{1.25}\smash{\begin{tabular}[t]{c}\footnotesize{$60$}\end{tabular}}}}%
    \put(0.16741589,0.33406282){\makebox(0,0)[t]{\lineheight{1.25}\smash{\begin{tabular}[t]{c}\footnotesize{$10$}\end{tabular}}}}%
    \put(0.28787416,0.33406282){\makebox(0,0)[t]{\lineheight{1.25}\smash{\begin{tabular}[t]{c}\footnotesize{$30$}\end{tabular}}}}%
    \put(0.3491475,0.33406282){\makebox(0,0)[t]{\lineheight{1.25}\smash{\begin{tabular}[t]{c}\footnotesize{$40$}\end{tabular}}}}%
    \put(0.22846545,0.33406282){\makebox(0,0)[t]{\lineheight{1.25}\smash{\begin{tabular}[t]{c}\footnotesize{$20$}\end{tabular}}}}%
    \put(0.58720038,0.33406282){\makebox(0,0)[t]{\lineheight{1.25}\smash{\begin{tabular}[t]{c}\footnotesize{$0$}\end{tabular}}}}%
    \put(0.89291722,0.33406282){\makebox(0,0)[t]{\lineheight{1.25}\smash{\begin{tabular}[t]{c}\footnotesize{$50$}\end{tabular}}}}%
    \put(0.95333272,0.33406282){\makebox(0,0)[t]{\lineheight{1.25}\smash{\begin{tabular}[t]{c}\footnotesize{$60$}\end{tabular}}}}%
    \put(0.64883969,0.33406282){\makebox(0,0)[t]{\lineheight{1.25}\smash{\begin{tabular}[t]{c}\footnotesize{$10$}\end{tabular}}}}%
    \put(0.76929805,0.33406282){\makebox(0,0)[t]{\lineheight{1.25}\smash{\begin{tabular}[t]{c}\footnotesize{$30$}\end{tabular}}}}%
    \put(0.8305713,0.33406282){\makebox(0,0)[t]{\lineheight{1.25}\smash{\begin{tabular}[t]{c}\footnotesize{$40$}\end{tabular}}}}%
    \put(0.70988925,0.33406282){\makebox(0,0)[t]{\lineheight{1.25}\smash{\begin{tabular}[t]{c}\footnotesize{$20$}\end{tabular}}}}%
    \put(0.28763205,0.31686175){\color[rgb]{0,0,0}\makebox(0,0)[t]{\lineheight{1.25}\smash{\begin{tabular}[t]{c}\footnotesize{time [s]}\end{tabular}}}}%
    \put(0.76996201,0.31686175){\color[rgb]{0,0,0}\makebox(0,0)[t]{\lineheight{1.25}\smash{\begin{tabular}[t]{c}\footnotesize{time [s]}\end{tabular}}}}%
    \put(0.10514389,0.02553446){\makebox(0,0)[t]{\lineheight{1.25}\smash{\begin{tabular}[t]{c}\footnotesize{$0$}\end{tabular}}}}%
    \put(0.41086064,0.02553446){\makebox(0,0)[t]{\lineheight{1.25}\smash{\begin{tabular}[t]{c}\footnotesize{$50$}\end{tabular}}}}%
    \put(0.47127623,0.02553446){\makebox(0,0)[t]{\lineheight{1.25}\smash{\begin{tabular}[t]{c}\footnotesize{$60$}\end{tabular}}}}%
    \put(0.16678311,0.02553446){\makebox(0,0)[t]{\lineheight{1.25}\smash{\begin{tabular}[t]{c}\footnotesize{$10$}\end{tabular}}}}%
    \put(0.28724147,0.02553446){\makebox(0,0)[t]{\lineheight{1.25}\smash{\begin{tabular}[t]{c}\footnotesize{$30$}\end{tabular}}}}%
    \put(0.3485148,0.02553446){\makebox(0,0)[t]{\lineheight{1.25}\smash{\begin{tabular}[t]{c}\footnotesize{$40$}\end{tabular}}}}%
    \put(0.22783276,0.02553446){\makebox(0,0)[t]{\lineheight{1.25}\smash{\begin{tabular}[t]{c}\footnotesize{$20$}\end{tabular}}}}%
    \put(0.58656769,0.02553446){\makebox(0,0)[t]{\lineheight{1.25}\smash{\begin{tabular}[t]{c}\footnotesize{$0$}\end{tabular}}}}%
    \put(0.89228453,0.02553446){\makebox(0,0)[t]{\lineheight{1.25}\smash{\begin{tabular}[t]{c}\footnotesize{$50$}\end{tabular}}}}%
    \put(0.95269994,0.02553446){\makebox(0,0)[t]{\lineheight{1.25}\smash{\begin{tabular}[t]{c}\footnotesize{$60$}\end{tabular}}}}%
    \put(0.648207,0.02553446){\makebox(0,0)[t]{\lineheight{1.25}\smash{\begin{tabular}[t]{c}\footnotesize{$10$}\end{tabular}}}}%
    \put(0.76866536,0.02553446){\makebox(0,0)[t]{\lineheight{1.25}\smash{\begin{tabular}[t]{c}\footnotesize{$30$}\end{tabular}}}}%
    \put(0.8299386,0.02553446){\makebox(0,0)[t]{\lineheight{1.25}\smash{\begin{tabular}[t]{c}\footnotesize{$40$}\end{tabular}}}}%
    \put(0.70925656,0.02553446){\makebox(0,0)[t]{\lineheight{1.25}\smash{\begin{tabular}[t]{c}\footnotesize{$20$}\end{tabular}}}}%
    \put(0.28699936,0.00833339){\color[rgb]{0,0,0}\makebox(0,0)[t]{\lineheight{1.25}\smash{\begin{tabular}[t]{c}\footnotesize{time [s]}\end{tabular}}}}%
    \put(0.76932932,0.00833339){\color[rgb]{0,0,0}\makebox(0,0)[t]{\lineheight{1.25}\smash{\begin{tabular}[t]{c}\footnotesize{time [s]}\end{tabular}}}}%
    \put(0.08248874,0.81595569){\makebox(0,0)[rt]{\lineheight{1.25}\smash{\begin{tabular}[t]{r}\footnotesize{$ 5$}\end{tabular}}}}%
    \put(0.08248874,0.87126044){\makebox(0,0)[rt]{\lineheight{1.25}\smash{\begin{tabular}[t]{r}\footnotesize{$ 10$}\end{tabular}}}}%
    \put(0.08248874,0.76318354){\makebox(0,0)[rt]{\lineheight{1.25}\smash{\begin{tabular}[t]{r}\footnotesize{$ 0$}\end{tabular}}}}%
    \put(0.08248874,0.70901){\makebox(0,0)[rt]{\lineheight{1.25}\smash{\begin{tabular}[t]{r}\footnotesize{$ 5$}\end{tabular}}}}%
    \put(0.08166098,0.23833912){\makebox(0,0)[rt]{\lineheight{1.25}\smash{\begin{tabular}[t]{r}\footnotesize{$0.00$}\end{tabular}}}}%
    \put(0.08166098,0.29364387){\makebox(0,0)[rt]{\lineheight{1.25}\smash{\begin{tabular}[t]{r}\footnotesize{$0.05$}\end{tabular}}}}%
    \put(0.08166098,0.19188932){\makebox(0,0)[rt]{\lineheight{1.25}\smash{\begin{tabular}[t]{r}\footnotesize{$-0.05$}\end{tabular}}}}%
    \put(0.08166098,0.13929637){\makebox(0,0)[rt]{\lineheight{1.25}\smash{\begin{tabular}[t]{r}\footnotesize{$-0.10$}\end{tabular}}}}%
    \put(0.08166098,0.08855419){\makebox(0,0)[rt]{\lineheight{1.25}\smash{\begin{tabular}[t]{r}\footnotesize{$-0.15$}\end{tabular}}}}%
    \put(0.08166098,0.52664999){\makebox(0,0)[rt]{\lineheight{1.25}\smash{\begin{tabular}[t]{r}\footnotesize{$ 12$}\end{tabular}}}}%
    \put(0.08166098,0.5578802){\makebox(0,0)[rt]{\lineheight{1.25}\smash{\begin{tabular}[t]{r}\footnotesize{$ 14$}\end{tabular}}}}%
    \put(0.08166098,0.58827709){\makebox(0,0)[rt]{\lineheight{1.25}\smash{\begin{tabular}[t]{r}\footnotesize{$ 16$}\end{tabular}}}}%
    \put(0.08166098,0.49600607){\makebox(0,0)[rt]{\lineheight{1.25}\smash{\begin{tabular}[t]{r}\footnotesize{$ 10$}\end{tabular}}}}%
    \put(0.08166098,0.46554136){\makebox(0,0)[rt]{\lineheight{1.25}\smash{\begin{tabular}[t]{r}\footnotesize{$ 8$}\end{tabular}}}}%
    \put(0.08166098,0.43507666){\makebox(0,0)[rt]{\lineheight{1.25}\smash{\begin{tabular}[t]{r}\footnotesize{$ 6$}\end{tabular}}}}%
    \put(0.08166098,0.40443274){\makebox(0,0)[rt]{\lineheight{1.25}\smash{\begin{tabular}[t]{r}\footnotesize{$ 4$}\end{tabular}}}}%
    \put(0.08166098,0.37396804){\makebox(0,0)[rt]{\lineheight{1.25}\smash{\begin{tabular}[t]{r}\footnotesize{$ 2$}\end{tabular}}}}%
    \put(0.0085608,0.93670957){\color[rgb]{0,0,0}\makebox(0,0)[lt]{\lineheight{1.25}\smash{\begin{tabular}[t]{l}(a)\end{tabular}}}}%
    \put(0.03349154,0.79420492){\color[rgb]{0,0,0}\rotatebox{90}{\makebox(0,0)[t]{\lineheight{1.25}\smash{\begin{tabular}[t]{c}${\boldsymbol q}_1$ [m]\end{tabular}}}}}%
    \put(0.51352717,0.79420491){\color[rgb]{0,0,0}\rotatebox{90}{\makebox(0,0)[t]{\lineheight{1.25}\smash{\begin{tabular}[t]{c}${\boldsymbol p}_1$ [kg·m/s]\end{tabular}}}}}%
    \put(0.03349154,0.48605137){\color[rgb]{0,0,0}\rotatebox{90}{\makebox(0,0)[t]{\lineheight{1.25}\smash{\begin{tabular}[t]{c}${\boldsymbol q}_2$ [m]\end{tabular}}}}}%
    \put(0.51352717,0.48605137){\color[rgb]{0,0,0}\rotatebox{90}{\makebox(0,0)[t]{\lineheight{1.25}\smash{\begin{tabular}[t]{c}${\boldsymbol p}_2$ [kg·m/s]\end{tabular}}}}}%
    \put(0.51362288,0.17445107){\color[rgb]{0,0,0}\rotatebox{90}{\makebox(0,0)[t]{\lineheight{1.25}\smash{\begin{tabular}[t]{c}$s_2$ [J/K]\end{tabular}}}}}%
    \put(0.51508825,0.93580688){\color[rgb]{0,0,0}\makebox(0,0)[t]{\lineheight{1.25}\smash{\begin{tabular}[t]{c}GENERIC\end{tabular}}}}%
    \put(0.5644296,0.84167462){\makebox(0,0)[rt]{\lineheight{1.25}\smash{\begin{tabular}[t]{r}\footnotesize{$ 1.0$}\end{tabular}}}}%
    \put(0.5644296,0.87448542){\makebox(0,0)[rt]{\lineheight{1.25}\smash{\begin{tabular}[t]{r}\footnotesize{$ 1.5$}\end{tabular}}}}%
    \put(0.5644296,0.90619792){\makebox(0,0)[rt]{\lineheight{1.25}\smash{\begin{tabular}[t]{r}\footnotesize{$ 2.0$}\end{tabular}}}}%
    \put(0.5644296,0.80945012){\makebox(0,0)[rt]{\lineheight{1.25}\smash{\begin{tabular}[t]{r}\footnotesize{$ 0.5$}\end{tabular}}}}%
    \put(0.5644296,0.77740482){\makebox(0,0)[rt]{\lineheight{1.25}\smash{\begin{tabular}[t]{r}\footnotesize{$ 0.0$}\end{tabular}}}}%
    \put(0.5644296,0.74535953){\makebox(0,0)[rt]{\lineheight{1.25}\smash{\begin{tabular}[t]{r}\footnotesize{$ -0.5$}\end{tabular}}}}%
    \put(0.5644296,0.71313502){\makebox(0,0)[rt]{\lineheight{1.25}\smash{\begin{tabular}[t]{r}\footnotesize{$ -1.0$}\end{tabular}}}}%
    \put(0.5644296,0.68108973){\makebox(0,0)[rt]{\lineheight{1.25}\smash{\begin{tabular}[t]{r}\footnotesize{$ -1.5$}\end{tabular}}}}%
    \put(0.5644296,0.53675409){\makebox(0,0)[rt]{\lineheight{1.25}\smash{\begin{tabular}[t]{r}\footnotesize{$ 1.0$}\end{tabular}}}}%
    \put(0.5644296,0.57114547){\makebox(0,0)[rt]{\lineheight{1.25}\smash{\begin{tabular}[t]{r}\footnotesize{$ 1.5$}\end{tabular}}}}%
    \put(0.5644296,0.60443865){\makebox(0,0)[rt]{\lineheight{1.25}\smash{\begin{tabular}[t]{r}\footnotesize{$ 2.0$}\end{tabular}}}}%
    \put(0.5644296,0.502949){\makebox(0,0)[rt]{\lineheight{1.25}\smash{\begin{tabular}[t]{r}\footnotesize{$ 0.5$}\end{tabular}}}}%
    \put(0.5644296,0.46932311){\makebox(0,0)[rt]{\lineheight{1.25}\smash{\begin{tabular}[t]{r}\footnotesize{$ 0.0$}\end{tabular}}}}%
    \put(0.5644296,0.43569722){\makebox(0,0)[rt]{\lineheight{1.25}\smash{\begin{tabular}[t]{r}\footnotesize{$ -0.5$}\end{tabular}}}}%
    \put(0.5644296,0.40189213){\makebox(0,0)[rt]{\lineheight{1.25}\smash{\begin{tabular}[t]{r}\footnotesize{$ -1.0$}\end{tabular}}}}%
    \put(0.5644296,0.36826625){\makebox(0,0)[rt]{\lineheight{1.25}\smash{\begin{tabular}[t]{r}\footnotesize{$ -1.5$}\end{tabular}}}}%
    \put(0.5644296,0.15348539){\makebox(0,0)[rt]{\lineheight{1.25}\smash{\begin{tabular}[t]{r}\footnotesize{$ 0.05$}\end{tabular}}}}%
    \put(0.5644296,0.20684384){\makebox(0,0)[rt]{\lineheight{1.25}\smash{\begin{tabular}[t]{r}\footnotesize{$ 0.10$}\end{tabular}}}}%
    \put(0.5644296,0.25936897){\makebox(0,0)[rt]{\lineheight{1.25}\smash{\begin{tabular}[t]{r}\footnotesize{$ 0.15$}\end{tabular}}}}%
    \put(0.5644296,0.10071305){\makebox(0,0)[rt]{\lineheight{1.25}\smash{\begin{tabular}[t]{r}\footnotesize{$ 0.00$}\end{tabular}}}}%
    \put(0.5644296,0.0481201){\makebox(0,0)[rt]{\lineheight{1.25}\smash{\begin{tabular}[t]{r}\footnotesize{$ -0.05$}\end{tabular}}}}%
    \put(0.41215231,0.27782859){\makebox(0,0)[lt]{\lineheight{1.25}\smash{\begin{tabular}[t]{l}\footnotesize{Net}\end{tabular}}}}%
    \put(0.41412312,0.25960515){\makebox(0,0)[lt]{\lineheight{1.25}\smash{\begin{tabular}[t]{l}\footnotesize{GT}\end{tabular}}}}%
  \end{picture}%
\endgroup%

%% file: DP_variablesSingle.eps_tex
\begingroup%
  \makeatletter%
  \providecommand\color[2][]{%
    \errmessage{(Inkscape) Color is used for the text in Inkscape, but the package 'color.sty' is not loaded}%
    \renewcommand\color[2][]{}%
  }%
  \providecommand\transparent[1]{%
    \errmessage{(Inkscape) Transparency is used (non-zero) for the text in Inkscape, but the package 'transparent.sty' is not loaded}%
    \renewcommand\transparent[1]{}%
  }%
  \providecommand\rotatebox[2]{#2}%
  \newcommand*\fsize{\dimexpr\f@size pt\relax}%
  \newcommand*\lineheight[1]{\fontsize{\fsize}{#1\fsize}\selectfont}%
  \ifx\svgwidth\undefined%
    \setlength{\unitlength}{948.96619776bp}%
    \ifx\svgscale\undefined%
      \relax%
    \else%
      \setlength{\unitlength}{\unitlength * \real{\svgscale}}%
    \fi%
  \else%
    \setlength{\unitlength}{\svgwidth}%
  \fi%
  \global\let\svgwidth\undefined%
  \global\let\svgscale\undefined%
  \makeatother%
  \begin{picture}(1,0.95025574)%
    \lineheight{1}%
    \setlength\tabcolsep{0pt}%
    \put(0,0){\includegraphics[width=\unitlength]{DP_variablesSingle.eps}}%
    \put(0.10515388,0.64407192){\makebox(0,0)[t]{\lineheight{1.25}\smash{\begin{tabular}[t]{c}\footnotesize{$0$}\end{tabular}}}}%
    \put(0.41089044,0.64407192){\makebox(0,0)[t]{\lineheight{1.25}\smash{\begin{tabular}[t]{c}\footnotesize{$50$}\end{tabular}}}}%
    \put(0.47130996,0.64407192){\makebox(0,0)[t]{\lineheight{1.25}\smash{\begin{tabular}[t]{c}\footnotesize{$60$}\end{tabular}}}}%
    \put(0.16679708,0.64407192){\makebox(0,0)[t]{\lineheight{1.25}\smash{\begin{tabular}[t]{c}\footnotesize{$10$}\end{tabular}}}}%
    \put(0.2872632,0.64407192){\makebox(0,0)[t]{\lineheight{1.25}\smash{\begin{tabular}[t]{c}\footnotesize{$30$}\end{tabular}}}}%
    \put(0.34854053,0.64407192){\makebox(0,0)[t]{\lineheight{1.25}\smash{\begin{tabular}[t]{c}\footnotesize{$40$}\end{tabular}}}}%
    \put(0.22785061,0.64407192){\makebox(0,0)[t]{\lineheight{1.25}\smash{\begin{tabular}[t]{c}\footnotesize{$20$}\end{tabular}}}}%
    \put(0.58660894,0.64407192){\makebox(0,0)[t]{\lineheight{1.25}\smash{\begin{tabular}[t]{c}\footnotesize{$0$}\end{tabular}}}}%
    \put(0.89234572,0.64407192){\makebox(0,0)[t]{\lineheight{1.25}\smash{\begin{tabular}[t]{c}\footnotesize{$50$}\end{tabular}}}}%
    \put(0.95276516,0.64407192){\makebox(0,0)[t]{\lineheight{1.25}\smash{\begin{tabular}[t]{c}\footnotesize{$60$}\end{tabular}}}}%
    \put(0.64825227,0.64407192){\makebox(0,0)[t]{\lineheight{1.25}\smash{\begin{tabular}[t]{c}\footnotesize{$10$}\end{tabular}}}}%
    \put(0.76871849,0.64407192){\makebox(0,0)[t]{\lineheight{1.25}\smash{\begin{tabular}[t]{c}\footnotesize{$30$}\end{tabular}}}}%
    \put(0.82999573,0.64407192){\makebox(0,0)[t]{\lineheight{1.25}\smash{\begin{tabular}[t]{c}\footnotesize{$40$}\end{tabular}}}}%
    \put(0.70930581,0.64407192){\makebox(0,0)[t]{\lineheight{1.25}\smash{\begin{tabular}[t]{c}\footnotesize{$20$}\end{tabular}}}}%
    \put(0.28702108,0.62686973){\color[rgb]{0,0,0}\makebox(0,0)[t]{\lineheight{1.25}\smash{\begin{tabular}[t]{c}\footnotesize{time [s]}\end{tabular}}}}%
    \put(0.76938249,0.62686973){\color[rgb]{0,0,0}\makebox(0,0)[t]{\lineheight{1.25}\smash{\begin{tabular}[t]{c}\footnotesize{time [s]}\end{tabular}}}}%
    \put(0.10515388,0.33504011){\makebox(0,0)[t]{\lineheight{1.25}\smash{\begin{tabular}[t]{c}\footnotesize{$0$}\end{tabular}}}}%
    \put(0.41089053,0.33504011){\makebox(0,0)[t]{\lineheight{1.25}\smash{\begin{tabular}[t]{c}\footnotesize{$50$}\end{tabular}}}}%
    \put(0.47130996,0.33504011){\makebox(0,0)[t]{\lineheight{1.25}\smash{\begin{tabular}[t]{c}\footnotesize{$60$}\end{tabular}}}}%
    \put(0.16679708,0.33504011){\makebox(0,0)[t]{\lineheight{1.25}\smash{\begin{tabular}[t]{c}\footnotesize{$10$}\end{tabular}}}}%
    \put(0.28726329,0.33504011){\makebox(0,0)[t]{\lineheight{1.25}\smash{\begin{tabular}[t]{c}\footnotesize{$30$}\end{tabular}}}}%
    \put(0.34854053,0.33504011){\makebox(0,0)[t]{\lineheight{1.25}\smash{\begin{tabular}[t]{c}\footnotesize{$40$}\end{tabular}}}}%
    \put(0.22785061,0.33504011){\makebox(0,0)[t]{\lineheight{1.25}\smash{\begin{tabular}[t]{c}\footnotesize{$20$}\end{tabular}}}}%
    \put(0.58660894,0.33504011){\makebox(0,0)[t]{\lineheight{1.25}\smash{\begin{tabular}[t]{c}\footnotesize{$0$}\end{tabular}}}}%
    \put(0.89234572,0.33504011){\makebox(0,0)[t]{\lineheight{1.25}\smash{\begin{tabular}[t]{c}\footnotesize{$50$}\end{tabular}}}}%
    \put(0.95276516,0.33504011){\makebox(0,0)[t]{\lineheight{1.25}\smash{\begin{tabular}[t]{c}\footnotesize{$60$}\end{tabular}}}}%
    \put(0.64825227,0.33504011){\makebox(0,0)[t]{\lineheight{1.25}\smash{\begin{tabular}[t]{c}\footnotesize{$10$}\end{tabular}}}}%
    \put(0.76871849,0.33504011){\makebox(0,0)[t]{\lineheight{1.25}\smash{\begin{tabular}[t]{c}\footnotesize{$30$}\end{tabular}}}}%
    \put(0.82999573,0.33504011){\makebox(0,0)[t]{\lineheight{1.25}\smash{\begin{tabular}[t]{c}\footnotesize{$40$}\end{tabular}}}}%
    \put(0.70930581,0.33504011){\makebox(0,0)[t]{\lineheight{1.25}\smash{\begin{tabular}[t]{c}\footnotesize{$20$}\end{tabular}}}}%
    \put(0.28702117,0.31783792){\color[rgb]{0,0,0}\makebox(0,0)[t]{\lineheight{1.25}\smash{\begin{tabular}[t]{c}\footnotesize{time [s]}\end{tabular}}}}%
    \put(0.76938249,0.31783792){\color[rgb]{0,0,0}\makebox(0,0)[t]{\lineheight{1.25}\smash{\begin{tabular}[t]{c}\footnotesize{time [s]}\end{tabular}}}}%
    \put(0.10515388,0.02553938){\makebox(0,0)[t]{\lineheight{1.25}\smash{\begin{tabular}[t]{c}\footnotesize{$0$}\end{tabular}}}}%
    \put(0.41089044,0.02553938){\makebox(0,0)[t]{\lineheight{1.25}\smash{\begin{tabular}[t]{c}\footnotesize{$50$}\end{tabular}}}}%
    \put(0.47130996,0.02553938){\makebox(0,0)[t]{\lineheight{1.25}\smash{\begin{tabular}[t]{c}\footnotesize{$60$}\end{tabular}}}}%
    \put(0.16679708,0.02553938){\makebox(0,0)[t]{\lineheight{1.25}\smash{\begin{tabular}[t]{c}\footnotesize{$10$}\end{tabular}}}}%
    \put(0.2872632,0.02553938){\makebox(0,0)[t]{\lineheight{1.25}\smash{\begin{tabular}[t]{c}\footnotesize{$30$}\end{tabular}}}}%
    \put(0.34854053,0.02553938){\makebox(0,0)[t]{\lineheight{1.25}\smash{\begin{tabular}[t]{c}\footnotesize{$40$}\end{tabular}}}}%
    \put(0.22785061,0.02553938){\makebox(0,0)[t]{\lineheight{1.25}\smash{\begin{tabular}[t]{c}\footnotesize{$20$}\end{tabular}}}}%
    \put(0.58660894,0.02553938){\makebox(0,0)[t]{\lineheight{1.25}\smash{\begin{tabular}[t]{c}\footnotesize{$0$}\end{tabular}}}}%
    \put(0.89234572,0.02553938){\makebox(0,0)[t]{\lineheight{1.25}\smash{\begin{tabular}[t]{c}\footnotesize{$50$}\end{tabular}}}}%
    \put(0.95276516,0.02553938){\makebox(0,0)[t]{\lineheight{1.25}\smash{\begin{tabular}[t]{c}\footnotesize{$60$}\end{tabular}}}}%
    \put(0.64825227,0.02553938){\makebox(0,0)[t]{\lineheight{1.25}\smash{\begin{tabular}[t]{c}\footnotesize{$10$}\end{tabular}}}}%
    \put(0.76871849,0.02553938){\makebox(0,0)[t]{\lineheight{1.25}\smash{\begin{tabular}[t]{c}\footnotesize{$30$}\end{tabular}}}}%
    \put(0.82999573,0.02553938){\makebox(0,0)[t]{\lineheight{1.25}\smash{\begin{tabular}[t]{c}\footnotesize{$40$}\end{tabular}}}}%
    \put(0.70930581,0.02553938){\makebox(0,0)[t]{\lineheight{1.25}\smash{\begin{tabular}[t]{c}\footnotesize{$20$}\end{tabular}}}}%
    \put(0.28702108,0.00833719){\color[rgb]{0,0,0}\makebox(0,0)[t]{\lineheight{1.25}\smash{\begin{tabular}[t]{c}\footnotesize{time [s]}\end{tabular}}}}%
    \put(0.76938249,0.00833719){\color[rgb]{0,0,0}\makebox(0,0)[t]{\lineheight{1.25}\smash{\begin{tabular}[t]{c}\footnotesize{time [s]}\end{tabular}}}}%
    \put(0.08146534,0.8163511){\makebox(0,0)[rt]{\lineheight{1.25}\smash{\begin{tabular}[t]{r}\footnotesize{$ 5$}\end{tabular}}}}%
    \put(0.08146534,0.87165945){\makebox(0,0)[rt]{\lineheight{1.25}\smash{\begin{tabular}[t]{r}\footnotesize{$ 10$}\end{tabular}}}}%
	\put(0.08166098,0.29364387){\makebox(0,0)[rt]{\lineheight{1.25}\smash{\begin{tabular}[t]{r}\footnotesize{$0.05$}\end{tabular}}}}%
    \put(0.08166098,0.23833912){\makebox(0,0)[rt]{\lineheight{1.25}\smash{\begin{tabular}[t]{r}\footnotesize{$0.00$}\end{tabular}}}}%
    \put(0.08166098,0.18303437){\makebox(0,0)[rt]{\lineheight{1.25}\smash{\begin{tabular}[t]{r}\footnotesize{$-0.05$}\end{tabular}}}}%
    \put(0.08166098,0.12772962){\makebox(0,0)[rt]{\lineheight{1.25}\smash{\begin{tabular}[t]{r}\footnotesize{$-0.10$}\end{tabular}}}}%
    \put(0.08166098,0.07242487){\makebox(0,0)[rt]{\lineheight{1.25}\smash{\begin{tabular}[t]{r}\footnotesize{$-0.15$}\end{tabular}}}}%
    \put(0.08146534,0.52677292){\makebox(0,0)[rt]{\lineheight{1.25}\smash{\begin{tabular}[t]{r}\footnotesize{$ 12$}\end{tabular}}}}%
    \put(0.08146534,0.55800516){\makebox(0,0)[rt]{\lineheight{1.25}\smash{\begin{tabular}[t]{r}\footnotesize{$ 14$}\end{tabular}}}}%
    \put(0.08146534,0.58840404){\makebox(0,0)[rt]{\lineheight{1.25}\smash{\begin{tabular}[t]{r}\footnotesize{$ 16$}\end{tabular}}}}%
    \put(0.08146534,0.49612701){\makebox(0,0)[rt]{\lineheight{1.25}\smash{\begin{tabular}[t]{r}\footnotesize{$ 10$}\end{tabular}}}}%
    \put(0.08146534,0.46566031){\makebox(0,0)[rt]{\lineheight{1.25}\smash{\begin{tabular}[t]{r}\footnotesize{$ 8$}\end{tabular}}}}%
    \put(0.08146534,0.43519362){\makebox(0,0)[rt]{\lineheight{1.25}\smash{\begin{tabular}[t]{r}\footnotesize{$ 6$}\end{tabular}}}}%
    \put(0.08146534,0.40454771){\makebox(0,0)[rt]{\lineheight{1.25}\smash{\begin{tabular}[t]{r}\footnotesize{$ 4$}\end{tabular}}}}%
    \put(0.08146534,0.37408101){\makebox(0,0)[rt]{\lineheight{1.25}\smash{\begin{tabular}[t]{r}\footnotesize{$ 2$}\end{tabular}}}}%
    \put(0.5644695,0.84263816){\makebox(0,0)[rt]{\lineheight{1.25}\smash{\begin{tabular}[t]{r}\footnotesize{$ 1.0$}\end{tabular}}}}%
    \put(0.5644695,0.8754511){\makebox(0,0)[rt]{\lineheight{1.25}\smash{\begin{tabular}[t]{r}\footnotesize{$ 1.5$}\end{tabular}}}}%
    \put(0.5644695,0.90716576){\makebox(0,0)[rt]{\lineheight{1.25}\smash{\begin{tabular}[t]{r}\footnotesize{$ 2.0$}\end{tabular}}}}%
    \put(0.5644695,0.81041156){\makebox(0,0)[rt]{\lineheight{1.25}\smash{\begin{tabular}[t]{r}\footnotesize{$ 0.5$}\end{tabular}}}}%
    \put(0.5644695,0.77836417){\makebox(0,0)[rt]{\lineheight{1.25}\smash{\begin{tabular}[t]{r}\footnotesize{$ 0.0$}\end{tabular}}}}%
    \put(0.5644695,0.74631679){\makebox(0,0)[rt]{\lineheight{1.25}\smash{\begin{tabular}[t]{r}\footnotesize{$ -0.5$}\end{tabular}}}}%
    \put(0.5644695,0.71409018){\makebox(0,0)[rt]{\lineheight{1.25}\smash{\begin{tabular}[t]{r}\footnotesize{$ -1.0$}\end{tabular}}}}%
    \put(0.5644695,0.6820428){\makebox(0,0)[rt]{\lineheight{1.25}\smash{\begin{tabular}[t]{r}\footnotesize{$ -1.5$}\end{tabular}}}}%
    \put(0.5644695,0.53769774){\makebox(0,0)[rt]{\lineheight{1.25}\smash{\begin{tabular}[t]{r}\footnotesize{$ 1.0$}\end{tabular}}}}%
    \put(0.5644695,0.57209137){\makebox(0,0)[rt]{\lineheight{1.25}\smash{\begin{tabular}[t]{r}\footnotesize{$ 1.5$}\end{tabular}}}}%
    \put(0.5644695,0.60538672){\makebox(0,0)[rt]{\lineheight{1.25}\smash{\begin{tabular}[t]{r}\footnotesize{$ 2.0$}\end{tabular}}}}%
    \put(0.5644695,0.50389045){\makebox(0,0)[rt]{\lineheight{1.25}\smash{\begin{tabular}[t]{r}\footnotesize{$ 0.5$}\end{tabular}}}}%
    \put(0.5644695,0.47026237){\makebox(0,0)[rt]{\lineheight{1.25}\smash{\begin{tabular}[t]{r}\footnotesize{$ 0.0$}\end{tabular}}}}%
    \put(0.5644695,0.43663429){\makebox(0,0)[rt]{\lineheight{1.25}\smash{\begin{tabular}[t]{r}\footnotesize{$ -0.5$}\end{tabular}}}}%
    \put(0.5644695,0.40282699){\makebox(0,0)[rt]{\lineheight{1.25}\smash{\begin{tabular}[t]{r}\footnotesize{$ -1.0$}\end{tabular}}}}%
    \put(0.5644695,0.36919892){\makebox(0,0)[rt]{\lineheight{1.25}\smash{\begin{tabular}[t]{r}\footnotesize{$ -1.5$}\end{tabular}}}}%
    \put(0.56295517,0.15124267){\makebox(0,0)[rt]{\lineheight{1.25}\smash{\begin{tabular}[t]{r}\footnotesize{$ 0.05$}\end{tabular}}}}%
    \put(0.56295517,0.20144321){\makebox(0,0)[rt]{\lineheight{1.25}\smash{\begin{tabular}[t]{r}\footnotesize{$ 0.10$}\end{tabular}}}}%
    \put(0.56295517,0.25081039){\makebox(0,0)[rt]{\lineheight{1.25}\smash{\begin{tabular}[t]{r}\footnotesize{$ 0.15$}\end{tabular}}}}%
    \put(0.56295517,0.10004776){\makebox(0,0)[rt]{\lineheight{1.25}\smash{\begin{tabular}[t]{r}\footnotesize{$ 0.00$}\end{tabular}}}}%
    \put(0.56295517,0.05061277){\makebox(0,0)[rt]{\lineheight{1.25}\smash{\begin{tabular}[t]{r}\footnotesize{$ -0.05$}\end{tabular}}}}%
    \put(0.08146534,0.7635755){\makebox(0,0)[rt]{\lineheight{1.25}\smash{\begin{tabular}[t]{r}\footnotesize{$ 0$}\end{tabular}}}}%
    \put(0.08146534,0.70939843){\makebox(0,0)[rt]{\lineheight{1.25}\smash{\begin{tabular}[t]{r}\footnotesize{$ 5$}\end{tabular}}}}%
    \put(0.03349686,0.79717167){\color[rgb]{0,0,0}\rotatebox{90}{\makebox(0,0)[t]{\lineheight{1.25}\smash{\begin{tabular}[t]{c}${\boldsymbol q}_1$ [m]\end{tabular}}}}}%
    \put(0.51356375,0.79717167){\color[rgb]{0,0,0}\rotatebox{90}{\makebox(0,0)[t]{\lineheight{1.25}\smash{\begin{tabular}[t]{c}${\boldsymbol p}_1$ [kg·m/s]\end{tabular}}}}}%
    \put(0.03349686,0.48899803){\color[rgb]{0,0,0}\rotatebox{90}{\makebox(0,0)[t]{\lineheight{1.25}\smash{\begin{tabular}[t]{c}${\boldsymbol q}_2$ [m]\end{tabular}}}}}%
    \put(0.51356375,0.48899803){\color[rgb]{0,0,0}\rotatebox{90}{\makebox(0,0)[t]{\lineheight{1.25}\smash{\begin{tabular}[t]{c}${\boldsymbol p}_2$ [kg·m/s]\end{tabular}}}}}%
    \put(0.02881668,0.17737732){\color[rgb]{0,0,0}\rotatebox{90}{\makebox(0,0)[t]{\lineheight{1.25}\smash{\begin{tabular}[t]{c}$s_1$ [J/K]\end{tabular}}}}}%
    \put(0.51365947,0.17737732){\color[rgb]{0,0,0}\rotatebox{90}{\makebox(0,0)[t]{\lineheight{1.25}\smash{\begin{tabular}[t]{c}$s_2$ [J/K]\end{tabular}}}}}%
    \put(0.51479148,0.93610317){\color[rgb]{0,0,0}\makebox(0,0)[t]{\lineheight{1.25}\smash{\begin{tabular}[t]{c}Single generator\end{tabular}}}}%
    \put(0.61338789,0.90018249){\makebox(0,0)[lt]{\lineheight{1.25}\smash{\begin{tabular}[t]{l}\footnotesize{Net (X)}\end{tabular}}}}%
    \put(0.61338789,0.88372172){\makebox(0,0)[lt]{\lineheight{1.25}\smash{\begin{tabular}[t]{l}\footnotesize{Net (Y)}\end{tabular}}}}%
    \put(0.61338789,0.86582295){\makebox(0,0)[lt]{\lineheight{1.25}\smash{\begin{tabular}[t]{l}\footnotesize{GT}\end{tabular}}}}%
    \put(0.90371167,0.59137652){\makebox(0,0)[lt]{\lineheight{1.25}\smash{\begin{tabular}[t]{l}\footnotesize{Net (X)}\end{tabular}}}}%
    \put(0.90371167,0.57491575){\makebox(0,0)[lt]{\lineheight{1.25}\smash{\begin{tabular}[t]{l}\footnotesize{Net (Y)}\end{tabular}}}}%
    \put(0.90371167,0.55701698){\makebox(0,0)[lt]{\lineheight{1.25}\smash{\begin{tabular}[t]{l}\footnotesize{GT}\end{tabular}}}}%
    \put(0.42184054,0.59137652){\makebox(0,0)[lt]{\lineheight{1.25}\smash{\begin{tabular}[t]{l}\footnotesize{Net (X)}\end{tabular}}}}%
    \put(0.42184054,0.57491575){\makebox(0,0)[lt]{\lineheight{1.25}\smash{\begin{tabular}[t]{l}\footnotesize{Net (Y)}\end{tabular}}}}%
    \put(0.42184054,0.55701698){\makebox(0,0)[lt]{\lineheight{1.25}\smash{\begin{tabular}[t]{l}\footnotesize{GT}\end{tabular}}}}%
    \put(0.4212759,0.90018249){\makebox(0,0)[lt]{\lineheight{1.25}\smash{\begin{tabular}[t]{l}\footnotesize{Net (X)}\end{tabular}}}}%
    \put(0.4212759,0.88372172){\makebox(0,0)[lt]{\lineheight{1.25}\smash{\begin{tabular}[t]{l}\footnotesize{Net (Y)}\end{tabular}}}}%
    \put(0.4212759,0.86582295){\makebox(0,0)[lt]{\lineheight{1.25}\smash{\begin{tabular}[t]{l}\footnotesize{GT}\end{tabular}}}}%
    \put(0.62385376,0.27882522){\makebox(0,0)[lt]{\lineheight{1.25}\smash{\begin{tabular}[t]{l}\footnotesize{Net}\end{tabular}}}}%
    \put(0.62582471,0.26060059){\makebox(0,0)[lt]{\lineheight{1.25}\smash{\begin{tabular}[t]{l}\footnotesize{GT}\end{tabular}}}}%
    \put(0.412114,0.27882522){\makebox(0,0)[lt]{\lineheight{1.25}\smash{\begin{tabular}[t]{l}\footnotesize{Net}\end{tabular}}}}%
    \put(0.41408494,0.26060059){\makebox(0,0)[lt]{\lineheight{1.25}\smash{\begin{tabular}[t]{l}\footnotesize{GT}\end{tabular}}}}%
  \end{picture}%
\endgroup%

%% file: DP_BoxPlots.eps_tex
\begingroup%
  \makeatletter%
  \providecommand\color[2][]{%
    \errmessage{(Inkscape) Color is used for the text in Inkscape, but the package 'color.sty' is not loaded}%
    \renewcommand\color[2][]{}%
  }%
  \providecommand\transparent[1]{%
    \errmessage{(Inkscape) Transparency is used (non-zero) for the text in Inkscape, but the package 'transparent.sty' is not loaded}%
    \renewcommand\transparent[1]{}%
  }%
  \providecommand\rotatebox[2]{#2}%
  \newcommand*\fsize{\dimexpr\f@size pt\relax}%
  \newcommand*\lineheight[1]{\fontsize{\fsize}{#1\fsize}\selectfont}%
  \ifx\svgwidth\undefined%
    \setlength{\unitlength}{930.91393124bp}%
    \ifx\svgscale\undefined%
      \relax%
    \else%
      \setlength{\unitlength}{\unitlength * \real{\svgscale}}%
    \fi%
  \else%
    \setlength{\unitlength}{\svgwidth}%
  \fi%
  \global\let\svgwidth\undefined%
  \global\let\svgscale\undefined%
  \makeatother%
  \begin{picture}(1,0.37167773)%
    \lineheight{1}%
    \setlength\tabcolsep{0pt}%
    \put(0,0){\includegraphics[width=\unitlength]{DP_BoxPlots.eps}}%
    \put(0.19342745,0.00){\color[rgb]{0,0,0}\makebox(0,0)[t]{\lineheight{1.25}\smash{\begin{tabular}[t]{c}GENERIC\end{tabular}}}}%
    \put(0.38516103,0.00){\color[rgb]{0,0,0}\makebox(0,0)[t]{\lineheight{1.25}\smash{\begin{tabular}[t]{c}Single generator\end{tabular}}}}%
    \put(0.83105456,0.0239601){\makebox(0,0)[t]{\lineheight{1.25}\smash{\begin{tabular}[t]{c}\footnotesize{train}\end{tabular}}}}%
    \put(0.91406975,0.0239601){\makebox(0,0)[t]{\lineheight{1.25}\smash{\begin{tabular}[t]{c}\footnotesize{test}\end{tabular}}}}%
    \put(0.63924305,0.0239601){\makebox(0,0)[t]{\lineheight{1.25}\smash{\begin{tabular}[t]{c}\footnotesize{train}\end{tabular}}}}%
    \put(0.72225825,0.0239601){\makebox(0,0)[t]{\lineheight{1.25}\smash{\begin{tabular}[t]{c}\footnotesize{test}\end{tabular}}}}%
    \put(0.01925293,0.35543903){\color[rgb]{0,0,0}\makebox(0,0)[lt]{\lineheight{1.25}\smash{\begin{tabular}[t]{l}(a)\end{tabular}}}}%
    \put(0.02965019,0.17077104){\rotatebox{90}{\makebox(0,0)[lt]{\lineheight{1.25}\smash{\begin{tabular}[t]{l}MSE\end{tabular}}}}}%
    \put(0.06514043,0.26375479){\makebox(0,0)[t]{\lineheight{1.25}\smash{\begin{tabular}[t]{c}\footnotesize{$10^{-2}$}\end{tabular}}}}%
    \put(0.06514043,0.17228642){\makebox(0,0)[t]{\lineheight{1.25}\smash{\begin{tabular}[t]{c}\footnotesize{$10^{-3}$}\end{tabular}}}}%
    \put(0.06514043,0.07985061){\makebox(0,0)[t]{\lineheight{1.25}\smash{\begin{tabular}[t]{c}\footnotesize{$10^{-4}$}\end{tabular}}}}%
    \put(0.68036828,0.00){\color[rgb]{0,0,0}\makebox(0,0)[t]{\lineheight{1.25}\smash{\begin{tabular}[t]{c}GENERIC\end{tabular}}}}%
    \put(0.87210186,0.00){\color[rgb]{0,0,0}\makebox(0,0)[t]{\lineheight{1.25}\smash{\begin{tabular}[t]{c}Single generator\end{tabular}}}}%
    \put(0.52941503,0.17077107){\rotatebox{90}{\makebox(0,0)[lt]{\lineheight{1.25}\smash{\begin{tabular}[t]{l}MSE\end{tabular}}}}}%
    \put(0.55691522,0.27342274){\makebox(0,0)[t]{\lineheight{1.25}\smash{\begin{tabular}[t]{c}\footnotesize{$10^{2}$}\end{tabular}}}}%
    \put(0.55691522,0.16100719){\makebox(0,0)[t]{\lineheight{1.25}\smash{\begin{tabular}[t]{c}\footnotesize{$10^{1}$}\end{tabular}}}}%
    \put(0.15230224,0.0239601){\makebox(0,0)[t]{\lineheight{1.25}\smash{\begin{tabular}[t]{c}\footnotesize{train}\end{tabular}}}}%
    \put(0.23531743,0.0239601){\makebox(0,0)[t]{\lineheight{1.25}\smash{\begin{tabular}[t]{c}\footnotesize{test}\end{tabular}}}}%
    \put(0.55691522,0.04923556){\makebox(0,0)[t]{\lineheight{1.25}\smash{\begin{tabular}[t]{c}\footnotesize{$10^{0}$}\end{tabular}}}}%
    \put(0.34411373,0.0239601){\makebox(0,0)[t]{\lineheight{1.25}\smash{\begin{tabular}[t]{c}\footnotesize{train}\end{tabular}}}}%
    \put(0.42712892,0.0239601){\makebox(0,0)[t]{\lineheight{1.25}\smash{\begin{tabular}[t]{c}\footnotesize{test}\end{tabular}}}}%
    \put(0.51875579,0.35543905){\color[rgb]{0,0,0}\makebox(0,0)[lt]{\lineheight{1.25}\smash{\begin{tabular}[t]{l}(b)\end{tabular}}}}%
  \end{picture}%
\endgroup%

%% file: VE_Losses.eps_tex
\begingroup%
  \makeatletter%
  \providecommand\color[2][]{%
    \errmessage{(Inkscape) Color is used for the text in Inkscape, but the package 'color.sty' is not loaded}%
    \renewcommand\color[2][]{}%
  }%
  \providecommand\transparent[1]{%
    \errmessage{(Inkscape) Transparency is used (non-zero) for the text in Inkscape, but the package 'transparent.sty' is not loaded}%
    \renewcommand\transparent[1]{}%
  }%
  \providecommand\rotatebox[2]{#2}%
  \newcommand*\fsize{\dimexpr\f@size pt\relax}%
  \newcommand*\lineheight[1]{\fontsize{\fsize}{#1\fsize}\selectfont}%
  \ifx\svgwidth\undefined%
    \setlength{\unitlength}{983.81042257bp}%
    \ifx\svgscale\undefined%
      \relax%
    \else%
      \setlength{\unitlength}{\unitlength * \real{\svgscale}}%
    \fi%
  \else%
    \setlength{\unitlength}{\svgwidth}%
  \fi%
  \global\let\svgwidth\undefined%
  \global\let\svgscale\undefined%
  \makeatother%
  \begin{picture}(1,0.3320007)%
    \lineheight{1}%
    \setlength\tabcolsep{0pt}%
    \put(0,0){\includegraphics[width=\unitlength]{VE_Losses.eps}}%
    \put(0.35920851,0.24645356){\makebox(0,0)[lt]{\lineheight{1.25}\smash{\begin{tabular}[t]{l}\footnotesize{$\mathcal{L}^{\text{total}}$}\end{tabular}}}}%
    \put(0.35920851,0.28295704){\makebox(0,0)[lt]{\lineheight{1.25}\smash{\begin{tabular}[t]{l}\footnotesize{$\mathcal{L}^{\text{data}}$}\end{tabular}}}}%
    \put(0.35920851,0.2647053){\makebox(0,0)[lt]{\lineheight{1.25}\smash{\begin{tabular}[t]{l}\footnotesize{$\mathcal{L}^{\text{degen}}$}\end{tabular}}}}%
    \put(0.01432251,0.3130355){\color[rgb]{0,0,0}\makebox(0,0)[lt]{\lineheight{1.25}\smash{\begin{tabular}[t]{l}(a)\end{tabular}}}}%
    \put(0.05213928,0.10799645){\makebox(0,0)[t]{\lineheight{1.25}\smash{\begin{tabular}[t]{c}\footnotesize{$10^{-4}$}\end{tabular}}}}%
    \put(0.02039947,0.17326322){\rotatebox{90}{\makebox(0,0)[t]{\lineheight{1.25}\smash{\begin{tabular}[t]{c}Loss\end{tabular}}}}}%
    \put(0.1518641,0.02548739){\makebox(0,0)[t]{\lineheight{1.25}\smash{\begin{tabular}[t]{c}\footnotesize{$1000$}\end{tabular}}}}%
    \put(0.26120994,0.00173004){\makebox(0,0)[t]{\lineheight{1.25}\smash{\begin{tabular}[t]{c}Epoch\end{tabular}}}}%
    \put(0.05213928,0.29692458){\makebox(0,0)[t]{\lineheight{1.25}\smash{\begin{tabular}[t]{c}\footnotesize{$10^{1}$}\end{tabular}}}}%
    \put(0.05213928,0.25898888){\makebox(0,0)[t]{\lineheight{1.25}\smash{\begin{tabular}[t]{c}\footnotesize{$10^{0}$}\end{tabular}}}}%
    \put(0.05213928,0.2214904){\makebox(0,0)[t]{\lineheight{1.25}\smash{\begin{tabular}[t]{c}\footnotesize{$10^{-1}$}\end{tabular}}}}%
    \put(0.05213928,0.18407983){\makebox(0,0)[t]{\lineheight{1.25}\smash{\begin{tabular}[t]{c}\footnotesize{$10^{-2}$}\end{tabular}}}}%
    \put(0.05230009,0.14619239){\makebox(0,0)[t]{\lineheight{1.25}\smash{\begin{tabular}[t]{c}\footnotesize{$10^{-3}$}\end{tabular}}}}%
    \put(0.05213928,0.06919906){\makebox(0,0)[t]{\lineheight{1.25}\smash{\begin{tabular}[t]{c}\footnotesize{$10^{-5}$}\end{tabular}}}}%
    \put(0.09654154,0.02548739){\makebox(0,0)[t]{\lineheight{1.25}\smash{\begin{tabular}[t]{c}\footnotesize{$0$}\end{tabular}}}}%
    \put(0.20656599,0.02548739){\makebox(0,0)[t]{\lineheight{1.25}\smash{\begin{tabular}[t]{c}\footnotesize{$2000$}\end{tabular}}}}%
    \put(0.31678658,0.02548739){\makebox(0,0)[t]{\lineheight{1.25}\smash{\begin{tabular}[t]{c}\footnotesize{$4000$}\end{tabular}}}}%
    \put(0.42644875,0.02548739){\makebox(0,0)[t]{\lineheight{1.25}\smash{\begin{tabular}[t]{c}\footnotesize{$6000$}\end{tabular}}}}%
    \put(0.37179904,0.02548739){\makebox(0,0)[t]{\lineheight{1.25}\smash{\begin{tabular}[t]{c}\footnotesize{$5000$}\end{tabular}}}}%
    \put(0.26161512,0.02548739){\makebox(0,0)[t]{\lineheight{1.25}\smash{\begin{tabular}[t]{c}\footnotesize{$3000$}\end{tabular}}}}%
    \put(0.48412606,0.31152567){\color[rgb]{0,0,0}\makebox(0,0)[lt]{\lineheight{1.25}\smash{\begin{tabular}[t]{l}(b)\end{tabular}}}}%
    \put(0.82978333,0.24494373){\makebox(0,0)[lt]{\lineheight{1.25}\smash{\begin{tabular}[t]{l}\footnotesize{$\mathcal{L}^{\text{total}}$}\end{tabular}}}}%
    \put(0.82978333,0.28144721){\makebox(0,0)[lt]{\lineheight{1.25}\smash{\begin{tabular}[t]{l}\footnotesize{$\mathcal{L}^{\text{data}}$}\end{tabular}}}}%
    \put(0.82978333,0.26319547){\makebox(0,0)[lt]{\lineheight{1.25}\smash{\begin{tabular}[t]{l}\footnotesize{$\mathcal{L}^{\text{degen}}$}\end{tabular}}}}%
    \put(0.48982558,0.17320626){\rotatebox{90}{\makebox(0,0)[t]{\lineheight{1.25}\smash{\begin{tabular}[t]{c}Loss\end{tabular}}}}}%
    \put(0.62166707,0.02532547){\makebox(0,0)[t]{\lineheight{1.25}\smash{\begin{tabular}[t]{c}\footnotesize{$1000$}\end{tabular}}}}%
    \put(0.73101289,0.00156806){\makebox(0,0)[t]{\lineheight{1.25}\smash{\begin{tabular}[t]{c}Epoch\end{tabular}}}}%
    \put(0.52156544,0.22843827){\makebox(0,0)[t]{\lineheight{1.25}\smash{\begin{tabular}[t]{c}\footnotesize{$10^{-4}$}\end{tabular}}}}%
    \put(0.52156544,0.13546267){\makebox(0,0)[t]{\lineheight{1.25}\smash{\begin{tabular}[t]{c}\footnotesize{$10^{-5}$}\end{tabular}}}}%
    \put(0.52156544,0.04169783){\makebox(0,0)[t]{\lineheight{1.25}\smash{\begin{tabular}[t]{c}\footnotesize{$10^{-6}$}\end{tabular}}}}%
    \put(0.56634454,0.02532547){\makebox(0,0)[t]{\lineheight{1.25}\smash{\begin{tabular}[t]{c}\footnotesize{$0$}\end{tabular}}}}%
    \put(0.67636895,0.02532547){\makebox(0,0)[t]{\lineheight{1.25}\smash{\begin{tabular}[t]{c}\footnotesize{$2000$}\end{tabular}}}}%
    \put(0.78658953,0.02532547){\makebox(0,0)[t]{\lineheight{1.25}\smash{\begin{tabular}[t]{c}\footnotesize{$4000$}\end{tabular}}}}%
    \put(0.89625176,0.02532547){\makebox(0,0)[t]{\lineheight{1.25}\smash{\begin{tabular}[t]{c}\footnotesize{$6000$}\end{tabular}}}}%
    \put(0.84160199,0.02532547){\makebox(0,0)[t]{\lineheight{1.25}\smash{\begin{tabular}[t]{c}\footnotesize{$5000$}\end{tabular}}}}%
    \put(0.73141813,0.02532547){\makebox(0,0)[t]{\lineheight{1.25}\smash{\begin{tabular}[t]{c}\footnotesize{$3000$}\end{tabular}}}}%
  \end{picture}%
\endgroup%

%% file: VE_variables.eps_tex
\begingroup%
  \makeatletter%
  \providecommand\color[2][]{%
    \errmessage{(Inkscape) Color is used for the text in Inkscape, but the package 'color.sty' is not loaded}%
    \renewcommand\color[2][]{}%
  }%
  \providecommand\transparent[1]{%
    \errmessage{(Inkscape) Transparency is used (non-zero) for the text in Inkscape, but the package 'transparent.sty' is not loaded}%
    \renewcommand\transparent[1]{}%
  }%
  \providecommand\rotatebox[2]{#2}%
  \newcommand*\fsize{\dimexpr\f@size pt\relax}%
  \newcommand*\lineheight[1]{\fontsize{\fsize}{#1\fsize}\selectfont}%
  \ifx\svgwidth\undefined%
    \setlength{\unitlength}{947.87071889bp}%
    \ifx\svgscale\undefined%
      \relax%
    \else%
      \setlength{\unitlength}{\unitlength * \real{\svgscale}}%
    \fi%
  \else%
    \setlength{\unitlength}{\svgwidth}%
  \fi%
  \global\let\svgwidth\undefined%
  \global\let\svgscale\undefined%
  \makeatother%
  \begin{picture}(1,1.32216648)%
    \lineheight{1}%
    \setlength\tabcolsep{0pt}%
    \put(0,0){\includegraphics[width=\unitlength]{VE_variables.eps}}%
    \put(0.14653629,0.94299573){\makebox(0,0)[lt]{\lineheight{1.25}\smash{\begin{tabular}[t]{l}\footnotesize{Net}\end{tabular}}}}%
    \put(0.14850951,0.91271778){\makebox(0,0)[lt]{\lineheight{1.25}\smash{\begin{tabular}[t]{l}\footnotesize{GT}\end{tabular}}}}%
    \put(0.89932272,0.94299573){\makebox(0,0)[lt]{\lineheight{1.25}\smash{\begin{tabular}[t]{l}\footnotesize{Net}\end{tabular}}}}%
    \put(0.90129594,0.91271778){\makebox(0,0)[lt]{\lineheight{1.25}\smash{\begin{tabular}[t]{l}\footnotesize{GT}\end{tabular}}}}%
    \put(0.89932583,0.41222982){\makebox(0,0)[lt]{\lineheight{1.25}\smash{\begin{tabular}[t]{l}\footnotesize{Net}\end{tabular}}}}%
    \put(0.90129905,0.38195188){\makebox(0,0)[lt]{\lineheight{1.25}\smash{\begin{tabular}[t]{l}\footnotesize{GT}\end{tabular}}}}%
    \put(0.89932583,0.28022712){\makebox(0,0)[lt]{\lineheight{1.25}\smash{\begin{tabular}[t]{l}\footnotesize{Net}\end{tabular}}}}%
    \put(0.90129905,0.24994918){\makebox(0,0)[lt]{\lineheight{1.25}\smash{\begin{tabular}[t]{l}\footnotesize{GT}\end{tabular}}}}%
    \put(0.89932272,1.0727966){\makebox(0,0)[lt]{\lineheight{1.25}\smash{\begin{tabular}[t]{l}\footnotesize{Net}\end{tabular}}}}%
    \put(0.90129594,1.04251866){\makebox(0,0)[lt]{\lineheight{1.25}\smash{\begin{tabular}[t]{l}\footnotesize{GT}\end{tabular}}}}%
    \put(0.14653629,1.26045424){\makebox(0,0)[lt]{\lineheight{1.25}\smash{\begin{tabular}[t]{l}\footnotesize{Net}\end{tabular}}}}%
    \put(0.14850951,1.2301763){\makebox(0,0)[lt]{\lineheight{1.25}\smash{\begin{tabular}[t]{l}\footnotesize{GT}\end{tabular}}}}%
    \put(0.1465394,0.60171974){\makebox(0,0)[lt]{\lineheight{1.25}\smash{\begin{tabular}[t]{l}\footnotesize{Net}\end{tabular}}}}%
    \put(0.14851262,0.5714418){\makebox(0,0)[lt]{\lineheight{1.25}\smash{\begin{tabular}[t]{l}\footnotesize{GT}\end{tabular}}}}%
    \put(0.1465394,0.28022712){\makebox(0,0)[lt]{\lineheight{1.25}\smash{\begin{tabular}[t]{l}\footnotesize{Net}\end{tabular}}}}%
    \put(0.14851262,0.24994918){\makebox(0,0)[lt]{\lineheight{1.25}\smash{\begin{tabular}[t]{l}\footnotesize{GT}\end{tabular}}}}%
    \put(0.08591976,0.25456966){\makebox(0,0)[rt]{\lineheight{1.25}\smash{\begin{tabular}[t]{r}\footnotesize{$ 0.30$}\end{tabular}}}}%
    \put(0.08591976,0.28778667){\makebox(0,0)[rt]{\lineheight{1.25}\smash{\begin{tabular}[t]{r}\footnotesize{$ 0.35$}\end{tabular}}}}%
    \put(0.08591976,0.22135265){\makebox(0,0)[rt]{\lineheight{1.25}\smash{\begin{tabular}[t]{r}\footnotesize{$ 0.25$}\end{tabular}}}}%
    \put(0.08591976,0.18907292){\makebox(0,0)[rt]{\lineheight{1.25}\smash{\begin{tabular}[t]{r}\footnotesize{$ 0.20$}\end{tabular}}}}%
    \put(0.08591976,0.15382346){\makebox(0,0)[rt]{\lineheight{1.25}\smash{\begin{tabular}[t]{r}\footnotesize{$ 0.15$}\end{tabular}}}}%
    \put(0.08591976,0.12063748){\makebox(0,0)[rt]{\lineheight{1.25}\smash{\begin{tabular}[t]{r}\footnotesize{$ 0.10$}\end{tabular}}}}%
    \put(0.08591976,0.08745168){\makebox(0,0)[rt]{\lineheight{1.25}\smash{\begin{tabular}[t]{r}\footnotesize{$ 0.05$}\end{tabular}}}}%
    \put(0.08591976,0.05370296){\makebox(0,0)[rt]{\lineheight{1.25}\smash{\begin{tabular}[t]{r}\footnotesize{$ 0.00$}\end{tabular}}}}%
    \put(0.11002274,0.02556881){\makebox(0,0)[t]{\lineheight{1.25}\smash{\begin{tabular}[t]{c}\footnotesize{$0$}\end{tabular}}}}%
    \put(0.47783109,0.02556881){\makebox(0,0)[t]{\lineheight{1.25}\smash{\begin{tabular}[t]{c}\footnotesize{$1.0$}\end{tabular}}}}%
    \put(0.18281503,0.02556881){\makebox(0,0)[t]{\lineheight{1.25}\smash{\begin{tabular}[t]{c}\footnotesize{$0.2$}\end{tabular}}}}%
    \put(0.3303232,0.02556881){\makebox(0,0)[t]{\lineheight{1.25}\smash{\begin{tabular}[t]{c}\footnotesize{$0.6$}\end{tabular}}}}%
    \put(0.4043315,0.02556881){\makebox(0,0)[t]{\lineheight{1.25}\smash{\begin{tabular}[t]{c}\footnotesize{$0.8$}\end{tabular}}}}%
    \put(0.25659928,0.02556881){\makebox(0,0)[t]{\lineheight{1.25}\smash{\begin{tabular}[t]{c}\footnotesize{$0.4$}\end{tabular}}}}%
    \put(0.59234968,0.02556881){\makebox(0,0)[t]{\lineheight{1.25}\smash{\begin{tabular}[t]{c}\footnotesize{$0$}\end{tabular}}}}%
    \put(0.96015813,0.02556881){\makebox(0,0)[t]{\lineheight{1.25}\smash{\begin{tabular}[t]{c}\footnotesize{$1.0$}\end{tabular}}}}%
    \put(0.66514198,0.02556881){\makebox(0,0)[t]{\lineheight{1.25}\smash{\begin{tabular}[t]{c}\footnotesize{$0.2$}\end{tabular}}}}%
    \put(0.81265015,0.02556881){\makebox(0,0)[t]{\lineheight{1.25}\smash{\begin{tabular}[t]{c}\footnotesize{$0.6$}\end{tabular}}}}%
    \put(0.88665845,0.02556881){\makebox(0,0)[t]{\lineheight{1.25}\smash{\begin{tabular}[t]{c}\footnotesize{$0.8$}\end{tabular}}}}%
    \put(0.73892623,0.02556881){\makebox(0,0)[t]{\lineheight{1.25}\smash{\begin{tabular}[t]{c}\footnotesize{$0.4$}\end{tabular}}}}%
    \put(0.29368288,0.00834674){\color[rgb]{0,0,0}\makebox(0,0)[t]{\lineheight{1.25}\smash{\begin{tabular}[t]{c}\footnotesize{time [-]}\end{tabular}}}}%
    \put(0.77660167,0.00834674){\color[rgb]{0,0,0}\makebox(0,0)[t]{\lineheight{1.25}\smash{\begin{tabular}[t]{c}\footnotesize{time [-]}\end{tabular}}}}%
    \put(0.11002278,0.34686518){\makebox(0,0)[t]{\lineheight{1.25}\smash{\begin{tabular}[t]{c}\footnotesize{$0$}\end{tabular}}}}%
    \put(0.47783119,0.34686518){\makebox(0,0)[t]{\lineheight{1.25}\smash{\begin{tabular}[t]{c}\footnotesize{$1.0$}\end{tabular}}}}%
    \put(0.18281512,0.34686518){\makebox(0,0)[t]{\lineheight{1.25}\smash{\begin{tabular}[t]{c}\footnotesize{$0.2$}\end{tabular}}}}%
    \put(0.33032329,0.34686518){\makebox(0,0)[t]{\lineheight{1.25}\smash{\begin{tabular}[t]{c}\footnotesize{$0.6$}\end{tabular}}}}%
    \put(0.40433168,0.34686518){\makebox(0,0)[t]{\lineheight{1.25}\smash{\begin{tabular}[t]{c}\footnotesize{$0.8$}\end{tabular}}}}%
    \put(0.25659937,0.34686518){\makebox(0,0)[t]{\lineheight{1.25}\smash{\begin{tabular}[t]{c}\footnotesize{$0.4$}\end{tabular}}}}%
    \put(0.59234986,0.34686518){\makebox(0,0)[t]{\lineheight{1.25}\smash{\begin{tabular}[t]{c}\footnotesize{$0$}\end{tabular}}}}%
    \put(0.96015813,0.34686518){\makebox(0,0)[t]{\lineheight{1.25}\smash{\begin{tabular}[t]{c}\footnotesize{$1.0$}\end{tabular}}}}%
    \put(0.66514207,0.34686518){\makebox(0,0)[t]{\lineheight{1.25}\smash{\begin{tabular}[t]{c}\footnotesize{$0.2$}\end{tabular}}}}%
    \put(0.81265033,0.34686518){\makebox(0,0)[t]{\lineheight{1.25}\smash{\begin{tabular}[t]{c}\footnotesize{$0.6$}\end{tabular}}}}%
    \put(0.88665854,0.34686518){\makebox(0,0)[t]{\lineheight{1.25}\smash{\begin{tabular}[t]{c}\footnotesize{$0.8$}\end{tabular}}}}%
    \put(0.73892632,0.34686518){\makebox(0,0)[t]{\lineheight{1.25}\smash{\begin{tabular}[t]{c}\footnotesize{$0.4$}\end{tabular}}}}%
    \put(0.29368297,0.3296431){\color[rgb]{0,0,0}\makebox(0,0)[t]{\lineheight{1.25}\smash{\begin{tabular}[t]{c}\footnotesize{time [-]}\end{tabular}}}}%
    \put(0.77660176,0.3296431){\color[rgb]{0,0,0}\makebox(0,0)[t]{\lineheight{1.25}\smash{\begin{tabular}[t]{c}\footnotesize{time [-]}\end{tabular}}}}%
    \put(0.11001963,0.6866489){\makebox(0,0)[t]{\lineheight{1.25}\smash{\begin{tabular}[t]{c}\footnotesize{$0$}\end{tabular}}}}%
    \put(0.47782808,0.6866489){\makebox(0,0)[t]{\lineheight{1.25}\smash{\begin{tabular}[t]{c}\footnotesize{$1.0$}\end{tabular}}}}%
    \put(0.18281202,0.6866489){\makebox(0,0)[t]{\lineheight{1.25}\smash{\begin{tabular}[t]{c}\footnotesize{$0.2$}\end{tabular}}}}%
    \put(0.33032019,0.6866489){\makebox(0,0)[t]{\lineheight{1.25}\smash{\begin{tabular}[t]{c}\footnotesize{$0.6$}\end{tabular}}}}%
    \put(0.40432849,0.6866489){\makebox(0,0)[t]{\lineheight{1.25}\smash{\begin{tabular}[t]{c}\footnotesize{$0.8$}\end{tabular}}}}%
    \put(0.25659627,0.6866489){\makebox(0,0)[t]{\lineheight{1.25}\smash{\begin{tabular}[t]{c}\footnotesize{$0.4$}\end{tabular}}}}%
    \put(0.59234667,0.6866489){\makebox(0,0)[t]{\lineheight{1.25}\smash{\begin{tabular}[t]{c}\footnotesize{$0$}\end{tabular}}}}%
    \put(0.96015503,0.6866489){\makebox(0,0)[t]{\lineheight{1.25}\smash{\begin{tabular}[t]{c}\footnotesize{$1.0$}\end{tabular}}}}%
    \put(0.66513897,0.6866489){\makebox(0,0)[t]{\lineheight{1.25}\smash{\begin{tabular}[t]{c}\footnotesize{$0.2$}\end{tabular}}}}%
    \put(0.81264713,0.6866489){\makebox(0,0)[t]{\lineheight{1.25}\smash{\begin{tabular}[t]{c}\footnotesize{$0.6$}\end{tabular}}}}%
    \put(0.88665543,0.6866489){\makebox(0,0)[t]{\lineheight{1.25}\smash{\begin{tabular}[t]{c}\footnotesize{$0.8$}\end{tabular}}}}%
    \put(0.73892321,0.6866489){\makebox(0,0)[t]{\lineheight{1.25}\smash{\begin{tabular}[t]{c}\footnotesize{$0.4$}\end{tabular}}}}%
    \put(0.29367986,0.66942683){\color[rgb]{0,0,0}\makebox(0,0)[t]{\lineheight{1.25}\smash{\begin{tabular}[t]{c}\footnotesize{time [-]}\end{tabular}}}}%
    \put(0.77659866,0.66942683){\color[rgb]{0,0,0}\makebox(0,0)[t]{\lineheight{1.25}\smash{\begin{tabular}[t]{c}\footnotesize{time [-]}\end{tabular}}}}%
    \put(0.11001963,1.00794536){\makebox(0,0)[t]{\lineheight{1.25}\smash{\begin{tabular}[t]{c}\footnotesize{$0$}\end{tabular}}}}%
    \put(0.47782799,1.00794536){\makebox(0,0)[t]{\lineheight{1.25}\smash{\begin{tabular}[t]{c}\footnotesize{$1.0$}\end{tabular}}}}%
    \put(0.18281193,1.00794536){\makebox(0,0)[t]{\lineheight{1.25}\smash{\begin{tabular}[t]{c}\footnotesize{$0.2$}\end{tabular}}}}%
    \put(0.3303201,1.00794536){\makebox(0,0)[t]{\lineheight{1.25}\smash{\begin{tabular}[t]{c}\footnotesize{$0.6$}\end{tabular}}}}%
    \put(0.4043284,1.00794536){\makebox(0,0)[t]{\lineheight{1.25}\smash{\begin{tabular}[t]{c}\footnotesize{$0.8$}\end{tabular}}}}%
    \put(0.25659618,1.00794536){\makebox(0,0)[t]{\lineheight{1.25}\smash{\begin{tabular}[t]{c}\footnotesize{$0.4$}\end{tabular}}}}%
    \put(0.59234658,1.00794536){\makebox(0,0)[t]{\lineheight{1.25}\smash{\begin{tabular}[t]{c}\footnotesize{$0$}\end{tabular}}}}%
    \put(0.96015503,1.00794536){\makebox(0,0)[t]{\lineheight{1.25}\smash{\begin{tabular}[t]{c}\footnotesize{$1.0$}\end{tabular}}}}%
    \put(0.66513888,1.00794536){\makebox(0,0)[t]{\lineheight{1.25}\smash{\begin{tabular}[t]{c}\footnotesize{$0.2$}\end{tabular}}}}%
    \put(0.81264704,1.00794536){\makebox(0,0)[t]{\lineheight{1.25}\smash{\begin{tabular}[t]{c}\footnotesize{$0.6$}\end{tabular}}}}%
    \put(0.88665534,1.00794536){\makebox(0,0)[t]{\lineheight{1.25}\smash{\begin{tabular}[t]{c}\footnotesize{$0.8$}\end{tabular}}}}%
    \put(0.73892312,1.00794536){\makebox(0,0)[t]{\lineheight{1.25}\smash{\begin{tabular}[t]{c}\footnotesize{$0.4$}\end{tabular}}}}%
    \put(0.29367977,0.99072328){\color[rgb]{0,0,0}\makebox(0,0)[t]{\lineheight{1.25}\smash{\begin{tabular}[t]{c}\footnotesize{time [-]}\end{tabular}}}}%
    \put(0.77659857,0.99072328){\color[rgb]{0,0,0}\makebox(0,0)[t]{\lineheight{1.25}\smash{\begin{tabular}[t]{c}\footnotesize{time [-]}\end{tabular}}}}%
    \put(0.02578369,0.17074545){\color[rgb]{0,0,0}\rotatebox{90}{\makebox(0,0)[t]{\lineheight{1.25}\smash{\begin{tabular}[t]{c}$e$ [-]\end{tabular}}}}}%
    \put(0.02572186,0.51014824){\color[rgb]{0,0,0}\rotatebox{90}{\makebox(0,0)[t]{\lineheight{1.25}\smash{\begin{tabular}[t]{c}$q$ [-]\end{tabular}}}}}%
    \put(0.51427783,0.16998577){\color[rgb]{0,0,0}\rotatebox{90}{\makebox(0,0)[t]{\lineheight{1.25}\smash{\begin{tabular}[t]{c}$\tau$ [-]\end{tabular}}}}}%
    \put(0.52427783,0.5099805){\color[rgb]{0,0,0}\rotatebox{90}{\makebox(0,0)[t]{\lineheight{1.25}\smash{\begin{tabular}[t]{c}$v$ [-]\end{tabular}}}}}%
    \put(0.51427473,0.83116607){\color[rgb]{0,0,0}\rotatebox{90}{\makebox(0,0)[t]{\lineheight{1.25}\smash{\begin{tabular}[t]{c}$\tau$ [-]\end{tabular}}}}}%
    \put(0.02578059,0.83192575){\color[rgb]{0,0,0}\rotatebox{90}{\makebox(0,0)[t]{\lineheight{1.25}\smash{\begin{tabular}[t]{c}$e$ [-]\end{tabular}}}}}%
    \put(0.02571876,1.17132836){\color[rgb]{0,0,0}\rotatebox{90}{\makebox(0,0)[t]{\lineheight{1.25}\smash{\begin{tabular}[t]{c}$q$ [-]\end{tabular}}}}}%
    \put(0.52427473,1.17116062){\color[rgb]{0,0,0}\rotatebox{90}{\makebox(0,0)[t]{\lineheight{1.25}\smash{\begin{tabular}[t]{c}$v$ [-]\end{tabular}}}}}%
    \put(0.00323983,0.64621418){\color[rgb]{0,0,0}\makebox(0,0)[lt]{\lineheight{1.25}\smash{\begin{tabular}[t]{l}(b)\end{tabular}}}}%
    \put(0.00313706,1.30729427){\color[rgb]{0,0,0}\makebox(0,0)[lt]{\lineheight{1.25}\smash{\begin{tabular}[t]{l}(a)\end{tabular}}}}%
    \put(0.08591666,0.91813231){\makebox(0,0)[rt]{\lineheight{1.25}\smash{\begin{tabular}[t]{r}\footnotesize{$ 0.30$}\end{tabular}}}}%
    \put(0.08591666,0.95134933){\makebox(0,0)[rt]{\lineheight{1.25}\smash{\begin{tabular}[t]{r}\footnotesize{$ 0.35$}\end{tabular}}}}%
    \put(0.08591666,0.8849153){\makebox(0,0)[rt]{\lineheight{1.25}\smash{\begin{tabular}[t]{r}\footnotesize{$ 0.25$}\end{tabular}}}}%
    \put(0.08591666,0.85263557){\makebox(0,0)[rt]{\lineheight{1.25}\smash{\begin{tabular}[t]{r}\footnotesize{$ 0.20$}\end{tabular}}}}%
    \put(0.08591666,0.81738611){\makebox(0,0)[rt]{\lineheight{1.25}\smash{\begin{tabular}[t]{r}\footnotesize{$ 0.15$}\end{tabular}}}}%
    \put(0.08591666,0.78420013){\makebox(0,0)[rt]{\lineheight{1.25}\smash{\begin{tabular}[t]{r}\footnotesize{$ 0.10$}\end{tabular}}}}%
    \put(0.08591666,0.75101433){\makebox(0,0)[rt]{\lineheight{1.25}\smash{\begin{tabular}[t]{r}\footnotesize{$ 0.05$}\end{tabular}}}}%
    \put(0.08908161,0.71726562){\makebox(0,0)[rt]{\lineheight{1.25}\smash{\begin{tabular}[t]{r}\footnotesize{$ 0.00$}\end{tabular}}}}%
    \put(0.56583675,1.22162014){\makebox(0,0)[rt]{\lineheight{1.25}\smash{\begin{tabular}[t]{r}\footnotesize{$ 0.5$}\end{tabular}}}}%
    \put(0.56583675,1.25800219){\makebox(0,0)[rt]{\lineheight{1.25}\smash{\begin{tabular}[t]{r}\footnotesize{$ 0.6$}\end{tabular}}}}%
    \put(0.56583675,1.18365558){\makebox(0,0)[rt]{\lineheight{1.25}\smash{\begin{tabular}[t]{r}\footnotesize{$ 0.4$}\end{tabular}}}}%
    \put(0.56583675,1.14662829){\makebox(0,0)[rt]{\lineheight{1.25}\smash{\begin{tabular}[t]{r}\footnotesize{$ 0.3$}\end{tabular}}}}%
    \put(0.56583675,1.10979631){\makebox(0,0)[rt]{\lineheight{1.25}\smash{\begin{tabular}[t]{r}\footnotesize{$ 0.2$}\end{tabular}}}}%
    \put(0.56583675,1.07186277){\makebox(0,0)[rt]{\lineheight{1.25}\smash{\begin{tabular}[t]{r}\footnotesize{$ 0.1$}\end{tabular}}}}%
    \put(0.56583675,1.03392942){\makebox(0,0)[rt]{\lineheight{1.25}\smash{\begin{tabular}[t]{r}\footnotesize{$ 0.0$}\end{tabular}}}}%
    \put(0.56583985,0.56048073){\makebox(0,0)[rt]{\lineheight{1.25}\smash{\begin{tabular}[t]{r}\footnotesize{$ 0.5$}\end{tabular}}}}%
    \put(0.56583985,0.59686278){\makebox(0,0)[rt]{\lineheight{1.25}\smash{\begin{tabular}[t]{r}\footnotesize{$ 0.6$}\end{tabular}}}}%
    \put(0.56583985,0.52251616){\makebox(0,0)[rt]{\lineheight{1.25}\smash{\begin{tabular}[t]{r}\footnotesize{$ 0.4$}\end{tabular}}}}%
    \put(0.56583985,0.48548888){\makebox(0,0)[rt]{\lineheight{1.25}\smash{\begin{tabular}[t]{r}\footnotesize{$ 0.3$}\end{tabular}}}}%
    \put(0.56583985,0.4486569){\makebox(0,0)[rt]{\lineheight{1.25}\smash{\begin{tabular}[t]{r}\footnotesize{$ 0.2$}\end{tabular}}}}%
    \put(0.56583985,0.41072336){\makebox(0,0)[rt]{\lineheight{1.25}\smash{\begin{tabular}[t]{r}\footnotesize{$ 0.1$}\end{tabular}}}}%
    \put(0.56583985,0.37279001){\makebox(0,0)[rt]{\lineheight{1.25}\smash{\begin{tabular}[t]{r}\footnotesize{$ 0.0$}\end{tabular}}}}%
    \put(0.56583675,0.91851544){\makebox(0,0)[rt]{\lineheight{1.25}\smash{\begin{tabular}[t]{r}\footnotesize{$-0.1$}\end{tabular}}}}%
    \put(0.56583675,0.95806253){\makebox(0,0)[rt]{\lineheight{1.25}\smash{\begin{tabular}[t]{r}\footnotesize{$ 0.0$}\end{tabular}}}}%
    \put(0.56583675,0.87896835){\makebox(0,0)[rt]{\lineheight{1.25}\smash{\begin{tabular}[t]{r}\footnotesize{$-0.2$}\end{tabular}}}}%
    \put(0.56583675,0.83877603){\makebox(0,0)[rt]{\lineheight{1.25}\smash{\begin{tabular}[t]{r}\footnotesize{$-0.3$}\end{tabular}}}}%
    \put(0.56583675,0.7971965){\makebox(0,0)[rt]{\lineheight{1.25}\smash{\begin{tabular}[t]{r}\footnotesize{$-0.4$}\end{tabular}}}}%
    \put(0.56583675,0.75768044){\makebox(0,0)[rt]{\lineheight{1.25}\smash{\begin{tabular}[t]{r}\footnotesize{$-0.5$}\end{tabular}}}}%
    \put(0.56583675,0.71816457){\makebox(0,0)[rt]{\lineheight{1.25}\smash{\begin{tabular}[t]{r}\footnotesize{$-0.6$}\end{tabular}}}}%
    \put(0.56583985,0.25591927){\makebox(0,0)[rt]{\lineheight{1.25}\smash{\begin{tabular}[t]{r}\footnotesize{$-0.1$}\end{tabular}}}}%
    \put(0.56583985,0.29546636){\makebox(0,0)[rt]{\lineheight{1.25}\smash{\begin{tabular}[t]{r}\footnotesize{$ 0.0$}\end{tabular}}}}%
    \put(0.56583985,0.21637219){\makebox(0,0)[rt]{\lineheight{1.25}\smash{\begin{tabular}[t]{r}\footnotesize{$-0.2$}\end{tabular}}}}%
    \put(0.56583985,0.17617986){\makebox(0,0)[rt]{\lineheight{1.25}\smash{\begin{tabular}[t]{r}\footnotesize{$-0.3$}\end{tabular}}}}%
    \put(0.56583985,0.13460015){\makebox(0,0)[rt]{\lineheight{1.25}\smash{\begin{tabular}[t]{r}\footnotesize{$-0.4$}\end{tabular}}}}%
    \put(0.56583985,0.09508409){\makebox(0,0)[rt]{\lineheight{1.25}\smash{\begin{tabular}[t]{r}\footnotesize{$-0.5$}\end{tabular}}}}%
    \put(0.56583985,0.05556822){\makebox(0,0)[rt]{\lineheight{1.25}\smash{\begin{tabular}[t]{r}\footnotesize{$-0.6$}\end{tabular}}}}%
    \put(0.08591977,0.57817939){\makebox(0,0)[rt]{\lineheight{1.25}\smash{\begin{tabular}[t]{r}\footnotesize{$ 0.4$}\end{tabular}}}}%
    \put(0.08591977,0.52692531){\makebox(0,0)[rt]{\lineheight{1.25}\smash{\begin{tabular}[t]{r}\footnotesize{$ 0.3$}\end{tabular}}}}%
    \put(0.08591977,0.47426815){\makebox(0,0)[rt]{\lineheight{1.25}\smash{\begin{tabular}[t]{r}\footnotesize{$ 0.2$}\end{tabular}}}}%
    \put(0.08591977,0.42372155){\makebox(0,0)[rt]{\lineheight{1.25}\smash{\begin{tabular}[t]{r}\footnotesize{$ 0.1$}\end{tabular}}}}%
    \put(0.08591977,0.37256513){\makebox(0,0)[rt]{\lineheight{1.25}\smash{\begin{tabular}[t]{r}\footnotesize{$ 0.0$}\end{tabular}}}}%
    \put(0.08591666,1.23975915){\makebox(0,0)[rt]{\lineheight{1.25}\smash{\begin{tabular}[t]{r}\footnotesize{$ 0.4$}\end{tabular}}}}%
    \put(0.08591666,1.18850507){\makebox(0,0)[rt]{\lineheight{1.25}\smash{\begin{tabular}[t]{r}\footnotesize{$ 0.3$}\end{tabular}}}}%
    \put(0.08591666,1.13584791){\makebox(0,0)[rt]{\lineheight{1.25}\smash{\begin{tabular}[t]{r}\footnotesize{$ 0.2$}\end{tabular}}}}%
    \put(0.08591666,1.08530131){\makebox(0,0)[rt]{\lineheight{1.25}\smash{\begin{tabular}[t]{r}\footnotesize{$ 0.1$}\end{tabular}}}}%
    \put(0.08591666,1.03414489){\makebox(0,0)[rt]{\lineheight{1.25}\smash{\begin{tabular}[t]{r}\footnotesize{$ 0.0$}\end{tabular}}}}%
    \put(0.51916203,0.64629924){\color[rgb]{0,0,0}\makebox(0,0)[t]{\lineheight{1.25}\smash{\begin{tabular}[t]{c}Single generator\end{tabular}}}}%
    \put(0.52014814,1.30435027){\color[rgb]{0,0,0}\makebox(0,0)[t]{\lineheight{1.25}\smash{\begin{tabular}[t]{c}GENERIC\end{tabular}}}}%
  \end{picture}%
\endgroup%

%% file: VE_BoxPlots.eps_tex
\begingroup%
  \makeatletter%
  \providecommand\color[2][]{%
    \errmessage{(Inkscape) Color is used for the text in Inkscape, but the package 'color.sty' is not loaded}%
    \renewcommand\color[2][]{}%
  }%
  \providecommand\transparent[1]{%
    \errmessage{(Inkscape) Transparency is used (non-zero) for the text in Inkscape, but the package 'transparent.sty' is not loaded}%
    \renewcommand\transparent[1]{}%
  }%
  \providecommand\rotatebox[2]{#2}%
  \newcommand*\fsize{\dimexpr\f@size pt\relax}%
  \newcommand*\lineheight[1]{\fontsize{\fsize}{#1\fsize}\selectfont}%
  \ifx\svgwidth\undefined%
    \setlength{\unitlength}{929.41390636bp}%
    \ifx\svgscale\undefined%
      \relax%
    \else%
      \setlength{\unitlength}{\unitlength * \real{\svgscale}}%
    \fi%
  \else%
    \setlength{\unitlength}{\svgwidth}%
  \fi%
  \global\let\svgwidth\undefined%
  \global\let\svgscale\undefined%
  \makeatother%
  \begin{picture}(1,0.3722776)%
    \lineheight{1}%
    \setlength\tabcolsep{0pt}%
    \put(0,0){\includegraphics[width=\unitlength]{VE_BoxPlots.eps}}%
    \put(0.01767005,0.35601269){\color[rgb]{0,0,0}\makebox(0,0)[lt]{\lineheight{1.25}\smash{\begin{tabular}[t]{l}(a)\end{tabular}}}}%
    \put(0.52973118,0.17104668){\rotatebox{90}{\makebox(0,0)[lt]{\lineheight{1.25}\smash{\begin{tabular}[t]{l}MSE\end{tabular}}}}}%
    \put(0.55404792,0.2238325){\makebox(0,0)[t]{\lineheight{1.25}\smash{\begin{tabular}[t]{c}\footnotesize{$10^{-4}$}\end{tabular}}}}%
    \put(0.55404792,0.11123552){\makebox(0,0)[t]{\lineheight{1.25}\smash{\begin{tabular}[t]{c}\footnotesize{$10^{-5}$}\end{tabular}}}}%
    \put(0.02808412,0.17060564){\rotatebox{90}{\makebox(0,0)[lt]{\lineheight{1.25}\smash{\begin{tabular}[t]{l}MSE\end{tabular}}}}}%
    \put(0.06524557,0.31927964){\makebox(0,0)[t]{\lineheight{1.25}\smash{\begin{tabular}[t]{c}\footnotesize{$10^{-3}$}\end{tabular}}}}%
    \put(0.06524557,0.19957087){\makebox(0,0)[t]{\lineheight{1.25}\smash{\begin{tabular}[t]{c}\footnotesize{$10^{-4}$}\end{tabular}}}}%
    \put(0.06524557,0.08211434){\makebox(0,0)[t]{\lineheight{1.25}\smash{\begin{tabular}[t]{c}\footnotesize{$10^{-5}$}\end{tabular}}}}%
    \put(0.19207839,0.00239295){\color[rgb]{0,0,0}\makebox(0,0)[t]{\lineheight{1.25}\smash{\begin{tabular}[t]{c}GENERIC\end{tabular}}}}%
    \put(0.38412143,0.00322985){\color[rgb]{0,0,0}\makebox(0,0)[t]{\lineheight{1.25}\smash{\begin{tabular}[t]{c}Single generator\end{tabular}}}}%
    \put(0.8307346,0.02357837){\makebox(0,0)[t]{\lineheight{1.25}\smash{\begin{tabular}[t]{c}\footnotesize{train}\end{tabular}}}}%
    \put(0.91388378,0.02357837){\makebox(0,0)[t]{\lineheight{1.25}\smash{\begin{tabular}[t]{c}\footnotesize{test}\end{tabular}}}}%
    \put(0.63861353,0.02357837){\makebox(0,0)[t]{\lineheight{1.25}\smash{\begin{tabular}[t]{c}\footnotesize{train}\end{tabular}}}}%
    \put(0.7217627,0.02357837){\makebox(0,0)[t]{\lineheight{1.25}\smash{\begin{tabular}[t]{c}\footnotesize{test}\end{tabular}}}}%
    \put(0.67980512,0.00239296){\color[rgb]{0,0,0}\makebox(0,0)[t]{\lineheight{1.25}\smash{\begin{tabular}[t]{c}GENERIC\end{tabular}}}}%
    \put(0.87184815,0.00322987){\color[rgb]{0,0,0}\makebox(0,0)[t]{\lineheight{1.25}\smash{\begin{tabular}[t]{c}Single generator\end{tabular}}}}%
    \put(0.15250073,0.02357837){\makebox(0,0)[t]{\lineheight{1.25}\smash{\begin{tabular}[t]{c}\footnotesize{train}\end{tabular}}}}%
    \put(0.2356499,0.02357837){\makebox(0,0)[t]{\lineheight{1.25}\smash{\begin{tabular}[t]{c}\footnotesize{test}\end{tabular}}}}%
    \put(0.34462178,0.02357837){\makebox(0,0)[t]{\lineheight{1.25}\smash{\begin{tabular}[t]{c}\footnotesize{train}\end{tabular}}}}%
    \put(0.42777098,0.02357837){\makebox(0,0)[t]{\lineheight{1.25}\smash{\begin{tabular}[t]{c}\footnotesize{test}\end{tabular}}}}%
    \put(0.51797908,0.35601271){\color[rgb]{0,0,0}\makebox(0,0)[lt]{\lineheight{1.25}\smash{\begin{tabular}[t]{l}(b)\end{tabular}}}}%
  \end{picture}%
\endgroup%

%% file: DP_Ndata.eps_tex
\begingroup%
  \makeatletter%
  \providecommand\color[2][]{%
    \errmessage{(Inkscape) Color is used for the text in Inkscape, but the package 'color.sty' is not loaded}%
    \renewcommand\color[2][]{}%
  }%
  \providecommand\transparent[1]{%
    \errmessage{(Inkscape) Transparency is used (non-zero) for the text in Inkscape, but the package 'transparent.sty' is not loaded}%
    \renewcommand\transparent[1]{}%
  }%
  \providecommand\rotatebox[2]{#2}%
  \newcommand*\fsize{\dimexpr\f@size pt\relax}%
  \newcommand*\lineheight[1]{\fontsize{\fsize}{#1\fsize}\selectfont}%
  \ifx\svgwidth\undefined%
    \setlength{\unitlength}{921.6bp}%
    \ifx\svgscale\undefined%
      \relax%
    \else%
      \setlength{\unitlength}{\unitlength * \real{\svgscale}}%
    \fi%
  \else%
    \setlength{\unitlength}{\svgwidth}%
  \fi%
  \global\let\svgwidth\undefined%
  \global\let\svgscale\undefined%
  \makeatother%
  \begin{picture}(1,0.37500001)%
    \lineheight{1}%
    \setlength\tabcolsep{0pt}%
    \put(0,0){\includegraphics[width=\unitlength]{DP_Ndata.eps}}%
    \put(0.00922151,0.35304661){\color[rgb]{0,0,0}\makebox(0,0)[lt]{\lineheight{1.25}\smash{\begin{tabular}[t]{l}(a)\end{tabular}}}}%
    \put(0.50880983,0.35304661){\color[rgb]{0,0,0}\makebox(0,0)[lt]{\lineheight{1.25}\smash{\begin{tabular}[t]{l}(b)\end{tabular}}}}%
    \put(0.18690366,0.00446483){\color[rgb]{0,0,0}\makebox(0,0)[t]{\lineheight{1.25}\smash{\begin{tabular}[t]{c}GENERIC\end{tabular}}}}%
    \put(0.3805749,0.00530883){\color[rgb]{0,0,0}\makebox(0,0)[t]{\lineheight{1.25}\smash{\begin{tabular}[t]{c}Single generator\end{tabular}}}}%
    \put(0.01984355,0.17951299){\rotatebox{90}{\makebox(0,0)[lt]{\lineheight{1.25}\smash{\begin{tabular}[t]{l}MSE\end{tabular}}}}}%
    \put(0.0573201,0.32781994){\makebox(0,0)[t]{\lineheight{1.25}\smash{\begin{tabular}[t]{c}\footnotesize{$10^{-1}$}\end{tabular}}}}%
    \put(0.0573201,0.24778632){\makebox(0,0)[t]{\lineheight{1.25}\smash{\begin{tabular}[t]{c}\footnotesize{$10^{-2}$}\end{tabular}}}}%
    \put(0.0573201,0.08701614){\makebox(0,0)[t]{\lineheight{1.25}\smash{\begin{tabular}[t]{c}\footnotesize{$10^{-4}$}\end{tabular}}}}%
    \put(0.12293909,0.02608109){\makebox(0,0)[t]{\lineheight{1.25}\smash{\begin{tabular}[t]{c}\footnotesize{$300$}\end{tabular}}}}%
    \put(0.18752244,0.02608109){\makebox(0,0)[t]{\lineheight{1.25}\smash{\begin{tabular}[t]{c}\footnotesize{$200$}\end{tabular}}}}%
    \put(0.25210577,0.02608109){\makebox(0,0)[t]{\lineheight{1.25}\smash{\begin{tabular}[t]{c}\footnotesize{$50$}\end{tabular}}}}%
    \put(0.3166891,0.02608109){\makebox(0,0)[t]{\lineheight{1.25}\smash{\begin{tabular}[t]{c}\footnotesize{$300$}\end{tabular}}}}%
    \put(0.38127243,0.02608109){\makebox(0,0)[t]{\lineheight{1.25}\smash{\begin{tabular}[t]{c}\footnotesize{$200$}\end{tabular}}}}%
    \put(0.44585577,0.02608109){\makebox(0,0)[t]{\lineheight{1.25}\smash{\begin{tabular}[t]{c}\footnotesize{$50$}\end{tabular}}}}%
    \put(0.0573201,0.16759636){\makebox(0,0)[t]{\lineheight{1.25}\smash{\begin{tabular}[t]{c}\footnotesize{$10^{-3}$}\end{tabular}}}}%
    \put(0.68690364,0.00696051){\color[rgb]{0,0,0}\makebox(0,0)[t]{\lineheight{1.25}\smash{\begin{tabular}[t]{c}GENERIC\end{tabular}}}}%
    \put(0.8805749,0.00780451){\color[rgb]{0,0,0}\makebox(0,0)[t]{\lineheight{1.25}\smash{\begin{tabular}[t]{c}Single generator\end{tabular}}}}%
    \put(0.51984353,0.18200863){\rotatebox{90}{\makebox(0,0)[lt]{\lineheight{1.25}\smash{\begin{tabular}[t]{l}MSE\end{tabular}}}}}%
    \put(0.55732007,0.32877303){\makebox(0,0)[t]{\lineheight{1.25}\smash{\begin{tabular}[t]{c}\footnotesize{$10^{-1}$}\end{tabular}}}}%
    \put(0.55732007,0.2649304){\makebox(0,0)[t]{\lineheight{1.25}\smash{\begin{tabular}[t]{c}\footnotesize{$10^{-2}$}\end{tabular}}}}%
    \put(0.55732007,0.13508469){\makebox(0,0)[t]{\lineheight{1.25}\smash{\begin{tabular}[t]{c}\footnotesize{$10^{-4}$}\end{tabular}}}}%
    \put(0.55732007,0.06996456){\makebox(0,0)[t]{\lineheight{1.25}\smash{\begin{tabular}[t]{c}\footnotesize{$10^{-5}$}\end{tabular}}}}%
    \put(0.62293908,0.02857675){\makebox(0,0)[t]{\lineheight{1.25}\smash{\begin{tabular}[t]{c}\footnotesize{$400$}\end{tabular}}}}%
    \put(0.68752242,0.02857675){\makebox(0,0)[t]{\lineheight{1.25}\smash{\begin{tabular}[t]{c}\footnotesize{$200$}\end{tabular}}}}%
    \put(0.75210575,0.02857675){\makebox(0,0)[t]{\lineheight{1.25}\smash{\begin{tabular}[t]{c}\footnotesize{$50$}\end{tabular}}}}%
    \put(0.81668908,0.02857675){\makebox(0,0)[t]{\lineheight{1.25}\smash{\begin{tabular}[t]{c}\footnotesize{$400$}\end{tabular}}}}%
    \put(0.88127243,0.02857675){\makebox(0,0)[t]{\lineheight{1.25}\smash{\begin{tabular}[t]{c}\footnotesize{$200$}\end{tabular}}}}%
    \put(0.94585578,0.02857675){\makebox(0,0)[t]{\lineheight{1.25}\smash{\begin{tabular}[t]{c}\footnotesize{$50$}\end{tabular}}}}%
    \put(0.55732007,0.19938891){\makebox(0,0)[t]{\lineheight{1.25}\smash{\begin{tabular}[t]{c}\footnotesize{$10^{-3}$}\end{tabular}}}}%
  \end{picture}%
\endgroup%

%% file: VE_Ndata.eps_tex
\begingroup%
  \makeatletter%
  \providecommand\color[2][]{%
    \errmessage{(Inkscape) Color is used for the text in Inkscape, but the package 'color.sty' is not loaded}%
    \renewcommand\color[2][]{}%
  }%
  \providecommand\transparent[1]{%
    \errmessage{(Inkscape) Transparency is used (non-zero) for the text in Inkscape, but the package 'transparent.sty' is not loaded}%
    \renewcommand\transparent[1]{}%
  }%
  \providecommand\rotatebox[2]{#2}%
  \newcommand*\fsize{\dimexpr\f@size pt\relax}%
  \newcommand*\lineheight[1]{\fontsize{\fsize}{#1\fsize}\selectfont}%
  \ifx\svgwidth\undefined%
    \setlength{\unitlength}{921.6bp}%
    \ifx\svgscale\undefined%
      \relax%
    \else%
      \setlength{\unitlength}{\unitlength * \real{\svgscale}}%
    \fi%
  \else%
    \setlength{\unitlength}{\svgwidth}%
  \fi%
  \global\let\svgwidth\undefined%
  \global\let\svgscale\undefined%
  \makeatother%
  \begin{picture}(1,0.375)%
    \lineheight{1}%
    \setlength\tabcolsep{0pt}%
    \put(0,0){\includegraphics[width=\unitlength]{VE_Ndata.eps}}%
    \put(0.18690366,0.00609245){\color[rgb]{0,0,0}\makebox(0,0)[t]{\lineheight{1.25}\smash{\begin{tabular}[t]{c}GENERIC\end{tabular}}}}%
    \put(0.38057485,0.00693645){\color[rgb]{0,0,0}\makebox(0,0)[t]{\lineheight{1.25}\smash{\begin{tabular}[t]{c}Single generator\end{tabular}}}}%
    \put(0.01984352,0.18114055){\rotatebox{90}{\makebox(0,0)[lt]{\lineheight{1.25}\smash{\begin{tabular}[t]{l}MSE\end{tabular}}}}}%
    \put(0.05732007,0.30800764){\makebox(0,0)[t]{\lineheight{1.25}\smash{\begin{tabular}[t]{c}\footnotesize{$10^{-3}$}\end{tabular}}}}%
    \put(0.05732007,0.08951588){\makebox(0,0)[t]{\lineheight{1.25}\smash{\begin{tabular}[t]{c}\footnotesize{$10^{-5}$}\end{tabular}}}}%
    \put(0.12293905,0.02770871){\makebox(0,0)[t]{\lineheight{1.25}\smash{\begin{tabular}[t]{c}\footnotesize{$400$}\end{tabular}}}}%
    \put(0.1875224,0.02770871){\makebox(0,0)[t]{\lineheight{1.25}\smash{\begin{tabular}[t]{c}\footnotesize{$150$}\end{tabular}}}}%
    \put(0.25210578,0.02770871){\makebox(0,0)[t]{\lineheight{1.25}\smash{\begin{tabular}[t]{c}\footnotesize{$50$}\end{tabular}}}}%
    \put(0.3166891,0.02770871){\makebox(0,0)[t]{\lineheight{1.25}\smash{\begin{tabular}[t]{c}\footnotesize{$400$}\end{tabular}}}}%
    \put(0.38127245,0.02770871){\makebox(0,0)[t]{\lineheight{1.25}\smash{\begin{tabular}[t]{c}\footnotesize{$150$}\end{tabular}}}}%
    \put(0.44585576,0.02770871){\makebox(0,0)[t]{\lineheight{1.25}\smash{\begin{tabular}[t]{c}\footnotesize{$50$}\end{tabular}}}}%
    \put(0.05732007,0.19886339){\makebox(0,0)[t]{\lineheight{1.25}\smash{\begin{tabular}[t]{c}\footnotesize{$10^{-4}$}\end{tabular}}}}%
    \put(0.68690364,0.00609242){\color[rgb]{0,0,0}\makebox(0,0)[t]{\lineheight{1.25}\smash{\begin{tabular}[t]{c}GENERIC\end{tabular}}}}%
    \put(0.88057486,0.00693642){\color[rgb]{0,0,0}\makebox(0,0)[t]{\lineheight{1.25}\smash{\begin{tabular}[t]{c}Single generator\end{tabular}}}}%
    \put(0.51984353,0.18114055){\rotatebox{90}{\makebox(0,0)[lt]{\lineheight{1.25}\smash{\begin{tabular}[t]{l}MSE\end{tabular}}}}}%
    \put(0.55732008,0.2933592){\makebox(0,0)[t]{\lineheight{1.25}\smash{\begin{tabular}[t]{c}\footnotesize{$10^{-3}$}\end{tabular}}}}%
    \put(0.55732008,0.16351349){\makebox(0,0)[t]{\lineheight{1.25}\smash{\begin{tabular}[t]{c}\footnotesize{$10^{-5}$}\end{tabular}}}}%
    \put(0.55732008,0.10002098){\makebox(0,0)[t]{\lineheight{1.25}\smash{\begin{tabular}[t]{c}\footnotesize{$10^{-6}$}\end{tabular}}}}%
    \put(0.62293909,0.02770868){\makebox(0,0)[t]{\lineheight{1.25}\smash{\begin{tabular}[t]{c}\footnotesize{$400$}\end{tabular}}}}%
    \put(0.68752242,0.02770868){\makebox(0,0)[t]{\lineheight{1.25}\smash{\begin{tabular}[t]{c}\footnotesize{$200$}\end{tabular}}}}%
    \put(0.75210577,0.02770868){\makebox(0,0)[t]{\lineheight{1.25}\smash{\begin{tabular}[t]{c}\footnotesize{$50$}\end{tabular}}}}%
    \put(0.81668908,0.02770868){\makebox(0,0)[t]{\lineheight{1.25}\smash{\begin{tabular}[t]{c}\footnotesize{$400$}\end{tabular}}}}%
    \put(0.88127243,0.02770868){\makebox(0,0)[t]{\lineheight{1.25}\smash{\begin{tabular}[t]{c}\footnotesize{$200$}\end{tabular}}}}%
    \put(0.94585574,0.02770868){\makebox(0,0)[t]{\lineheight{1.25}\smash{\begin{tabular}[t]{c}\footnotesize{$50$}\end{tabular}}}}%
    \put(0.55732008,0.22816027){\makebox(0,0)[t]{\lineheight{1.25}\smash{\begin{tabular}[t]{c}\footnotesize{$10^{-4}$}\end{tabular}}}}%
    \put(0.00922151,0.3530466){\color[rgb]{0,0,0}\makebox(0,0)[lt]{\lineheight{1.25}\smash{\begin{tabular}[t]{l}(a)\end{tabular}}}}%
    \put(0.50880983,0.3530466){\color[rgb]{0,0,0}\makebox(0,0)[lt]{\lineheight{1.25}\smash{\begin{tabular}[t]{l}(b)\end{tabular}}}}%
  \end{picture}%
\endgroup%

%% file: DP_epoch.eps_tex
\begingroup%
  \makeatletter%
  \providecommand\color[2][]{%
    \errmessage{(Inkscape) Color is used for the text in Inkscape, but the package 'color.sty' is not loaded}%
    \renewcommand\color[2][]{}%
  }%
  \providecommand\transparent[1]{%
    \errmessage{(Inkscape) Transparency is used (non-zero) for the text in Inkscape, but the package 'transparent.sty' is not loaded}%
    \renewcommand\transparent[1]{}%
  }%
  \providecommand\rotatebox[2]{#2}%
  \newcommand*\fsize{\dimexpr\f@size pt\relax}%
  \newcommand*\lineheight[1]{\fontsize{\fsize}{#1\fsize}\selectfont}%
  \ifx\svgwidth\undefined%
    \setlength{\unitlength}{460.8bp}%
    \ifx\svgscale\undefined%
      \relax%
    \else%
      \setlength{\unitlength}{\unitlength * \real{\svgscale}}%
    \fi%
  \else%
    \setlength{\unitlength}{\svgwidth}%
  \fi%
  \global\let\svgwidth\undefined%
  \global\let\svgscale\undefined%
  \makeatother%
  \begin{picture}(1,0.75)%
    \lineheight{1}%
    \setlength\tabcolsep{0pt}%
    \put(0,0){\includegraphics[width=\unitlength]{DP_epoch.eps}}%
    \put(0.37380734,0.00768839){\color[rgb]{0,0,0}\makebox(0,0)[t]{\lineheight{1.25}\smash{\begin{tabular}[t]{c}GENERIC\end{tabular}}}}%
    \put(0.76114982,0.0093764){\color[rgb]{0,0,0}\makebox(0,0)[t]{\lineheight{1.25}\smash{\begin{tabular}[t]{c}Single generator\end{tabular}}}}%
    \put(0.03968711,0.35778484){\rotatebox{90}{\makebox(0,0)[lt]{\lineheight{1.25}\smash{\begin{tabular}[t]{l}MSE\end{tabular}}}}}%
    \put(0.1146402,0.61533616){\makebox(0,0)[t]{\lineheight{1.25}\smash{\begin{tabular}[t]{c}\footnotesize{$10^{-1}$}\end{tabular}}}}%
    \put(0.1146402,0.46177939){\makebox(0,0)[t]{\lineheight{1.25}\smash{\begin{tabular}[t]{c}\footnotesize{$10^{-2}$}\end{tabular}}}}%
    \put(0.1146402,0.15325986){\makebox(0,0)[t]{\lineheight{1.25}\smash{\begin{tabular}[t]{c}\footnotesize{$10^{-4}$}\end{tabular}}}}%
    \put(0.24262297,0.05092093){\makebox(0,0)[t]{\lineheight{1.25}\smash{\begin{tabular}[t]{c}\footnotesize{$3000$}\end{tabular}}}}%
    \put(0.37178966,0.05092093){\makebox(0,0)[t]{\lineheight{1.25}\smash{\begin{tabular}[t]{c}\footnotesize{$6000$}\end{tabular}}}}%
    \put(0.50095633,0.05092093){\makebox(0,0)[t]{\lineheight{1.25}\smash{\begin{tabular}[t]{c}\footnotesize{$12000$}\end{tabular}}}}%
    \put(0.63012295,0.05092093){\makebox(0,0)[t]{\lineheight{1.25}\smash{\begin{tabular}[t]{c}\footnotesize{$3000$}\end{tabular}}}}%
    \put(0.75928958,0.05092093){\makebox(0,0)[t]{\lineheight{1.25}\smash{\begin{tabular}[t]{c}\footnotesize{$6000$}\end{tabular}}}}%
    \put(0.88845628,0.05092093){\makebox(0,0)[t]{\lineheight{1.25}\smash{\begin{tabular}[t]{c}\footnotesize{$12000$}\end{tabular}}}}%
    \put(0.1146402,0.30790983){\makebox(0,0)[t]{\lineheight{1.25}\smash{\begin{tabular}[t]{c}\footnotesize{$10^{-3}$}\end{tabular}}}}%
  \end{picture}%
\endgroup%

%% file: DP_lrate.eps_tex
\begingroup%
  \makeatletter%
  \providecommand\color[2][]{%
    \errmessage{(Inkscape) Color is used for the text in Inkscape, but the package 'color.sty' is not loaded}%
    \renewcommand\color[2][]{}%
  }%
  \providecommand\transparent[1]{%
    \errmessage{(Inkscape) Transparency is used (non-zero) for the text in Inkscape, but the package 'transparent.sty' is not loaded}%
    \renewcommand\transparent[1]{}%
  }%
  \providecommand\rotatebox[2]{#2}%
  \newcommand*\fsize{\dimexpr\f@size pt\relax}%
  \newcommand*\lineheight[1]{\fontsize{\fsize}{#1\fsize}\selectfont}%
  \ifx\svgwidth\undefined%
    \setlength{\unitlength}{463.25049591bp}%
    \ifx\svgscale\undefined%
      \relax%
    \else%
      \setlength{\unitlength}{\unitlength * \real{\svgscale}}%
    \fi%
  \else%
    \setlength{\unitlength}{\svgwidth}%
  \fi%
  \global\let\svgwidth\undefined%
  \global\let\svgscale\undefined%
  \makeatother%
  \begin{picture}(1,0.74603267)%
    \lineheight{1}%
    \setlength\tabcolsep{0pt}%
    \put(0,0){\includegraphics[width=\unitlength]{DP_lrate.eps}}%
    \put(0.37270243,0.0088824){\color[rgb]{0,0,0}\makebox(0,0)[t]{\lineheight{1.25}\smash{\begin{tabular}[t]{c}GENERIC\end{tabular}}}}%
    \put(0.75799595,0.01056148){\color[rgb]{0,0,0}\makebox(0,0)[t]{\lineheight{1.25}\smash{\begin{tabular}[t]{c}Single generator\end{tabular}}}}%
    \put(0.04034966,0.35712667){\rotatebox{90}{\makebox(0,0)[lt]{\lineheight{1.25}\smash{\begin{tabular}[t]{l}MSE\end{tabular}}}}}%
    \put(0.11490626,0.66188559){\makebox(0,0)[t]{\lineheight{1.25}\smash{\begin{tabular}[t]{c}\footnotesize{$10^{-1}$}\end{tabular}}}}%
    \put(0.11490626,0.496189){\makebox(0,0)[t]{\lineheight{1.25}\smash{\begin{tabular}[t]{c}\footnotesize{$10^{-2}$}\end{tabular}}}}%
    \put(0.11490626,0.16663563){\makebox(0,0)[t]{\lineheight{1.25}\smash{\begin{tabular}[t]{c}\footnotesize{$10^{-4}$}\end{tabular}}}}%
    \put(0.27135399,0.0518861){\makebox(0,0)[t]{\lineheight{1.25}\smash{\begin{tabular}[t]{c}\footnotesize{$10^{-3}$}\end{tabular}}}}%
    \put(0.46679907,0.0518861){\makebox(0,0)[t]{\lineheight{1.25}\smash{\begin{tabular}[t]{c}\footnotesize{$10^{-4}$}\end{tabular}}}}%
    \put(0.65680435,0.0518861){\makebox(0,0)[t]{\lineheight{1.25}\smash{\begin{tabular}[t]{c}\footnotesize{$10^{-3}$}\end{tabular}}}}%
    \put(0.8522489,0.0518861){\makebox(0,0)[t]{\lineheight{1.25}\smash{\begin{tabular}[t]{c}\footnotesize{$10^{-4}$}\end{tabular}}}}%
    \put(0.11490626,0.33063002){\makebox(0,0)[t]{\lineheight{1.25}\smash{\begin{tabular}[t]{c}\footnotesize{$10^{-3}$}\end{tabular}}}}%
  \end{picture}%
\endgroup%

%% file: DP_hidden.eps_tex
\begingroup%
  \makeatletter%
  \providecommand\color[2][]{%
    \errmessage{(Inkscape) Color is used for the text in Inkscape, but the package 'color.sty' is not loaded}%
    \renewcommand\color[2][]{}%
  }%
  \providecommand\transparent[1]{%
    \errmessage{(Inkscape) Transparency is used (non-zero) for the text in Inkscape, but the package 'transparent.sty' is not loaded}%
    \renewcommand\transparent[1]{}%
  }%
  \providecommand\rotatebox[2]{#2}%
  \newcommand*\fsize{\dimexpr\f@size pt\relax}%
  \newcommand*\lineheight[1]{\fontsize{\fsize}{#1\fsize}\selectfont}%
  \ifx\svgwidth\undefined%
    \setlength{\unitlength}{460.8bp}%
    \ifx\svgscale\undefined%
      \relax%
    \else%
      \setlength{\unitlength}{\unitlength * \real{\svgscale}}%
    \fi%
  \else%
    \setlength{\unitlength}{\svgwidth}%
  \fi%
  \global\let\svgwidth\undefined%
  \global\let\svgscale\undefined%
  \makeatother%
  \begin{picture}(1,0.75)%
    \lineheight{1}%
    \setlength\tabcolsep{0pt}%
    \put(0,0){\includegraphics[width=\unitlength]{DP_hidden.eps}}%
    \put(0.37380728,0.00892962){\color[rgb]{0,0,0}\makebox(0,0)[t]{\lineheight{1.25}\smash{\begin{tabular}[t]{c}GENERIC\end{tabular}}}}%
    \put(0.76114979,0.01061763){\color[rgb]{0,0,0}\makebox(0,0)[t]{\lineheight{1.25}\smash{\begin{tabular}[t]{c}Single generator\end{tabular}}}}%
    \put(0.03968709,0.35902595){\rotatebox{90}{\makebox(0,0)[lt]{\lineheight{1.25}\smash{\begin{tabular}[t]{l}MSE\end{tabular}}}}}%
    \put(0.11464017,0.52868677){\makebox(0,0)[t]{\lineheight{1.25}\smash{\begin{tabular}[t]{c}\footnotesize{$10^{-2}$}\end{tabular}}}}%
    \put(0.11464017,0.38815083){\makebox(0,0)[t]{\lineheight{1.25}\smash{\begin{tabular}[t]{c}\footnotesize{$10^{-3}$}\end{tabular}}}}%
    \put(0.11464017,0.10241766){\makebox(0,0)[t]{\lineheight{1.25}\smash{\begin{tabular}[t]{c}\footnotesize{$10^{-5}$}\end{tabular}}}}%
    \put(0.24262289,0.05216208){\makebox(0,0)[t]{\lineheight{1.25}\smash{\begin{tabular}[t]{c}\footnotesize{$50$}\end{tabular}}}}%
    \put(0.37178967,0.05216208){\makebox(0,0)[t]{\lineheight{1.25}\smash{\begin{tabular}[t]{c}\footnotesize{$100$}\end{tabular}}}}%
    \put(0.50095634,0.05216208){\makebox(0,0)[t]{\lineheight{1.25}\smash{\begin{tabular}[t]{c}\footnotesize{$200$}\end{tabular}}}}%
    \put(0.63012293,0.05216208){\makebox(0,0)[t]{\lineheight{1.25}\smash{\begin{tabular}[t]{c}\footnotesize{$50$}\end{tabular}}}}%
    \put(0.75928963,0.05216208){\makebox(0,0)[t]{\lineheight{1.25}\smash{\begin{tabular}[t]{c}\footnotesize{$100$}\end{tabular}}}}%
    \put(0.88845633,0.05216208){\makebox(0,0)[t]{\lineheight{1.25}\smash{\begin{tabular}[t]{c}\footnotesize{$200$}\end{tabular}}}}%
    \put(0.11464017,0.2440469){\makebox(0,0)[t]{\lineheight{1.25}\smash{\begin{tabular}[t]{c}\footnotesize{$10^{-4}$}\end{tabular}}}}%
  \end{picture}%
\endgroup%

%% file: DP_LearnedEnergy.eps_tex
\begingroup%
  \makeatletter%
  \providecommand\color[2][]{%
    \errmessage{(Inkscape) Color is used for the text in Inkscape, but the package 'color.sty' is not loaded}%
    \renewcommand\color[2][]{}%
  }%
  \providecommand\transparent[1]{%
    \errmessage{(Inkscape) Transparency is used (non-zero) for the text in Inkscape, but the package 'transparent.sty' is not loaded}%
    \renewcommand\transparent[1]{}%
  }%
  \providecommand\rotatebox[2]{#2}%
  \newcommand*\fsize{\dimexpr\f@size pt\relax}%
  \newcommand*\lineheight[1]{\fontsize{\fsize}{#1\fsize}\selectfont}%
  \ifx\svgwidth\undefined%
    \setlength{\unitlength}{1279.93841216bp}%
    \ifx\svgscale\undefined%
      \relax%
    \else%
      \setlength{\unitlength}{\unitlength * \real{\svgscale}}%
    \fi%
  \else%
    \setlength{\unitlength}{\svgwidth}%
  \fi%
  \global\let\svgwidth\undefined%
  \global\let\svgscale\undefined%
  \makeatother%
  \begin{picture}(1,0.75163901)%
    \lineheight{1}%
    \setlength\tabcolsep{0pt}%
    \put(0,0){\includegraphics[width=\unitlength]{DP_LearnedEnergy.eps}}%
    \put(0.17962054,0.00415095){\color[rgb]{0,0,0}\makebox(0,0)[t]{\lineheight{1.25}\smash{\begin{tabular}[t]{c}Snapshots\end{tabular}}}}%
    \put(0.18861312,0.24913831){\color[rgb]{0,0,0}\makebox(0,0)[t]{\lineheight{1.25}\smash{\begin{tabular}[t]{c}Single generator ($\Delta t =0.2$ s)\end{tabular}}}}%
    \put(0.01178163,0.13623614){\color[rgb]{0,0,0}\rotatebox{90}{\makebox(0,0)[t]{\lineheight{1.25}\smash{\begin{tabular}[t]{c}$\mathcal{F}$\end{tabular}}}}}%
    \put(0.05766109,0.02339653){\makebox(0,0)[t]{\lineheight{1.25}\smash{\begin{tabular}[t]{c}\footnotesize{$0$}\end{tabular}}}}%
    \put(0.10961237,0.02339653){\makebox(0,0)[t]{\lineheight{1.25}\smash{\begin{tabular}[t]{c}\footnotesize{$60$}\end{tabular}}}}%
    \put(0.16024546,0.02339653){\makebox(0,0)[t]{\lineheight{1.25}\smash{\begin{tabular}[t]{c}\footnotesize{$120$}\end{tabular}}}}%
    \put(0.21150601,0.02339653){\makebox(0,0)[t]{\lineheight{1.25}\smash{\begin{tabular}[t]{c}\footnotesize{$180$}\end{tabular}}}}%
    \put(0.26228535,0.02339653){\makebox(0,0)[t]{\lineheight{1.25}\smash{\begin{tabular}[t]{c}\footnotesize{$240$}\end{tabular}}}}%
    \put(0.31291843,0.02339653){\makebox(0,0)[t]{\lineheight{1.25}\smash{\begin{tabular}[t]{c}\footnotesize{$300$}\end{tabular}}}}%
    \put(0.05589478,0.09136683){\makebox(0,0)[rt]{\lineheight{1.25}\smash{\begin{tabular}[t]{r}\footnotesize{$ -0.6$}\end{tabular}}}}%
    \put(0.05589478,0.05134153){\makebox(0,0)[rt]{\lineheight{1.25}\smash{\begin{tabular}[t]{r}\footnotesize{$ -0.9$}\end{tabular}}}}%
    \put(0.05589478,0.13167692){\makebox(0,0)[rt]{\lineheight{1.25}\smash{\begin{tabular}[t]{r}\footnotesize{$ -0.3$}\end{tabular}}}}%
    \put(0.05589478,0.17209341){\makebox(0,0)[rt]{\lineheight{1.25}\smash{\begin{tabular}[t]{r}\footnotesize{$ 0.0$}\end{tabular}}}}%
    \put(0.05589478,0.21212395){\makebox(0,0)[rt]{\lineheight{1.25}\smash{\begin{tabular}[t]{r}\footnotesize{$ 0.3$}\end{tabular}}}}%
    \put(0.83596386,0.74369336){\color[rgb]{0,0,0}\makebox(0,0)[t]{\lineheight{1.25}\smash{\begin{tabular}[t]{c}GENERIC ($\Delta t =1.2$ s)\end{tabular}}}}%
    \put(0.67040702,0.62844733){\color[rgb]{0,0,0}\rotatebox{90}{\makebox(0,0)[t]{\lineheight{1.25}\smash{\begin{tabular}[t]{c}$\mathcal{H}$\end{tabular}}}}}%
    \put(0.71438743,0.51560771){\makebox(0,0)[t]{\lineheight{1.25}\smash{\begin{tabular}[t]{c}\footnotesize{$0$}\end{tabular}}}}%
    \put(0.76516676,0.51560771){\makebox(0,0)[t]{\lineheight{1.25}\smash{\begin{tabular}[t]{c}\footnotesize{$10$}\end{tabular}}}}%
    \put(0.81579984,0.51560771){\makebox(0,0)[t]{\lineheight{1.25}\smash{\begin{tabular}[t]{c}\footnotesize{$20$}\end{tabular}}}}%
    \put(0.8670604,0.51560771){\makebox(0,0)[t]{\lineheight{1.25}\smash{\begin{tabular}[t]{c}\footnotesize{$30$}\end{tabular}}}}%
    \put(0.91783974,0.51560771){\makebox(0,0)[t]{\lineheight{1.25}\smash{\begin{tabular}[t]{c}\footnotesize{$40$}\end{tabular}}}}%
    \put(0.96847282,0.51560771){\makebox(0,0)[t]{\lineheight{1.25}\smash{\begin{tabular}[t]{c}\footnotesize{$50$}\end{tabular}}}}%
    \put(0.71027722,0.59764118){\makebox(0,0)[rt]{\lineheight{1.25}\smash{\begin{tabular}[t]{r}\footnotesize{$ -0.2$}\end{tabular}}}}%
    \put(0.71027722,0.6309197){\makebox(0,0)[rt]{\lineheight{1.25}\smash{\begin{tabular}[t]{r}\footnotesize{$ 0.0$}\end{tabular}}}}%
    \put(0.71027722,0.66430459){\makebox(0,0)[rt]{\lineheight{1.25}\smash{\begin{tabular}[t]{r}\footnotesize{$ 0.2$}\end{tabular}}}}%
    \put(0.71027722,0.69730354){\makebox(0,0)[rt]{\lineheight{1.25}\smash{\begin{tabular}[t]{r}\footnotesize{$ 0.4$}\end{tabular}}}}%
    \put(0.71027722,0.56401011){\makebox(0,0)[rt]{\lineheight{1.25}\smash{\begin{tabular}[t]{r}\footnotesize{$ -0.4$}\end{tabular}}}}%
    \put(0.51675389,0.74369336){\color[rgb]{0,0,0}\makebox(0,0)[t]{\lineheight{1.25}\smash{\begin{tabular}[t]{c}GENERIC ($\Delta t =0.3$ s)\end{tabular}}}}%
    \put(0.33992241,0.62844733){\color[rgb]{0,0,0}\rotatebox{90}{\makebox(0,0)[t]{\lineheight{1.25}\smash{\begin{tabular}[t]{c}$\mathcal{H}$\end{tabular}}}}}%
    \put(0.39517746,0.51560771){\makebox(0,0)[t]{\lineheight{1.25}\smash{\begin{tabular}[t]{c}\footnotesize{$0$}\end{tabular}}}}%
    \put(0.43775315,0.51560771){\makebox(0,0)[t]{\lineheight{1.25}\smash{\begin{tabular}[t]{c}\footnotesize{$40$}\end{tabular}}}}%
    \put(0.48838623,0.51560771){\makebox(0,0)[t]{\lineheight{1.25}\smash{\begin{tabular}[t]{c}\footnotesize{$80$}\end{tabular}}}}%
    \put(0.53964678,0.51560771){\makebox(0,0)[t]{\lineheight{1.25}\smash{\begin{tabular}[t]{c}\footnotesize{$120$}\end{tabular}}}}%
    \put(0.59042612,0.51560771){\makebox(0,0)[t]{\lineheight{1.25}\smash{\begin{tabular}[t]{c}\footnotesize{$160$}\end{tabular}}}}%
    \put(0.6410592,0.51560771){\makebox(0,0)[t]{\lineheight{1.25}\smash{\begin{tabular}[t]{c}\footnotesize{$200$}\end{tabular}}}}%
    \put(0.38403556,0.59764118){\makebox(0,0)[rt]{\lineheight{1.25}\smash{\begin{tabular}[t]{r}\footnotesize{$ -0.2$}\end{tabular}}}}%
    \put(0.38403556,0.6309197){\makebox(0,0)[rt]{\lineheight{1.25}\smash{\begin{tabular}[t]{r}\footnotesize{$ 0.0$}\end{tabular}}}}%
    \put(0.38403556,0.66430459){\makebox(0,0)[rt]{\lineheight{1.25}\smash{\begin{tabular}[t]{r}\footnotesize{$ 0.2$}\end{tabular}}}}%
    \put(0.38403556,0.69730354){\makebox(0,0)[rt]{\lineheight{1.25}\smash{\begin{tabular}[t]{r}\footnotesize{$ 0.4$}\end{tabular}}}}%
    \put(0.38403556,0.56401011){\makebox(0,0)[rt]{\lineheight{1.25}\smash{\begin{tabular}[t]{r}\footnotesize{$ -0.4$}\end{tabular}}}}%
    \put(0.18861312,0.74369336){\color[rgb]{0,0,0}\makebox(0,0)[t]{\lineheight{1.25}\smash{\begin{tabular}[t]{c}GENERIC ($\Delta t =0.2$ s)\end{tabular}}}}%
    \put(0.01178163,0.62844733){\color[rgb]{0,0,0}\rotatebox{90}{\makebox(0,0)[t]{\lineheight{1.25}\smash{\begin{tabular}[t]{c}$\mathcal{H}$\end{tabular}}}}}%
    \put(0.05766109,0.51560771){\makebox(0,0)[t]{\lineheight{1.25}\smash{\begin{tabular}[t]{c}\footnotesize{$0$}\end{tabular}}}}%
    \put(0.10961237,0.51560771){\makebox(0,0)[t]{\lineheight{1.25}\smash{\begin{tabular}[t]{c}\footnotesize{$60$}\end{tabular}}}}%
    \put(0.16024546,0.51560771){\makebox(0,0)[t]{\lineheight{1.25}\smash{\begin{tabular}[t]{c}\footnotesize{$120$}\end{tabular}}}}%
    \put(0.21150601,0.51560771){\makebox(0,0)[t]{\lineheight{1.25}\smash{\begin{tabular}[t]{c}\footnotesize{$180$}\end{tabular}}}}%
    \put(0.26228535,0.51560771){\makebox(0,0)[t]{\lineheight{1.25}\smash{\begin{tabular}[t]{c}\footnotesize{$240$}\end{tabular}}}}%
    \put(0.31291843,0.51560771){\makebox(0,0)[t]{\lineheight{1.25}\smash{\begin{tabular}[t]{c}\footnotesize{$300$}\end{tabular}}}}%
    \put(0.05589478,0.59764118){\makebox(0,0)[rt]{\lineheight{1.25}\smash{\begin{tabular}[t]{r}\footnotesize{$ -0.2$}\end{tabular}}}}%
    \put(0.05589478,0.6309197){\makebox(0,0)[rt]{\lineheight{1.25}\smash{\begin{tabular}[t]{r}\footnotesize{$ 0.0$}\end{tabular}}}}%
    \put(0.05589478,0.66430459){\makebox(0,0)[rt]{\lineheight{1.25}\smash{\begin{tabular}[t]{r}\footnotesize{$ 0.2$}\end{tabular}}}}%
    \put(0.05589478,0.69730354){\makebox(0,0)[rt]{\lineheight{1.25}\smash{\begin{tabular}[t]{r}\footnotesize{$ 0.4$}\end{tabular}}}}%
    \put(0.05589478,0.56401011){\makebox(0,0)[rt]{\lineheight{1.25}\smash{\begin{tabular}[t]{r}\footnotesize{$ -0.4$}\end{tabular}}}}%
    \put(0.18861312,0.49758777){\color[rgb]{0,0,0}\makebox(0,0)[t]{\lineheight{1.25}\smash{\begin{tabular}[t]{c}GENERIC ($\Delta t =0.2$ s)\end{tabular}}}}%
    \put(0.01178163,0.38234173){\color[rgb]{0,0,0}\rotatebox{90}{\makebox(0,0)[t]{\lineheight{1.25}\smash{\begin{tabular}[t]{c}$\mathcal{S}$\end{tabular}}}}}%
    \put(0.05766109,0.26950212){\makebox(0,0)[t]{\lineheight{1.25}\smash{\begin{tabular}[t]{c}\footnotesize{$0$}\end{tabular}}}}%
    \put(0.10961237,0.26950212){\makebox(0,0)[t]{\lineheight{1.25}\smash{\begin{tabular}[t]{c}\footnotesize{$60$}\end{tabular}}}}%
    \put(0.16024546,0.26950212){\makebox(0,0)[t]{\lineheight{1.25}\smash{\begin{tabular}[t]{c}\footnotesize{$120$}\end{tabular}}}}%
    \put(0.21150601,0.26950212){\makebox(0,0)[t]{\lineheight{1.25}\smash{\begin{tabular}[t]{c}\footnotesize{$180$}\end{tabular}}}}%
    \put(0.26228535,0.26950212){\makebox(0,0)[t]{\lineheight{1.25}\smash{\begin{tabular}[t]{c}\footnotesize{$240$}\end{tabular}}}}%
    \put(0.31291843,0.26950212){\makebox(0,0)[t]{\lineheight{1.25}\smash{\begin{tabular}[t]{c}\footnotesize{$300$}\end{tabular}}}}%
    \put(0.05589478,0.32458117){\makebox(0,0)[rt]{\lineheight{1.25}\smash{\begin{tabular}[t]{r}\footnotesize{$ -0.4$}\end{tabular}}}}%
    \put(0.05589478,0.36489126){\makebox(0,0)[rt]{\lineheight{1.25}\smash{\begin{tabular}[t]{r}\footnotesize{$ -0.3$}\end{tabular}}}}%
    \put(0.05589478,0.40530774){\makebox(0,0)[rt]{\lineheight{1.25}\smash{\begin{tabular}[t]{r}\footnotesize{$ -0.2$}\end{tabular}}}}%
    \put(0.05589478,0.44533829){\makebox(0,0)[rt]{\lineheight{1.25}\smash{\begin{tabular}[t]{r}\footnotesize{$ -0.1$}\end{tabular}}}}%
    \put(0.51675389,0.49758777){\color[rgb]{0,0,0}\makebox(0,0)[t]{\lineheight{1.25}\smash{\begin{tabular}[t]{c}GENERIC ($\Delta t =0.3$ s)\end{tabular}}}}%
    \put(0.33992241,0.38234173){\color[rgb]{0,0,0}\rotatebox{90}{\makebox(0,0)[t]{\lineheight{1.25}\smash{\begin{tabular}[t]{c}$\mathcal{S}$\end{tabular}}}}}%
    \put(0.38580186,0.26950212){\makebox(0,0)[t]{\lineheight{1.25}\smash{\begin{tabular}[t]{c}\footnotesize{$0$}\end{tabular}}}}%
    \put(0.43775315,0.26950212){\makebox(0,0)[t]{\lineheight{1.25}\smash{\begin{tabular}[t]{c}\footnotesize{$40$}\end{tabular}}}}%
    \put(0.48838623,0.26950212){\makebox(0,0)[t]{\lineheight{1.25}\smash{\begin{tabular}[t]{c}\footnotesize{$80$}\end{tabular}}}}%
    \put(0.53964678,0.26950212){\makebox(0,0)[t]{\lineheight{1.25}\smash{\begin{tabular}[t]{c}\footnotesize{$120$}\end{tabular}}}}%
    \put(0.59042612,0.26950212){\makebox(0,0)[t]{\lineheight{1.25}\smash{\begin{tabular}[t]{c}\footnotesize{$160$}\end{tabular}}}}%
    \put(0.6410592,0.26950212){\makebox(0,0)[t]{\lineheight{1.25}\smash{\begin{tabular}[t]{c}\footnotesize{$200$}\end{tabular}}}}%
    \put(0.38403556,0.32458117){\makebox(0,0)[rt]{\lineheight{1.25}\smash{\begin{tabular}[t]{r}\footnotesize{$ -0.4$}\end{tabular}}}}%
    \put(0.38403556,0.36489126){\makebox(0,0)[rt]{\lineheight{1.25}\smash{\begin{tabular}[t]{r}\footnotesize{$ -0.3$}\end{tabular}}}}%
    \put(0.38403556,0.40530774){\makebox(0,0)[rt]{\lineheight{1.25}\smash{\begin{tabular}[t]{r}\footnotesize{$ -0.2$}\end{tabular}}}}%
    \put(0.38403556,0.44533829){\makebox(0,0)[rt]{\lineheight{1.25}\smash{\begin{tabular}[t]{r}\footnotesize{$ -0.1$}\end{tabular}}}}%
    \put(0.84489467,0.49758777){\color[rgb]{0,0,0}\makebox(0,0)[t]{\lineheight{1.25}\smash{\begin{tabular}[t]{c}GENERIC ($\Delta t =1.2$ s)\end{tabular}}}}%
    \put(0.67040708,0.38234173){\color[rgb]{0,0,0}\rotatebox{90}{\makebox(0,0)[t]{\lineheight{1.25}\smash{\begin{tabular}[t]{c}$\mathcal{S}$\end{tabular}}}}}%
    \put(0.71394263,0.26950212){\makebox(0,0)[t]{\lineheight{1.25}\smash{\begin{tabular}[t]{c}\footnotesize{$0$}\end{tabular}}}}%
    \put(0.76589392,0.26950212){\makebox(0,0)[t]{\lineheight{1.25}\smash{\begin{tabular}[t]{c}\footnotesize{$10$}\end{tabular}}}}%
    \put(0.816527,0.26950212){\makebox(0,0)[t]{\lineheight{1.25}\smash{\begin{tabular}[t]{c}\footnotesize{$20$}\end{tabular}}}}%
    \put(0.86778755,0.26950212){\makebox(0,0)[t]{\lineheight{1.25}\smash{\begin{tabular}[t]{c}\footnotesize{$30$}\end{tabular}}}}%
    \put(0.91856689,0.26950212){\makebox(0,0)[t]{\lineheight{1.25}\smash{\begin{tabular}[t]{c}\footnotesize{$40$}\end{tabular}}}}%
    \put(0.96919997,0.26950212){\makebox(0,0)[t]{\lineheight{1.25}\smash{\begin{tabular}[t]{c}\footnotesize{$50$}\end{tabular}}}}%
    \put(0.71217633,0.32458117){\makebox(0,0)[rt]{\lineheight{1.25}\smash{\begin{tabular}[t]{r}\footnotesize{$ -0.4$}\end{tabular}}}}%
    \put(0.71217633,0.36489126){\makebox(0,0)[rt]{\lineheight{1.25}\smash{\begin{tabular}[t]{r}\footnotesize{$ -0.3$}\end{tabular}}}}%
    \put(0.71217633,0.40530774){\makebox(0,0)[rt]{\lineheight{1.25}\smash{\begin{tabular}[t]{r}\footnotesize{$ -0.2$}\end{tabular}}}}%
    \put(0.71217633,0.44533829){\makebox(0,0)[rt]{\lineheight{1.25}\smash{\begin{tabular}[t]{r}\footnotesize{$ -0.1$}\end{tabular}}}}%
    \put(0.84489466,0.24913831){\color[rgb]{0,0,0}\makebox(0,0)[t]{\lineheight{1.25}\smash{\begin{tabular}[t]{c}Single generator ($\Delta t =1.2$ s)\end{tabular}}}}%
    \put(0.67040708,0.13623614){\color[rgb]{0,0,0}\rotatebox{90}{\makebox(0,0)[t]{\lineheight{1.25}\smash{\begin{tabular}[t]{c}$\mathcal{F}$\end{tabular}}}}}%
    \put(0.71394263,0.02339653){\makebox(0,0)[t]{\lineheight{1.25}\smash{\begin{tabular}[t]{c}\footnotesize{$0$}\end{tabular}}}}%
    \put(0.76589392,0.02339653){\makebox(0,0)[t]{\lineheight{1.25}\smash{\begin{tabular}[t]{c}\footnotesize{$10$}\end{tabular}}}}%
    \put(0.816527,0.02339653){\makebox(0,0)[t]{\lineheight{1.25}\smash{\begin{tabular}[t]{c}\footnotesize{$20$}\end{tabular}}}}%
    \put(0.86778755,0.02339653){\makebox(0,0)[t]{\lineheight{1.25}\smash{\begin{tabular}[t]{c}\footnotesize{$30$}\end{tabular}}}}%
    \put(0.91856689,0.02339653){\makebox(0,0)[t]{\lineheight{1.25}\smash{\begin{tabular}[t]{c}\footnotesize{$40$}\end{tabular}}}}%
    \put(0.96919997,0.02339653){\makebox(0,0)[t]{\lineheight{1.25}\smash{\begin{tabular}[t]{c}\footnotesize{$50$}\end{tabular}}}}%
    \put(0.71217633,0.09136683){\makebox(0,0)[rt]{\lineheight{1.25}\smash{\begin{tabular}[t]{r}\footnotesize{$ -0.6$}\end{tabular}}}}%
    \put(0.71217633,0.05134153){\makebox(0,0)[rt]{\lineheight{1.25}\smash{\begin{tabular}[t]{r}\footnotesize{$ -0.9$}\end{tabular}}}}%
    \put(0.71217633,0.13167692){\makebox(0,0)[rt]{\lineheight{1.25}\smash{\begin{tabular}[t]{r}\footnotesize{$ -0.3$}\end{tabular}}}}%
    \put(0.71217633,0.17209341){\makebox(0,0)[rt]{\lineheight{1.25}\smash{\begin{tabular}[t]{r}\footnotesize{$ 0.0$}\end{tabular}}}}%
    \put(0.71217633,0.21212395){\makebox(0,0)[rt]{\lineheight{1.25}\smash{\begin{tabular}[t]{r}\footnotesize{$ 0.3$}\end{tabular}}}}%
    \put(0.51675389,0.24913831){\color[rgb]{0,0,0}\makebox(0,0)[t]{\lineheight{1.25}\smash{\begin{tabular}[t]{c}Single generator ($\Delta t =0.3$ s)\end{tabular}}}}%
    \put(0.33992241,0.13623614){\color[rgb]{0,0,0}\rotatebox{90}{\makebox(0,0)[t]{\lineheight{1.25}\smash{\begin{tabular}[t]{c}$\mathcal{F}$\end{tabular}}}}}%
    \put(0.38580186,0.02339653){\makebox(0,0)[t]{\lineheight{1.25}\smash{\begin{tabular}[t]{c}\footnotesize{$0$}\end{tabular}}}}%
    \put(0.43775315,0.02339653){\makebox(0,0)[t]{\lineheight{1.25}\smash{\begin{tabular}[t]{c}\footnotesize{$40$}\end{tabular}}}}%
    \put(0.48838623,0.02339653){\makebox(0,0)[t]{\lineheight{1.25}\smash{\begin{tabular}[t]{c}\footnotesize{$80$}\end{tabular}}}}%
    \put(0.53964678,0.02339653){\makebox(0,0)[t]{\lineheight{1.25}\smash{\begin{tabular}[t]{c}\footnotesize{$120$}\end{tabular}}}}%
    \put(0.59042612,0.02339653){\makebox(0,0)[t]{\lineheight{1.25}\smash{\begin{tabular}[t]{c}\footnotesize{$160$}\end{tabular}}}}%
    \put(0.6410592,0.0257404){\makebox(0,0)[t]{\lineheight{1.25}\smash{\begin{tabular}[t]{c}\footnotesize{$200$}\end{tabular}}}}%
    \put(0.38403556,0.09136683){\makebox(0,0)[rt]{\lineheight{1.25}\smash{\begin{tabular}[t]{r}\footnotesize{$ -0.6$}\end{tabular}}}}%
    \put(0.38403556,0.05134153){\makebox(0,0)[rt]{\lineheight{1.25}\smash{\begin{tabular}[t]{r}\footnotesize{$ -0.9$}\end{tabular}}}}%
    \put(0.38403556,0.13167692){\makebox(0,0)[rt]{\lineheight{1.25}\smash{\begin{tabular}[t]{r}\footnotesize{$ -0.3$}\end{tabular}}}}%
    \put(0.38403556,0.17209341){\makebox(0,0)[rt]{\lineheight{1.25}\smash{\begin{tabular}[t]{r}\footnotesize{$ 0.0$}\end{tabular}}}}%
    \put(0.38403556,0.21212395){\makebox(0,0)[rt]{\lineheight{1.25}\smash{\begin{tabular}[t]{r}\footnotesize{$ 0.3$}\end{tabular}}}}%
    \put(0.50776131,0.00415095){\color[rgb]{0,0,0}\makebox(0,0)[t]{\lineheight{1.25}\smash{\begin{tabular}[t]{c}Snapshots\end{tabular}}}}%
    \put(0.83590208,0.00415095){\color[rgb]{0,0,0}\makebox(0,0)[t]{\lineheight{1.25}\smash{\begin{tabular}[t]{c}Snapshots\end{tabular}}}}%
  \end{picture}%
\endgroup%

%% file: VE_RexxWexx.eps_tex
\begingroup%
  \makeatletter%
  \providecommand\color[2][]{%
    \errmessage{(Inkscape) Color is used for the text in Inkscape, but the package 'color.sty' is not loaded}%
    \renewcommand\color[2][]{}%
  }%
  \providecommand\transparent[1]{%
    \errmessage{(Inkscape) Transparency is used (non-zero) for the text in Inkscape, but the package 'transparent.sty' is not loaded}%
    \renewcommand\transparent[1]{}%
  }%
  \providecommand\rotatebox[2]{#2}%
  \newcommand*\fsize{\dimexpr\f@size pt\relax}%
  \newcommand*\lineheight[1]{\fontsize{\fsize}{#1\fsize}\selectfont}%
  \ifx\svgwidth\undefined%
    \setlength{\unitlength}{921.59999bp}%
    \ifx\svgscale\undefined%
      \relax%
    \else%
      \setlength{\unitlength}{\unitlength * \real{\svgscale}}%
    \fi%
  \else%
    \setlength{\unitlength}{\svgwidth}%
  \fi%
  \global\let\svgwidth\undefined%
  \global\let\svgscale\undefined%
  \makeatother%
  \begin{picture}(1,0.37575958)%
    \lineheight{1}%
    \setlength\tabcolsep{0pt}%
    \put(0,0){\includegraphics[width=\unitlength]{VE_RexxWexx.eps}}%
    \put(0.01489766,0.35154374){\makebox(0,0)[t]{\lineheight{1.25}\smash{\begin{tabular}[t]{c}(a)\end{tabular}}}}%
    \put(0.51489764,0.35154374){\makebox(0,0)[t]{\lineheight{1.25}\smash{\begin{tabular}[t]{c}(b)\end{tabular}}}}%
    \put(0.18690363,0.00106483){\color[rgb]{0,0,0}\makebox(0,0)[t]{\lineheight{1.25}\smash{\begin{tabular}[t]{c}GENERIC\end{tabular}}}}%
    \put(0.38057487,0.00106483){\color[rgb]{0,0,0}\makebox(0,0)[t]{\lineheight{1.25}\smash{\begin{tabular}[t]{c}Single generator\end{tabular}}}}%
    \put(0.01984354,0.17951301){\rotatebox{90}{\makebox(0,0)[lt]{\lineheight{1.25}\smash{\begin{tabular}[t]{l}MSE\end{tabular}}}}}%
    \put(0.05732009,0.31289057){\makebox(0,0)[t]{\lineheight{1.25}\smash{\begin{tabular}[t]{c}\footnotesize{$10^{-2}$}\end{tabular}}}}%
    \put(0.05732009,0.13584432){\makebox(0,0)[t]{\lineheight{1.25}\smash{\begin{tabular}[t]{c}\footnotesize{$10^{-4}$}\end{tabular}}}}%
    \put(0.05732009,0.04684792){\makebox(0,0)[t]{\lineheight{1.25}\smash{\begin{tabular}[t]{c}\footnotesize{$10^{-5}$}\end{tabular}}}}%
    \put(0.12131147,0.0260811){\makebox(0,0)[t]{\lineheight{1.25}\smash{\begin{tabular}[t]{c}\footnotesize{$\textit{We}1.0$}\end{tabular}}}}%
    \put(0.18589482,0.0260811){\makebox(0,0)[t]{\lineheight{1.25}\smash{\begin{tabular}[t]{c}\footnotesize{$\textit{We}1.5$}\end{tabular}}}}%
    \put(0.25047815,0.0260811){\makebox(0,0)[t]{\lineheight{1.25}\smash{\begin{tabular}[t]{c}\footnotesize{$\textit{We}2.5$}\end{tabular}}}}%
    \put(0.31506147,0.0260811){\makebox(0,0)[t]{\lineheight{1.25}\smash{\begin{tabular}[t]{c}\footnotesize{$\textit{We}1.0$}\end{tabular}}}}%
    \put(0.37964478,0.0260811){\makebox(0,0)[t]{\lineheight{1.25}\smash{\begin{tabular}[t]{c}\footnotesize{$\textit{We}1.5$}\end{tabular}}}}%
    \put(0.44422813,0.0260811){\makebox(0,0)[t]{\lineheight{1.25}\smash{\begin{tabular}[t]{c}\footnotesize{$\textit{We}2.5$}\end{tabular}}}}%
    \put(0.05732009,0.22456256){\makebox(0,0)[t]{\lineheight{1.25}\smash{\begin{tabular}[t]{c}\footnotesize{$10^{-3}$}\end{tabular}}}}%
    \put(0.6869036,0.00106483){\color[rgb]{0,0,0}\makebox(0,0)[t]{\lineheight{1.25}\smash{\begin{tabular}[t]{c}GENERIC\end{tabular}}}}%
    \put(0.88057486,0.00106483){\color[rgb]{0,0,0}\makebox(0,0)[t]{\lineheight{1.25}\smash{\begin{tabular}[t]{c}Single generator\end{tabular}}}}%
    \put(0.51984349,0.18200866){\rotatebox{90}{\makebox(0,0)[lt]{\lineheight{1.25}\smash{\begin{tabular}[t]{l}MSE\end{tabular}}}}}%
    \put(0.55732011,0.15461598){\makebox(0,0)[t]{\lineheight{1.25}\smash{\begin{tabular}[t]{c}\footnotesize{$10^{-4}$}\end{tabular}}}}%
    \put(0.55732011,0.05857134){\makebox(0,0)[t]{\lineheight{1.25}\smash{\begin{tabular}[t]{c}\footnotesize{$10^{-5}$}\end{tabular}}}}%
    \put(0.62131148,0.02857676){\makebox(0,0)[t]{\lineheight{1.25}\smash{\begin{tabular}[t]{c}\footnotesize{$\textit{Re}0.1$}\end{tabular}}}}%
    \put(0.68589479,0.02857676){\makebox(0,0)[t]{\lineheight{1.25}\smash{\begin{tabular}[t]{c}\footnotesize{$\textit{Re}0.4$}\end{tabular}}}}%
    \put(0.75047811,0.02857676){\makebox(0,0)[t]{\lineheight{1.25}\smash{\begin{tabular}[t]{c}\footnotesize{$\textit{Re}0.8$}\end{tabular}}}}%
    \put(0.81506149,0.02857676){\makebox(0,0)[t]{\lineheight{1.25}\smash{\begin{tabular}[t]{c}\footnotesize{$\textit{Re}0.1$}\end{tabular}}}}%
    \put(0.87964481,0.02857676){\makebox(0,0)[t]{\lineheight{1.25}\smash{\begin{tabular}[t]{c}\footnotesize{$\textit{Re}0.4$}\end{tabular}}}}%
    \put(0.94422819,0.02857676){\makebox(0,0)[t]{\lineheight{1.25}\smash{\begin{tabular}[t]{c}\footnotesize{$\textit{Re}0.8$}\end{tabular}}}}%
    \put(0.55732011,0.25147229){\makebox(0,0)[t]{\lineheight{1.25}\smash{\begin{tabular}[t]{c}\footnotesize{$10^{-3}$}\end{tabular}}}}%
  \end{picture}%
\endgroup%